%% file: main.tex
\documentclass[sigplan,screen,authorversion,10pt]{acmart}

\include{header}

\AtBeginDocument{%
  \providecommand\BibTeX{{%
    \normalfont B\kern-0.5em{\scshape i\kern-0.25em b}\kern-0.8em\TeX}}}

\copyrightyear{2020}
\acmYear{2020}
\setcopyright{acmcopyright}
\acmConference[Middleware '20]{21st International
Middleware Conference}{December 7--11, 2020}{Delft, Netherlands}
\acmBooktitle{21st International Middleware Conference (Middleware '20), December
7--11, 2020, Delft, Netherlands}
\acmPrice{15.00}
\acmDOI{10.1145/3423211.3425685}
\acmISBN{978-1-4503-8153-6/20/12}

\begin{document}

\title{\system: Online Federated Learning via Staleness Awareness and Performance Prediction}

\author{Georgios Damaskinos}
\email{georgios.damaskinos@epfl.ch}
\affiliation{%
  \institution{EPFL, Switzerland}
}

\author{Rachid Guerraoui}
\email{rachid.guerraoui@epfl.ch}
\affiliation{%
  \institution{EPFL, Switzerland}
}

\author{Anne-Marie Kermarrec}
\email{anne-marie.kermarrec@epfl.ch}
\affiliation{%
  \institution{EPFL, Switzerland}
}

\author{Vlad Nitu}
\authornote{Work conducted while at EPFL as a postdoctoral researcher.}
\email{vlad.nitu@insa-lyon.fr}
\affiliation{%
  \institution{INSA Lyon, France}
}

\author{Rhicheek Patra}
\email{rhicheek.patra@epfl.ch}
\affiliation{%
  \institution{EPFL, Switzerland}
}

\author{Francois Taiani}
\email{francois.taiani@irisa.fr}
\affiliation{%
  \institution{Univ Rennes, Inria, CNRS, IRISA}
}

\renewcommand{\shortauthors}{Georgios Damaskinos, et al.}

\input{Abstract}




\maketitle
\input{Intro}
\input{Framework}
\input{Evaluation}
\input{RelWork}

\input{Conclusion}

\bibliographystyle{ACM-Reference-Format}
\bibliography{bib}

\end{document}

%% file: header.tex
\usepackage{graphicx}
\usepackage{diagbox}
\usepackage{xr,xfrac,xspace,mathtools,bm}
\usepackage{thmtools, thm-restate}
\usepackage{xspace}
\usepackage{epstopdf}
\usepackage{booktabs}
\usepackage{epsfig}
\usepackage[normalem]{ulem}
\PassOptionsToPackage{hyphens}{url}
\usepackage{hyperref}
\usepackage{url}
\usepackage[font=small]{subfig}
\usepackage{float}
\usepackage[noabbrev,capitalise]{cleveref}
\usepackage{xcolor,colortbl}
\usepackage{lipsum}
\usepackage{capt-of}
\usepackage{algorithm} %
\usepackage[noend]{algorithmic}
\usepackage{fancyhdr}
\usepackage{multirow}
\usepackage[normalem]{ulem}
\usepackage{soul}
\usepackage[utf8]{inputenc}
\usepackage{relsize}
\usepackage{stmaryrd}
\usepackage{wrapfig}
\usepackage{array}
\usepackage[inline]{enumitem}
\usepackage{tikz}
\usepackage{nth}

\usetikzlibrary{arrows.meta,calc,backgrounds,angles,quotes}

\makeatletter

\def\NAT@def@citea{\def\@citea{\NAT@separator}} %

\creflabelformat{equation}{#2#1#3} %

\crefformat{section}{\S#2#1#3} %
\crefformat{subsection}{\S#2#1#3}
\crefformat{footnote}{#2\footnotemark[#1]#3}

\newcommand{\nospaceitemize}{\vspace{-0.5\topsep}}

\newlist{myitemize}{itemize}{1}
\setlist[myitemize,1]{label=\textbullet,leftmargin=0.2in,wide, labelwidth=!, labelindent=0pt}

\newlist{myenumerate}{enumerate}{1}
\setlist[myenumerate,1]{label*=(\arabic*),leftmargin=0.2in,wide, labelwidth=!, labelindent=0pt}

\newenvironment{myalign*}{\par\nobreak\noindent\nonumber\align}{\endalign}

\newcounter{letter}
\@whilenum\value{letter}<27\do{%
  \expandafter\edef\csname \Alph{letter}v\endcsname{\noexpand\bm{\Alph{letter}}}
  \expandafter\edef\csname c\Alph{letter}\endcsname{\noexpand\mathcal{\Alph{letter}}}
  \expandafter\edef\csname bb\Alph{letter}\endcsname{\noexpand\mathbb{\Alph{letter}}}
  \expandafter\edef\csname \alph{letter}v\endcsname{\noexpand\bm{\alph{letter}}}
  \stepcounter{letter}%
}

\newcommand{\thetav}{{\boldsymbol \theta}}

\newcommand\footnoteref[1]{\protected@xdef\@thefnmark{\ref{#1}}\@footnotemark}

\allowdisplaybreaks

\newcommand*{\rom}[1]{\expandafter\@slowromancap\romannumeral #1@}
\newcommand{\norm}[1]{\left\lVert#1\right\rVert}

\renewcommand\paragraph{\@startsection{paragraph}{4}{\parindent}%
  {0\baselineskip \@plus 0\p@ \@minus 0\p@}%
  {-3.5\p@}%
  {\ACM@NRadjust{\@parfont}\bfseries}}

\makeatother

\newcommand{\annote}[3]{{\color{#3}%
		\colorbox{#3}{\bfseries\sffamily\tiny\textcolor{white}{#2}}
		\color{#3}
		$\blacktriangleright$\footnotesize\emph{#1}$\blacktriangleleft$}%
}

\newboolean{withinSout}
\setboolean{withinSout}{false}

\newcommand{\ftasout}[1]{%
  \ifthenelse{\boolean{withinSout}}{
    #1
  }
  {%
    \setboolean{withinSout}{true}\sout{#1}\setboolean{withinSout}{false}%
  }%
}%

\DeclareRobustCommand{\hsout}[1]{\texorpdfstring{\ftasout{#1}}{#1}}

\DeclareRobustCommand{\change}[4]{{%
    {\color{#4}#1}\ifx\relax#2\relax\else{ \color{gray}{\hsout{#2}}}\fi%
  }%
}

\renewcommand{\annote}[3]{}

\newcommand{\system}{\textsc{\mbox{FLeet}}\xspace}
\newcommand{\profiler}{\textsc{\mbox{I-Prof}}\xspace}

\newcommand{\dynsgd}{\textsc{\mbox{DynSGD}}\xspace}
\newcommand{\algo}{\textsc{\mbox{AdaSGD}}\xspace}
\newcommand{\baselinesgd}{\textsc{\mbox{FedAvg}}\xspace}

%% file: Abstract.tex
\begin{abstract}

Federated Learning (FL) is very appealing for its privacy benefits: 
essentially, a global model is trained with updates computed on mobile devices while keeping the data of users local. 
Standard FL infrastructures are however designed to have no energy or performance impact on mobile devices, and are therefore not suitable for applications that require frequent (\emph{online}) model updates, such as news recommenders.%

This paper presents \system, the first \emph{Online FL} system, acting as a middleware between the Android OS and the machine learning application.
\system combines the privacy of Standard FL with the precision of online learning thanks to two core components: (i) \profiler, a new lightweight profiler that predicts and controls the impact of learning tasks on mobile devices, and
(ii) \algo, a new adaptive learning algorithm that is resilient to delayed updates. %

Our extensive evaluation shows that Online FL, as implemented by \system, can deliver a 2.3$\times$ quality boost compared to Standard FL, while only consuming 0.036\% of the battery per day.
\profiler can accurately control the impact of learning tasks by improving the prediction accuracy up to 3.6$\times$ (computation time) and up to 19$\times$ (energy). 
\algo outperforms alternative FL approaches by 18.4\% in terms of convergence speed on heterogeneous data.

\end{abstract}

%% file: Intro.tex
\section{Introduction}
\label{sec:intro}
The number of edge devices and the data produced by these devices have grown tremendously over the last 10 years.
While in 2009, mobile phones only generated 0.7\% of the worldwide data traffic, in 2018 this number exceeded 50\%~\cite{traffic}.
This exponential growth is raising challenges both in terms of scalability and privacy.
As the volume of data produced by mobile devices explodes, users expose increasingly detailed and sensitive information, which in turn becomes more costly to store, process, and protect.
This dual challenge of privacy and scalability is pervasive in machine learning (ML) applications such as recommenders, image-recognition apps, and personal assistants.
These ML-based application often operate on highly personal and possibly sensitive content, including conversations, geolocation, or physical traits (faces, fingerprints), and typically require tremendous volumes of data for training their underlying ML models.
For example, people in the USA of age 18-24, type on average around 900 words per day (128 messages per day~\cite{messagesPerDay} with an average of 7 words per message~\cite{messageLength}). The Android next-word prediction service is trained on average with sequences of 4.1 words~\cite{hard2018federated} which means that each user generates around 220 training samples daily.
With tens of millions or even billions of user devices~\cite{bonawitz2019towards} scalability issues arise.

\paragraph{Federated Learning.}
To address this dual privacy and scalability challenge, large industrial players are now seeking to exploit the rising power of mobile devices to reduce the demand on their server infrastructures while, at the same time, protecting the privacy of their users. \emph{Federated Learning} (FL) is a new computing paradigm (spearheaded among
others by Google~\cite{konevcny2016federated,smith2017federated,chen2018federated}) where
a central server iteratively trains a global model (used by an ML-based application) without the need to centralize the data.
The iterative training orchestrated by the server consists of the following \emph{synchronous} steps for each update.
Initially, the server selects the contributing mobile devices and sends them the latest version of the model.
Each device then performs a learning task based on its local data and sends the result back to the server. 
The server aggregates a predefined number of results (typically a few hundreds~\cite{bonawitz2019towards}) and finally updates the model.
The server drops any results received after the update.
FL is ``privacy-ready'' and can provide formal privacy guarantees by using standard techniques such as secure aggregation and differential privacy~\cite{bonawitz2017practical}.

The standard use of FL has so far been limited to a few lightweight and extremely privacy-sensitive services, such as next-word prediction~\cite{yang2018applied}, but its popularity is bound to grow. %
Privacy-related scandals continue to unfold \cite{prism,fbca},
and new data protection regulations come into force~\cite{gdpr,ccpa}.
The popularity of FL is clearly visible in two of the most popular ML frameworks (namely TensorFlow and PyTorch)~\cite{tffl,ryffel2018generic}, and also in the rise of startups such as S20.ai~\cite{s20ai} or SNIPS (now part of Sonos)~\cite{snips}, which are betting on private decentralized learning.
\paragraph{Limitation of Standard FL.}
These are encouraging signs, but we argue
in this paper that Standard FL~\cite{bonawitz2019towards} is unfortunately not effective for a large segment of ML-based applications, mainly due to its constraint for 
\emph{high device availability}: the selected mobile devices need to be idle, charging and connected to an unmetered network. 
This constraint removes any impact perceived by users, but also limits the availability of devices for learning tasks. 
Google observed lower prediction accuracy during the day as few devices fulfill this policy and these generally represent a skewed population~\cite{yang2018applied}. 
With most devices available at night the model is generally updated every 24 hours.

\begin{figure}[!t]
\centering
\includegraphics[width=\linewidth,keepaspectratio]{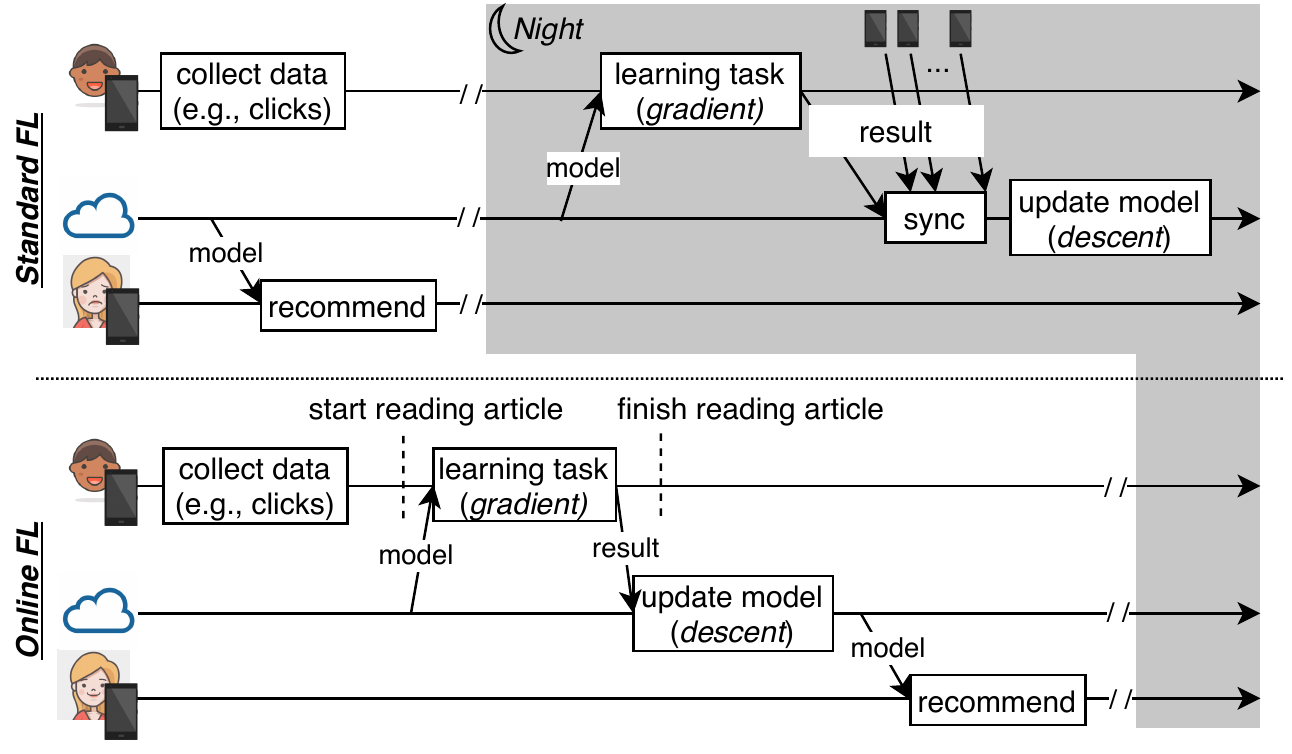}
\caption{Online FL enables frequent updates without requiring idle-charging-WiFi connected mobile devices.
}
\label{fig:onlineFed}
\end{figure}

This constraint may be acceptable for some ML-based services but is problematic to what we call \emph{online learning} systems, which underlie many popular applications such as news recommenders or interactive social networks (e.g., Facebook, Twitter, Linkedin).
These systems involve large amounts of data with high \emph{temporality}, that generally become obsolete in a matter of hours or even minutes~\cite{mishne2013fast}.
To illustrate the limitation of Standard FL, consider two users, Alice and Bob, who belong to a population that trains the ML model underlying a news recommendation system (\cref{fig:onlineFed}).
Bob wakes up earlier than Alice and clicks on some news articles. 
To deliver fresh and relevant recommendations, these clicks should be used to compute recommendations for Alice when she uses the app, slightly after Bob.
In Standard FL (upper half \cref{fig:onlineFed}), the device of Bob would wait until much later (when idle, charging and connected to WiFi) to perform the learning task thus negating the value of the task results for Alice.
In an online learning setup (lower half of \cref{fig:onlineFed}), the activity of Bob is rapidly incorporated into the model, thereby improving the experience of Alice.

\paragraph{Challenges and contributions.}
In this paper we address the aforementioned limitation and enable \emph{Online} FL.
We introduce \system, the first FL system that specifically targets online learning, 
acting as a middleware between the operating system of the mobile device and the ML-based application.
\system addresses two major problems that arise after forfeiting the high device availability constraint.

First, learning tasks may have an energy impact on mobile devices now powered on a battery.
Given that learning tasks are generally compute intensive, they can quickly discharge the device battery and thereby degrade user experience. 
To this end, \system includes \profiler (\cref{sec:profiler}), our new profiling tool which predicts and controls the computation time and the energy consumption of each learning task on mobile devices. 
The goal of \profiler is not trivial given the high heterogeneity of the devices and the performance variability even for the same device over time~\cite{nishio2019client} (as we show in \cref{sec:eval}).

Second, as mentioned above, synchronous training
discards all late results arriving after the model is updated thus wasting the battery of the corresponding devices and their potentially useful data.
Frequent model updates call for small synchronization windows that given the high performance variability, amplify this waste.
We therefore replace the synchronous scheme of Standard FL with asynchronous updates.
However, asynchronous updates introduce
the challenge of \emph{staleness}
as multiple users are now free to perform learning tasks at arbitrary times.
A stale result occurs when the learning task was computed on an outdated model version; meanwhile the global model has progressed to a new version. 
Stale results add noise to the training procedure, slow down or even prevent its convergence~\cite{jiang2017heterogeneity,zhang2015staleness}.
Therefore, \system includes \algo (\cref{sec:ada}), our new  Stochastic Gradient Descent (SGD) algorithm that tolerates staleness by dampening the impact of outdated results.
This dampening depends on (a) the past observed staleness values and (b) the similarity with past learning tasks.

We fully implemented the server side and the Android client of \system\footnote{\url{https://github.com/gdamaskinos/fleet}}.
We evaluate the potential of \system and show that it can increase the accuracy of a recommendation system (that employs Standard FL) by 2.3$\times$ on average, by performing the same number of updates but in a more timely (online) manner. 
Even though the learning tasks drain energy directly from the battery of the phone, they consume on average only 0.036\% of the battery capacity of a modern smartphone per user per day.
We also evaluate the components of \system on 40 commercial Android devices, by using popular benchmarks for image classification.
Regarding \profiler, we show that 90\% of the learning tasks
deviate from a fixed Service Level Objective (SLO) of 3 seconds by at most 0.75 seconds in comparison to 2.7 seconds for the competitor (the profiler of MAUI~\cite{cuervo2010maui}).
The energy deviation from an SLO of 0.075\% battery drop is 0.01\% for \profiler and 0.19\% for the competitor. 
We also show that our staleness-aware learning algorithm (\algo) learns 18.4\% faster than its competitor (\dynsgd~\cite{jiang2017heterogeneity}) on heterogeneous data.

%% file: Framework.tex
\section{{\large \textbf{\system}}} \label{sec:frame}

\system incorporates two components we consider necessary in any system that has the ambition to provide both, the (a) privacy of FL and (b) the precision of online learning systems.
The first component is \profiler, a \emph{lightweight ML-based profiling} mechanism that controls the computation time and energy of the learning task by using ML-based estimators. 
The second component of \system is \algo, a new adaptive learning algorithm that tolerates stale updates by automatically adjusting their weight.

\subsection{Architectural Overview}
\label{sec:netprot}

Similar to the implementation of Standard FL~\cite{bonawitz2019towards},
\system follows a client-server architecture (\cref{fig:arch}) where each user hosts a \emph{worker} and the service provider hosts the \emph{server} (typically in the cloud).
In \system, 
the worker is a library that can be used by any mobile ML-based application (e.g., a news articles application).
The model training protocol of \system is the following (the numbers refer to \cref{fig:arch}):

\begin{figure}
\centering
\includegraphics[width=\linewidth]{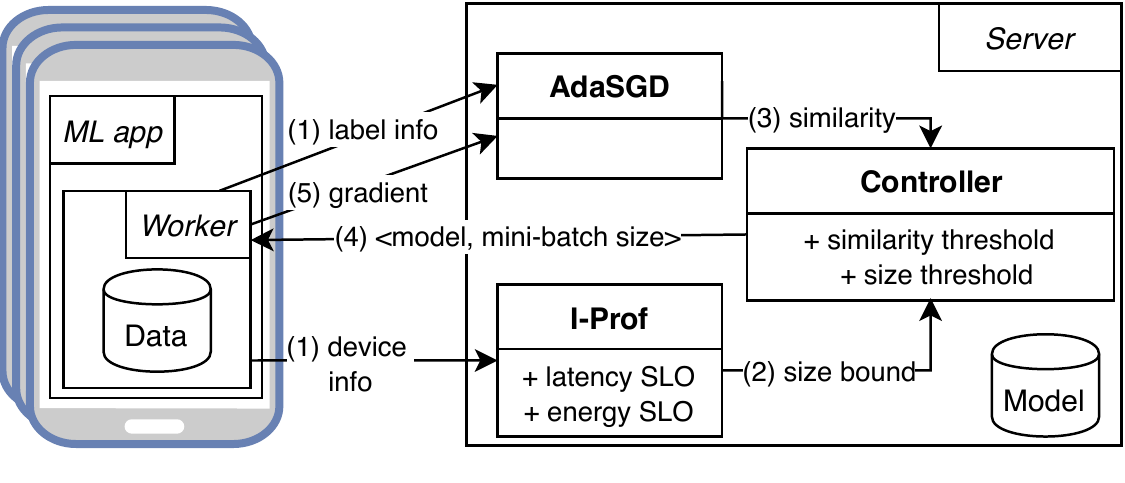}
\caption{The architecture of \system. 
}
\label{fig:arch}
\end{figure}

\begin{myenumerate}[noitemsep,topsep=0pt]
    \item 
    The worker requests a learning task and sends information regarding the labels of the local data along with information about the state of the mobile device.
    We introduce the purpose of this information in Steps~2 and 3.
    \item \profiler employs the device information to bound the workload size (i.e., set a \emph{mini-batch size bound}) that will be allocated to this worker such that the computation time and energy consumption approximate an SLO set by the service provider or negotiated with the user (details in \cref{sec:profiler}).
    The \emph{mini-batch size} is then set as the $min$(mini-batch size bound, local data size\footnote{The size of the local data is available to the server via the label info.}).
    \item \algo computes a \emph{similarity} for the requested learning task with past learning tasks in order to adapt to updates with new data (details in \cref{sec:ada}).
    \item In order to prevent the computation of learning tasks with low or no utility for the learning procedure, the controller checks if both the mini-batch size and the similarity values pass certain \emph{thresholds} set by the service provider.
    If the check fails, the request of the worker is rejected, otherwise the controller sends the model parameters and the mini-batch size to the worker and the learning task execution begins (details about setting these thresholds in \cref{sec:impl}).
    \item Based on the mini-batch size returned by the server, the worker samples from its locally collected data, performs the learning task (i.e., one or multiple local model updates) and sends the result (i.e., the \emph{gradient}) back to the server. 
    On the server side,
    \algo updates the model after dynamically adapting this gradient based on its \emph{staleness} and on its similarity value (details in \cref{sec:ada}).
\end{myenumerate}
The above protocol maintains the key ``privacy-readiness'' of Standard FL: the  user data never leave the device.

\subsection{Workload Bound via Profiling}
\label{sec:profiler}
In Online FL, a mobile device should be able to compute model updates at any time, not only during the night, when the mobile device is idle, charging and connected to WiFi. 
Therefore, \system drops the constraint of Standard FL for high device availability.
Hence, the learning task now drains energy directly from the battery of the device.
Controlling the impact of a learning task on the user application in terms of energy consumption and computation time becomes crucial. 
To this end, \system incorporates a profiling mechanism that determines the workload size (i.e., the mini-batch size) appropriate for each mobile device.

\paragraph{Best-effort solution.}
To highlight the need for a specific profiling tool, 
we first consider a naive solution %
in which users process data points until they reach the SLO either in terms of computation time or energy. At this point, a worker sends back the resulting ``best-effort'' gradient.
The service provider cannot decide beforehand whether for a given device, the cost (in terms of energy, time and bandwidth) to download the model, compute and upload the gradient is worth the benefit to the model.
Updates computed on very small mini-batch sizes (by weak devices) will perturb the convergence of the overall model, and might even negate the benefit of other workers.

\begin{figure}[tb]
\vspace{2mm}
\centering
\includegraphics[width=0.9\linewidth,keepaspectratio]{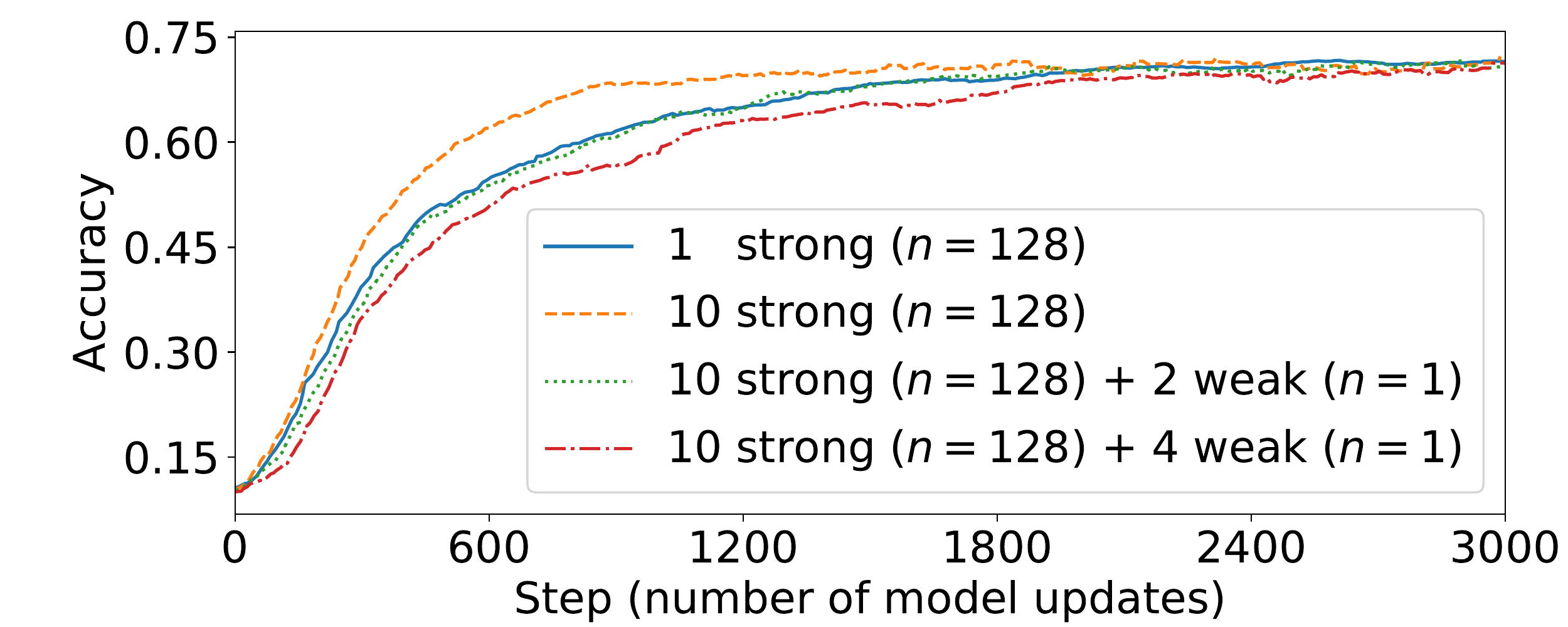}
\caption{Motivation for lower bounding the mini-batch size. The noise introduced by weak workers (i.e., with small mini-batch sizes) may be detrimental to learning.
}
\label{fig:batch_gap}
\end{figure}

To illustrate this point, consider the experiment of \cref{fig:batch_gap}.
The figure charts the result of training a Convolutional Neural Network on CIFAR10~\cite{cifar} under different combinations of ``strong'' and ``weak'' workers. 
The strong workers compute on a mini-batch size of 128 while the weak workers compute on a mini-batch size of 1.
We observe that even 2 weak workers are enough to cancel the benefit of distributed learning, i.e., the performance with 10 strong + 2 weak workers is the same as training with a single strong worker.

One way to avoid this issue could be to drop all the gradients computed on a mini-batch size lower than a given bound or weigh them with a tiny factor according to the size of their underlying mini-batch.
This way would however waste the energy required to obtain these gradients. 
A profiler tool that can estimate the maximum mini-batch size (workload bound) that a worker can compute is necessary for the controller to decide whether to reject the computation request of this worker, before the gradient computation.
Unfortunately, existing profiling approaches~\cite{kwon2013mantis,chowdhury2015system,yoon2012appscope,hao2013estimating,carroll2010analysis,chu2011balancing,cuervo2010maui} are not suitable because they are either relatively inaccurate (see \cref{sec:profilerEval}) or they require privileged access (e.g., rooted Android devices) to low-level system performance counters.

\paragraph{\profiler.}
Mobile devices have a significantly lower level of parallelism in comparison with cloud servers.
For example, the graphical accelerators of mobile devices generally have 10-20 cores~\cite{mali2020, adreno2020} while the GPUs on a server have thousands of cores~\cite{nvidia2020}.
Given this low level of parallelism, even a relatively small mini-batch size can fill the processing pipelines.
Hence, any additional workload will linearly increase the computation time and the energy consumption.
Based on this observation, we built \profiler, a lightweight profiler specifically designed for Online FL systems. 
We design \profiler with three goals in mind: (a) operate effectively with data from a wide range of device types, (b) do so in a lightweight manner, i.e., introduce only a negligible latency to the learning task and (c) rely only on the data available on a stock (non-rooted) Android device.

\profiler employs an ML-based scheme to capture how the device features affect the computation time and energy consumption of the learning task.
\profiler predicts the largest mini-batch size a device can process while respecting both the time and the energy limits set by the SLO. 
To this aim, \profiler uses two predictors, one for computation time and one for energy.
Each predictor updates its state with data from the device information sent by the workers.

\begin{figure}[tb]
\centering 
\subfloat{\includegraphics[width=0.5\linewidth,keepaspectratio]{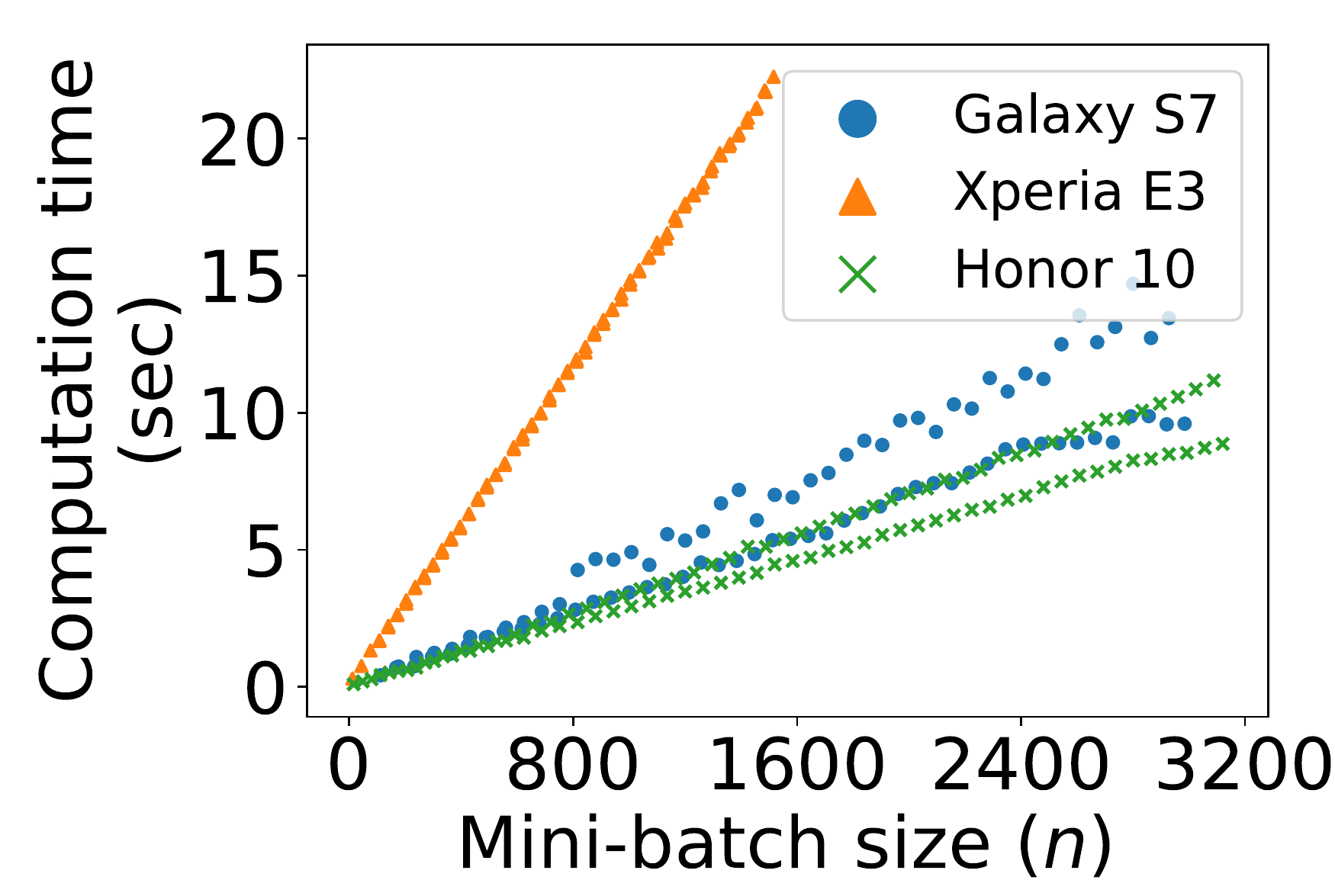}\label{subfig:latencyHypo}}
\subfloat{\includegraphics[width=0.5\linewidth,keepaspectratio]{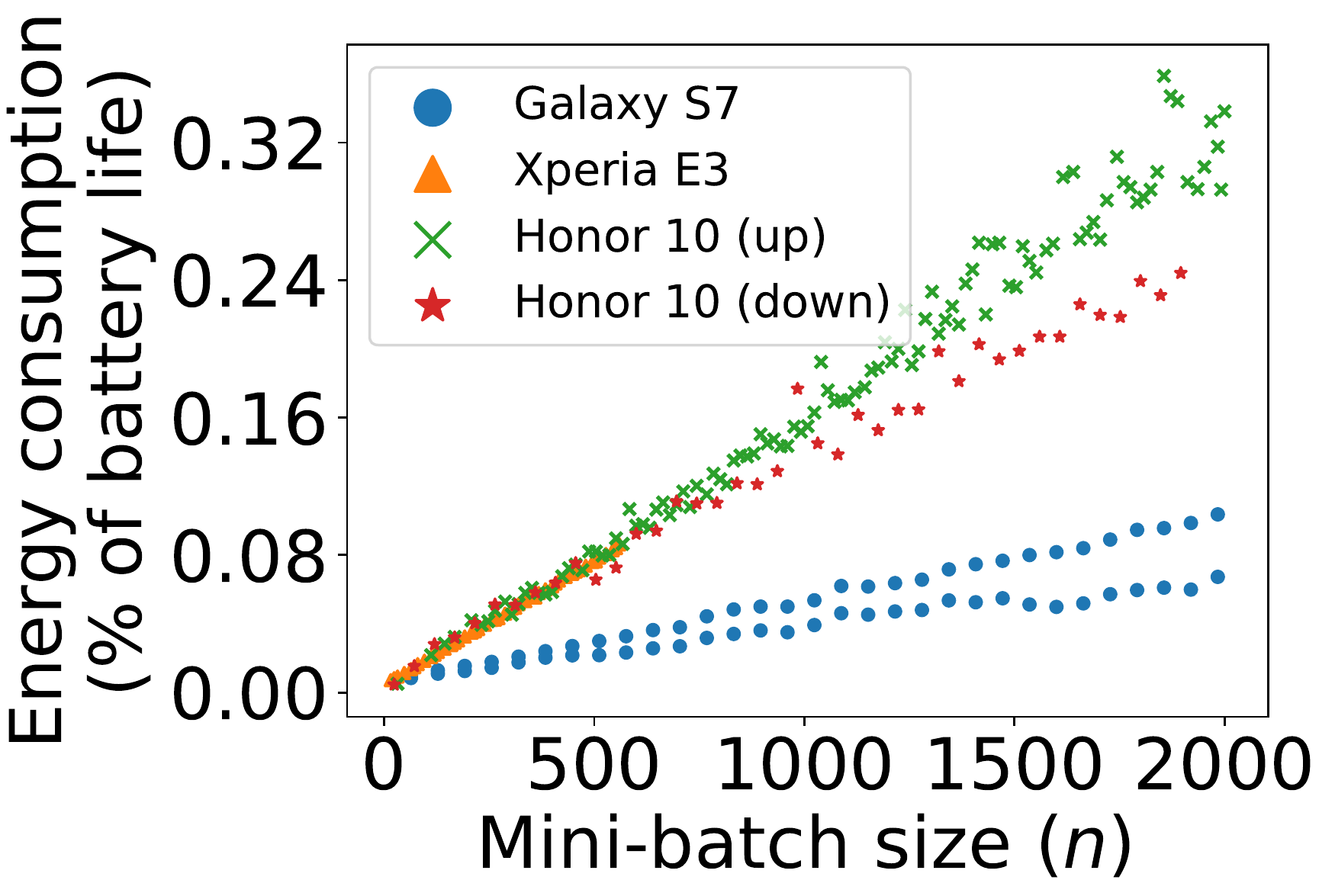}\label{subfig:energyHypo}}
\caption{The linear relation between the computation time and the mini-batch size depends on the specific device, and may even vary for the same device, depending on operation conditions such temperature.}
\label{fig:hypothesis}
\end{figure}

Designing such predictors is however tricky, as modern mobile phones exhibit a wide range of capabilities.
For example, in a matrix multiplication benchmark, Galaxy S6 performs 7.11~Gflops whereas Galaxy S10 performs 51.4~Gflops~\cite{matrixBench}. 
\cref{fig:hypothesis} illustrates this heterogeneity on three different mobile devices by executing successive learning tasks of increasing mini-batch size (``up'').
After reaching the maximum value, we let the devices cool down and execute subsequent learning tasks with decreasing mini-batch size (``down'').
We present the results for the up-down part with the same color-pattern, except for Honor 10 in \cref{subfig:energyHypo} that we split for highlighting the difference.
\cref{fig:hypothesis} illustrates that the linear relation changes for each device and for certain devices (Honor~10, Galaxy S7) also changes with the temperature.
Note that Honor~10 shows an increased variance at the end of the ``up'' part (\cref{subfig:energyHypo}) that is attributed to the high temperature of the device.
The variance is significantly smaller for the ``down'' part.

In the following, we describe how \profiler predicts the mini-batch size ($n$) given a computation time SLO\footnote{The prediction method given an energy SLO is the same.} ($t_{SLO}$).
The computation time linearly increases with the workload size, i.e., $t_{comp} = \alpha \cdot n$, where $\alpha$ depends on the device and its state.
Considering the goal (i.e., $t_{comp} \rightarrow t_{SLO}$), the optimal mini-batch size is predicted as: 
\begin{equation} \label{eq:output}
\hat{n} = \max\left(1, \frac{t_{SLO}}{\hat{\alpha}}\right)
\end{equation}
\profiler estimates the slope $\hat{\alpha}$ from the device characteristics and operational conditions using a method that combines linear regression and online passive-aggressive learning~\cite{crammer2006online}.

The input to this method is a set of device features based on measurements available through the Android API, namely available memory, total memory, temperature and sum of the maximum frequency over all the CPU cores.
However, these features only encode the computing power of a device. 
For the prediction based on the energy SLO, \profiler also needs a feature that encodes the energy efficiency of each device.
We choose this additional feature as the energy consumption per non-idle CPU time\footnote{CPU time spent by processes executing in user or kernel mode.}.
We show in our evaluation (\cref{sec:profilerEval}) that these features achieve our three design goals.
Given a vector of device features ($\xv$), and a vector of model parameters ($\thetav_{\text{prof}}$), the slope $\hat{\alpha}$ is estimated as $\hat{\alpha} = \xv^T\thetav_{\text{prof}}$.

\profiler uses a cold-start linear regression model for the first request of each user device.
We pre-train the cold-start model using ordinary least squares with an offline dataset.
This dataset consists of data collected by executing requests from a set of training devices with a mini-batch size increasing from 1 till a value such that the computation time reaches twice the SLO.
\profiler periodically re-trains the cold-start model after appending new data (device features).
The training cost is negligible given the small number of features.

Furthermore, \profiler creates a personalized model for every new device model (e.g., Galaxy S7) and employs it for every following request coming from this particular model.
\profiler bootstraps the new model with the first request (for which the cold-start model is used to estimate the computation time).
For all the following learning tasks that result in pairs of ($\xv^{(k)}, \alpha^{(k)}$),
\profiler incrementally updates a Passive-Aggressive (PA) model~\cite{crammer2006online} as: ${\thetav_{\text{prof}}^{(k+1)} = \thetav_{\text{prof}}^{(k)} + \frac{f^{(k)}}{\norm{\bm{x}^{(k)}}^2} \bm{v}^{(k)}}$
where 
${\bm{v}^{(k)} = sign(\alpha^{(k)} - \xv^{(k)\,T}\thetav^{(k)}_{\text{prof}}) \xv^{(k)}}$ 
denotes the update direction, and $f$ the loss function:
\begin{equation} \label{eq:paloss}
f(\thetav_{\text{prof}}, \bm{x}, \alpha) =  
\begin{cases}
0 & \text{if } |\xv^T\thetav_{\text{prof}} - \alpha| \leq \epsilon \\
|\xv^T\thetav_{\text{prof}} - \alpha| - \epsilon & \text{otherwise.}
\end{cases}
\end{equation}
The parameter $\epsilon$ controls the sensitivity to prediction error and thereby the aggressiveness of the regression algorithm, i.e., the smaller the value of $\epsilon$ 
the larger the update for each new data instance (more aggressive).

\profiler focuses solely on the time and energy spent during an SGD computation.
Despite network costs (in particular when transferring models) having also an important impact, they fall outside the scope of this work as one can rely on prior work~\cite{altamimi2015energy, liu2015empirical, qian2011profiling} to estimate the time and energy of network transfers within \system.

\subsection{Adaptive Stochastic Gradient Descent} 
\label{sec:ada}
The server-driven synchronous training of Standard FL is not suitable for Online FL, as the latter requires frequent updates and needs to exploit contributions from all workers, including slow ones (\cref{sec:intro}).
Therefore, we introduce \algo, an asynchronous learning algorithm that is robust to stale updates. 
\algo is responsible for aggregating the gradients sent by the workers and updating the application model ($\thetav_{\text{app}}$)\footnote{Not to be confused with the model of the profiler ($\theta_{\mathrm{prof}}$).}.
Each update takes place after \algo receives $K$ gradients. 
The aggregation parameter $K$ can be either fixed or based on a time window (e.g., update the model every 1 hour).
The model update is:
\begin{equation} \label{eq:sgdUpdate}
\thetav_{\text{app}}^{(t+1)} = \thetav_{\text{app}}^{(t)} - \gamma_t \sum_{i=1}^K \min\left(1, \Lambda(\tau_i) \cdot \frac{1}{sim(\xv_i)}\right) \cdot \bm{G}(\thetav_{\text{app}}^{(t_i)}, \xi_i) 
\end{equation}
where $\gamma_t$ is the learning rate, $t \in \mathbb{N}$ denotes the global logical clock (or step) of the model at the server
(i.e., the number of past model updates) and $t_i \leq t$ denotes the logical clock of the model that the worker receives.
$\bm{G}(\thetav_{\text{app}}^{(t_i)}, \xi_i)$ is the gradient computed by the client w.r.t the model $\thetav_{\text{app}}^{(t_i)}$ on the mini-batch $\xi_i$ drawn uniformly from the local dataset $\xv_i$.

The workers send gradients asynchronously that can result in \emph{stale} updates.
The \emph{staleness} of the gradient (${\tau_i := t - t_i}$) shows the number of model updates between the model pull and gradient push of worker $i$. 
One option is to directly apply this gradient, at the risk of slowing down or even completely preventing convergence~\cite{jiang2017heterogeneity, zhang2015staleness}. 
The Standard FL algorithm (FedAvg~\cite{mcmahan2017communication}) simply drops stale gradients. 
However, even if computed on a stale model, the gradient may incorporate potentially valuable information.
Moreover, in \system, the gradient computation may drain energy directly from the battery of the phone, thus making the result even more valuable.
Therefore, \algo utilizes even stale gradients without jeopardizing the learning process, by multiplying each gradient with an additional weight to the learning rate.
This weight consists of (a) a dampening factor based on the staleness ($\Lambda(\tau_i)$) and (b) a boosting factor based on the user's data novelty ($\frac{1}{sim(\xv_i)}$), that we describe in the following.

\begin{figure}
\centering
\includegraphics[width=\linewidth,keepaspectratio]{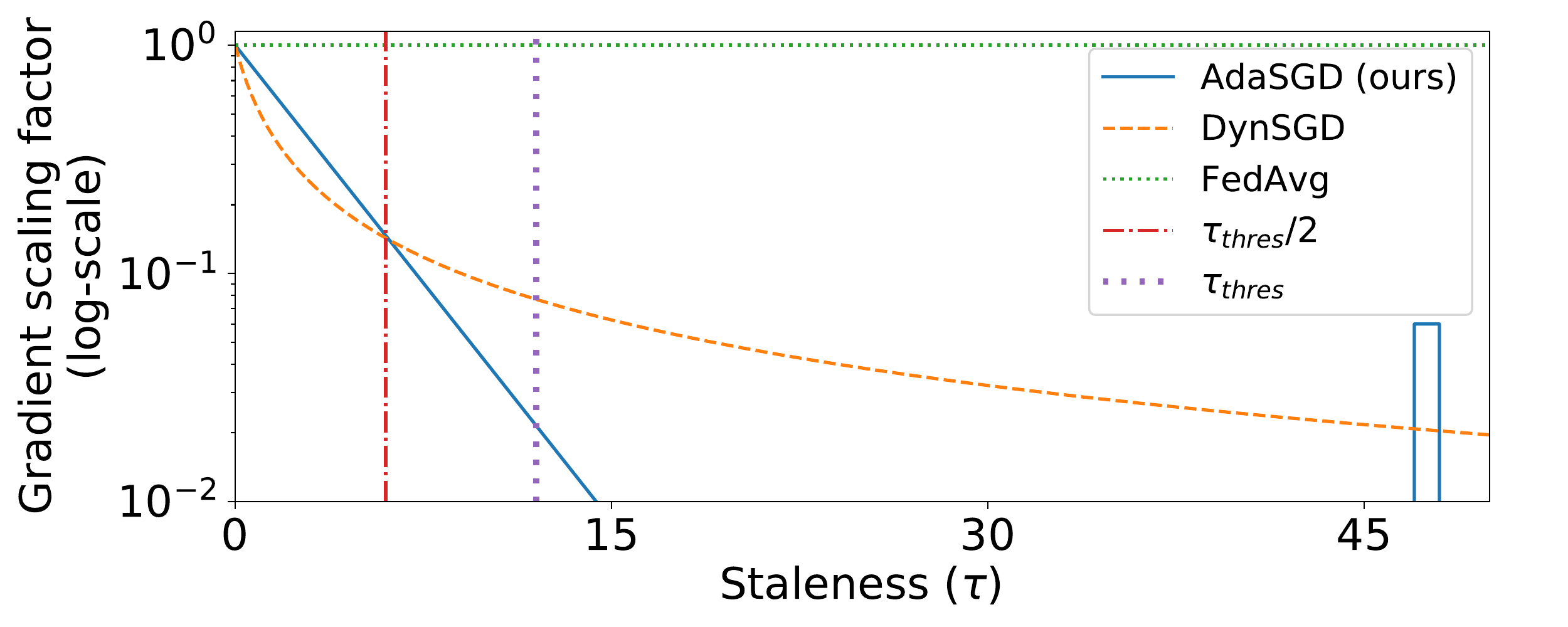}
\caption{Gradient scaling schemes of SGD algorithms. \algo, proposed in this paper, dampens stale gradients with an exponentially decreasing function ($\Lambda(\tau)$) based on the expected percentage of non-stragglers ($\tau_{\text{thres}}:= s$-th percentile of staleness values), and boosts the gradient of the straggler ($\tau=48$) due to its low similarity ($sim(\xv_i)$).}
\label{fig:scaling_factor}
\end{figure}

\paragraph{Staleness-based dampening.}
\algo builds on prior work on staleness-aware learning that has shown promising results~\cite{jiang2017heterogeneity, zhang2015staleness}.
In order to accelerate learning, \algo relies on a system parameter: \emph{the expected percentage of non-stragglers} (denoted by $s \%$).
We highlight that this value is not a hyperparameter that needs tuning for each ML application but a system parameter that solely depends on the computing and networking characteristics of the workers, while it can be adapted dynamically~\cite{ouyang2016straggler,phan2019new}.
We define the staleness-aware dampening factor $\Lambda(\tau) = e^{-\beta \tau}$, with $\beta$ chosen s.t. ${\frac{1}{\frac{\tau_{\text{thres}}}{2}+1} = e^{-\beta \frac{\tau_{\text{thres}}}{2}}}$ 
(i.e., the inverse dampening function~\cite{jiang2017heterogeneity} intersects with our exponential dampening function in $\frac{\tau_{\text{thres}}}{2}$),
where $\tau_{\text{thres}}$ is the $s$-th percentile of past staleness values.
\cref{fig:scaling_factor} shows the dampening factor of \algo compared to the inverse dampening function (employed by \dynsgd~\cite{jiang2017heterogeneity}).
Our hypothesis is that the perturbation to the learning process introduced by stale gradients, increases exponentially and not linearly with the staleness. 
We empirically verify the superior performance of our exponential dampening function compared to the inverse in \cref{sec:algoEval}.

As a quantile, $\tau_{\text{thres}}$ is estimated from the staleness distribution.
In practice, for the past staleness values to be representative of the actual distribution, an initial bootstrapping phase can employ the dampening factor of \dynsgd.
After this phase, the service provider can set $s\%$ and deploy \algo.
An underestimate of $s\%$ will slow down convergence, whereas an overestimate may lead to divergence.
As we empirically observe (\cref{sec:onlinefl}), the staleness distribution often has a long tail.
In such cases, the best choice of $s\%$ is the one that sets $\tau_{\text{thres}}$ at the beginning of the tail.

\paragraph{Similarity-based boosting.}
In the presence of stragglers with large delays (comparing to the mean latency), staleness can grow and drive $\Lambda(\tau)$ close to 0, i.e., almost neglect the gradients of these stragglers.
Nevertheless, these gradients may contain valuable information.
In particular, they may be computed on data that are not similar to the data used by past gradients.
Hence, \algo boosts these gradients by using the following similarity value:
\begin{equation} \label{eq:similarity}
sim(\xv_i) =  BC(\bm{LD}(\xv_i), \bm{LD}_{global})
\end{equation}
where $BC$ denotes the Bhattacharyya coefficient~\cite{bhattacharyya}, and $\bm{LD}$ the label distribution, that captures the importance of each gradient.
We choose this coefficient given our constraints ($sim(\xv_i) \in [0, 1]$).
For instance, given an application with 4 distinct labels and a local dataset ($\xv_i$) that has 1 example with label 0, and 2 examples with label 1: ${\bm{LD}(\xv_i) = [\frac{1}{3}, \frac{2}{3}, 0, 0]}$.
The global label distribution ($\bm{LD}_{global}$) is computed on the aggregate number of previously used samples for each label.
We highlight that $\bm{LD}$ is not specific to classification ML tasks; for regression tasks, $\bm{LD}$ would involve a histogram, with the length of the $\bm{LD}$ vector being equal to the number of bins instead of the number of classes.

The similarity value essentially captures how valuable the information of the gradient is.
For instance, if a gradient is computed on examples of an unseen label (e.g., a very rare animal), then its similarity value is less than 1 (i.e., has information not similar to the current knowledge of the model).
For the similarity computation, the server needs only the indices of the labels of the local datasets without any semantic information (e.g., label 3 corresponds to ``dogs'').

\subsection{Implementation} 
\label{sec:impl}

The server of \system is implemented as a web application (deployed on an HTTP server) and the worker as an Android library. 
The server transfers data with the workers via Java streams by using Kryo~\cite{kryo} and Gzip.
In total, \system accounts 26913 Java LoC, 3247 C/C++ LoC and 1222 Python LoC.

\paragraph{Worker runtime.}
We design the worker of our middleware (\system) as a library and execute it only when the overlying ML application (\cref{fig:arch}) is running in the foreground for two main reasons.
First, since Android is a UI-interactive operating system, background applications have low priority so their access to system resources is heavily restricted and they are likely to be killed by the operating system to free resources for the foreground running app. 
Therefore, allowing the worker to run in the background would make its performance very unpredictable and thus impact the predictions of \profiler.
Second, running the worker in the foreground alleviates the impact of collocated (background) workload.

We build our main library for Convolutional Neural Networks in C++ on top of \system.
We employ (i) the Java Native Interface (JNI) for the server, (ii) the Android NDK for the worker, (iii) an encoding scheme for transferring C++ objects through the java streams, and (iv) a thread-based parallelization scheme for the independent gradient computations of the worker.
On recent mobile devices that support NEON~\cite{neon}, \system accelerates the gradient computations by using SIMD instructions.
We also port a popular deep learning library (DL4J~\cite{dl4j}) to \system, to benefit from its rich ecosystem of ML algorithms.
However, as DL4J is implemented in Java, we do not have full control over the resource allocation.

\system relies on the developer of the overlying ML application to ensure the performance isolation between the running application and the worker runtime.
The worker can execute in a window of low user activity (e.g., while the user is reading an article) to minimize the impact of the overlying ML application on the predictive power of \profiler.

\paragraph{Resource allocation.}
Allocating system resources is a very challenging task given the latency and energy constraints of mobile devices~\cite{mishra2018caloree,ding2019}.
Our choice of employing only stock Android without root access means we can only control which cores execute the workload on the worker,
with no access, for instance, to low-level advanced tuning.
Given this limited control and the inherent mobile device heterogeneity, we opt for a simple yet effective scheme for allocating resources.

This scheme schedules the execution only on the ``big'' cores for ARM big.LITTLE architectures and on all the cores otherwise.
In the case of computationally intensive tasks (such as the learning tasks of \system), big cores are more energy efficient than LITTLE cores because they finish the computation much faster~\cite{greenhalgh2013big}. 
Regarding ARMv7 symmetric architectures with 2 and 4 cores that equip older mobile devices,
the energy consumption per workload is constant regardless of the number of cores: a higher level of parallelism will consume more energy but the workload will execute faster.
For this reason, our allocation policy relies on all the available cores so that we can take advantage of the embarrassingly parallel nature of the gradient computation tasks.
For such tasks, we empirically show (\cref{sec:evalResourceAlloc}) that this scheme outperforms more complex alternatives~\cite{mishra2018caloree}.

\paragraph{Controller thresholds.} 
In practice, the service provider can adopt various approaches to define the size and similarity thresholds of the controller (\cref{fig:arch}).
One option is A/B testing along with the gradual increase of the thresholds.
In particular, the system initializes the thresholds to zero and divides the users into two groups. 
The first group tests the impact of the mini-batch size and the second the impact of the label similarity. 
Both groups gradually increase the thresholds until the impact on the service quality is considered acceptable.
The server can execute this A/B testing procedure periodically, i.e., reset the thresholds after a time interval.
We empirically evaluate the impact of these thresholds on prediction quality in \cref{sec:evalController}.

%% file: Evaluation.tex
\section{Evaluation} 
\label{sec:eval}

Our evaluation consists of two main parts.
First, in \cref{sec:onlinefl}, we evaluate the claim that Online FL holds the potential to deliver better ML performance than Standard FL~\cite{bonawitz2019towards} for applications that employ data with high temporality (\cref{sec:intro}).
Second, we evaluate in more detail the internal mechanisms of \system, namely \algo (\cref{sec:algoEval}), \profiler (\cref{sec:profilerEval}), the resource allocation scheme (\cref{sec:evalResourceAlloc}) and the controller (\cref{sec:evalController}).

We deploy the server of \system on a machine with an Intel Xeon X3440 with four CPU cores, 16~GiB RAM and 1~Gb Ethernet, on Grid5000~\cite{g5k}.
The workers are deployed on a total of 40 different mobile phones that we either personally own or belong to the AWS Device Farm~\cite{deviceFarm} (Oregon, USA).
In \cref{sec:onlinefl}, we deploy the worker on a Raspberry Pi 4 as our hashtag recommender is implemented on TensorFlow that does not yet support training on Android devices.

\subsection{Online VS Standard Federated Learning}
\label{sec:onlinefl}

We compare Online with Standard FL on a Twitter hashtag recommender.
Tweepy~\cite{tweepy} enables us to collect around 2.6 million tweets spanning across 13 successive days and located in the west coast of the USA.
We preprocess these tweets (e.g., remove automatically generated tweets, remove special symbols) based on~\cite{dhingra2016tweet2vec}.
We then divide the data into shards, each spanning 2 days, and divide each shard into chunks of 1 hour.
We finally group the data into mini-batches based on the user id.

Our training and evaluation procedure follows an Online FL setup.
Our model is a basic Recurrent Neural Network implemented on TensorFlow with 123,330 parameters~\cite{tfTextClassification}, that predicts the $5$ hashtags with the largest values on the output layer.
The model training consists of successive gradient-descent operations, with each gradient derived from a single mini-batch (i.e., sent by a single user).
Every day includes 24 mini-batches and thus 24 distinct gradient computations, while each model update uses 1 gradient.
For the Online FL setup, %
the 24 daily updates are evenly distributed (one update every hour).
Training uses the data of the previous hour and testing uses the data of the next hour.
For the Standard FL setup, %
the 24 daily updates occur all at once (at the end of the day).
Training uses the data of the previous day and testing uses the data of the next day.
We highlight that under this setup, the two approaches employ the same number of gradient computations and the difference lies only in the time they perform the model updates.
We also compare against a baseline model that always predicts the most popular hashtags~\cite{kowald2017temporal,otsuka2014design}.
We evaluate the model on the data of each chunk and reset the model at the end of each shard.

\begin{figure}
\centering
\includegraphics[width=\linewidth,keepaspectratio]{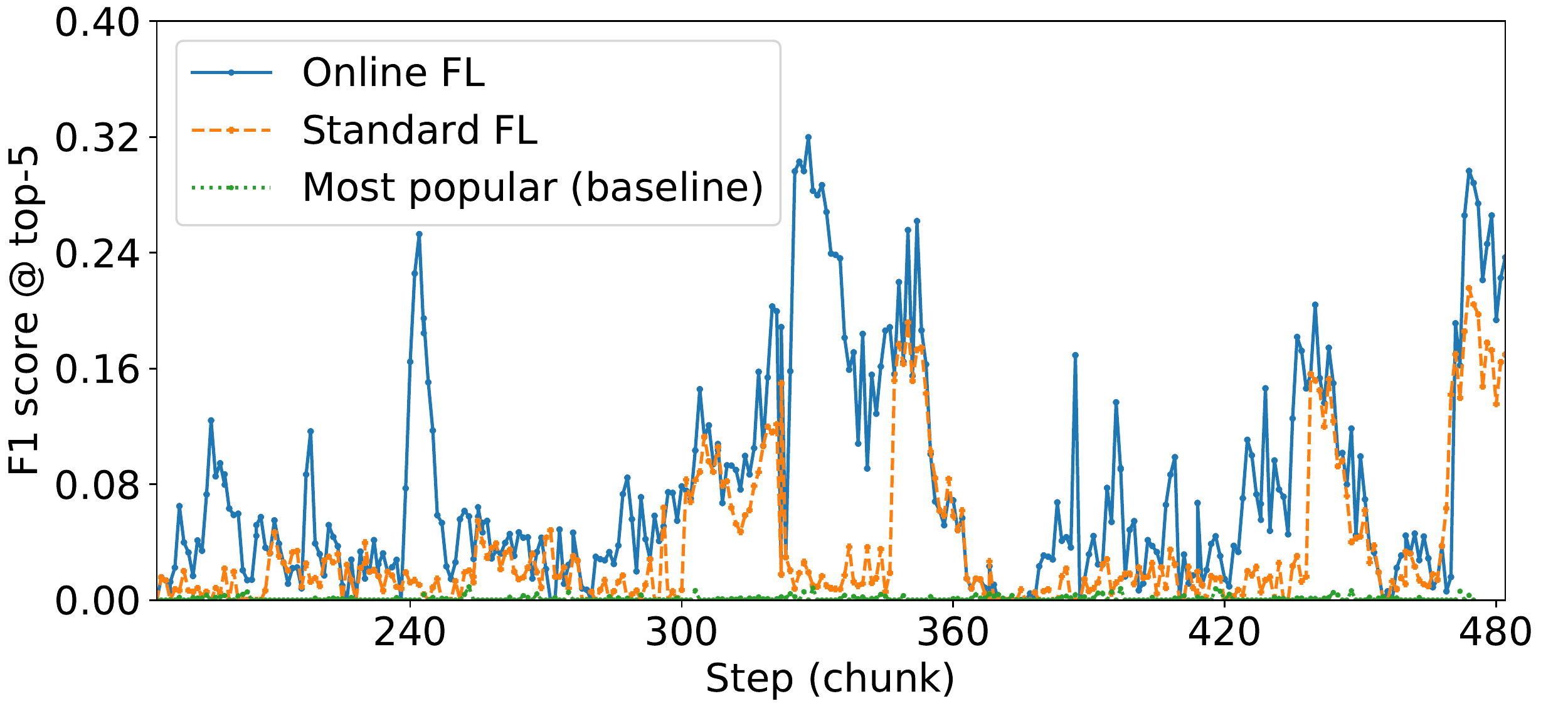}
\caption{Online FL boosts Twitter hashtag recommendations by an average of 2.3$\times$ comparing to Standard FL.
}
\label{fig:hashtagAcc}
\end{figure}

\paragraph{Quality boost.}
For assessing the quality of the hashtag recommender, we employ the F1-score @ top-5~\cite{kowald2017temporal,gong2016hashtag} to capture how many recommendations were used as hashtags (precision) and how many of the used hashtags were recommended (recall).
In particular, for each tweet in the evaluation set, we compare the output of the recommender (top-5 hashtags) with the actual hashtags of the tweet, and derive the F1-score.
\cref{fig:hashtagAcc} shows that Online FL outperforms Standard FL in terms of F1-score, with an average boost of 2.3$\times$.
Online FL updates the model in a more timely manner, i.e., soon after the data generation time, and can thus better predict (higher F1-score) the new hashtags than Standard FL.
The performance of the baseline model is quite low as the nature of the data is highly temporal~\cite{kywe2012recommending}.

\paragraph{Energy impact.}
We measure the energy impact of the gradient computation on the Raspberry Pi worker.
The Raspberry Pi has no screen; nevertheless recent trends in mobile/embedded processor design show that the processor is dominating the energy consumption, especially for compute intensive workloads such as the gradient computation~\cite{halpern2016mobile}.
We measure the power consumption of every update of Online FL by executing the corresponding gradient computation 10 times and by taking the median energy consumption.
We observe that the power depends on the batch size and increases from 1.9 Watts (idle) to 2.1 Watts (batch size of 1) and to 2.3 Watts (batch size of 100).
The computation latency is 5.6 seconds for batch size of 1 and 8.4 for batch size of 100.
Across all the updates of Online FL (that employ various batch sizes and result in the quality boost shown in \cref{fig:hashtagAcc}), we measure the average, median, $99^{\text{th}}$ percentile and maximum values of the daily energy consumption as 4, 3.3, 13.4 and 44 mWh respectively.
Given that most modern smartphones have battery capacities over 11000~mWh, we argue that Online FL imposes a minor energy consumption overhead for boosting the prediction quality.

\begin{figure}
\centering
\subfloat[]{\label{subfig:gaussHashtag} \includegraphics[width=0.5\linewidth,keepaspectratio]{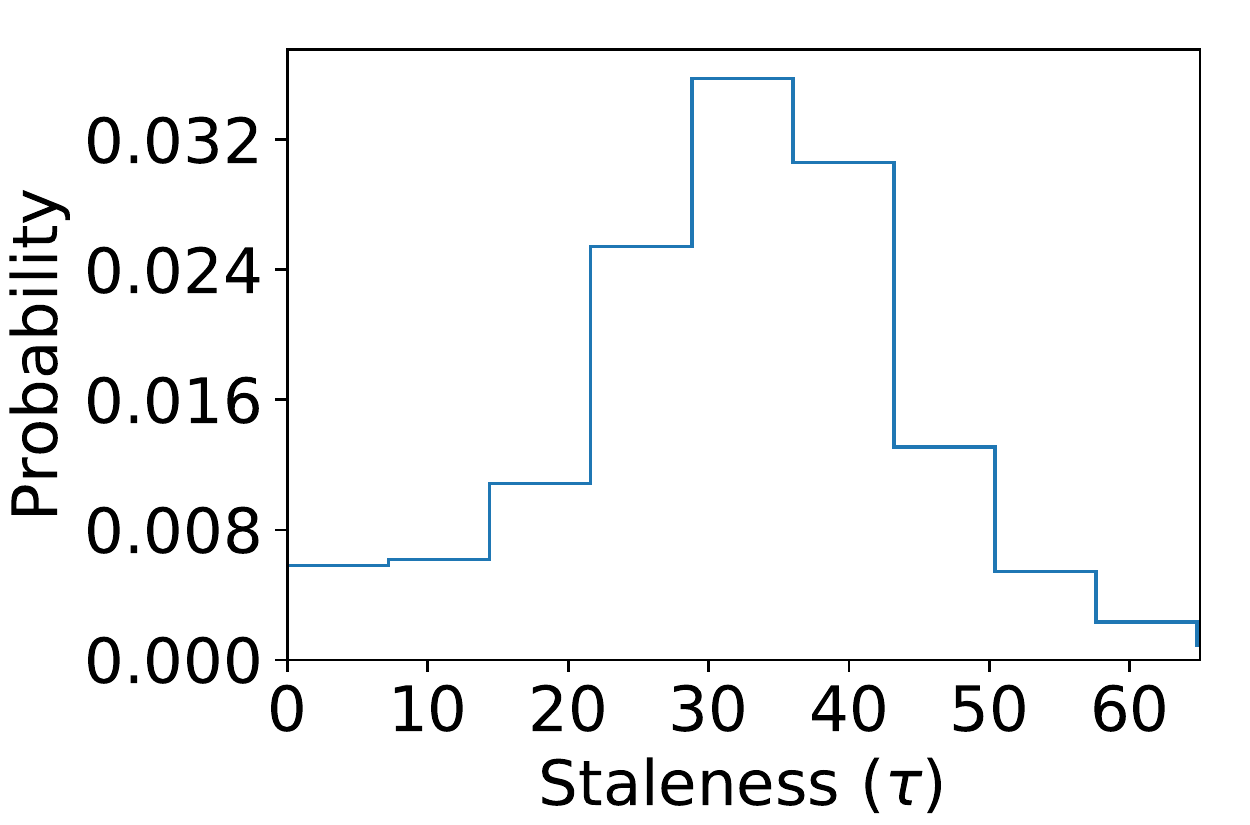}}
\subfloat[]{\label{subfig:tailHashtag} \includegraphics[width=0.5\linewidth,keepaspectratio]{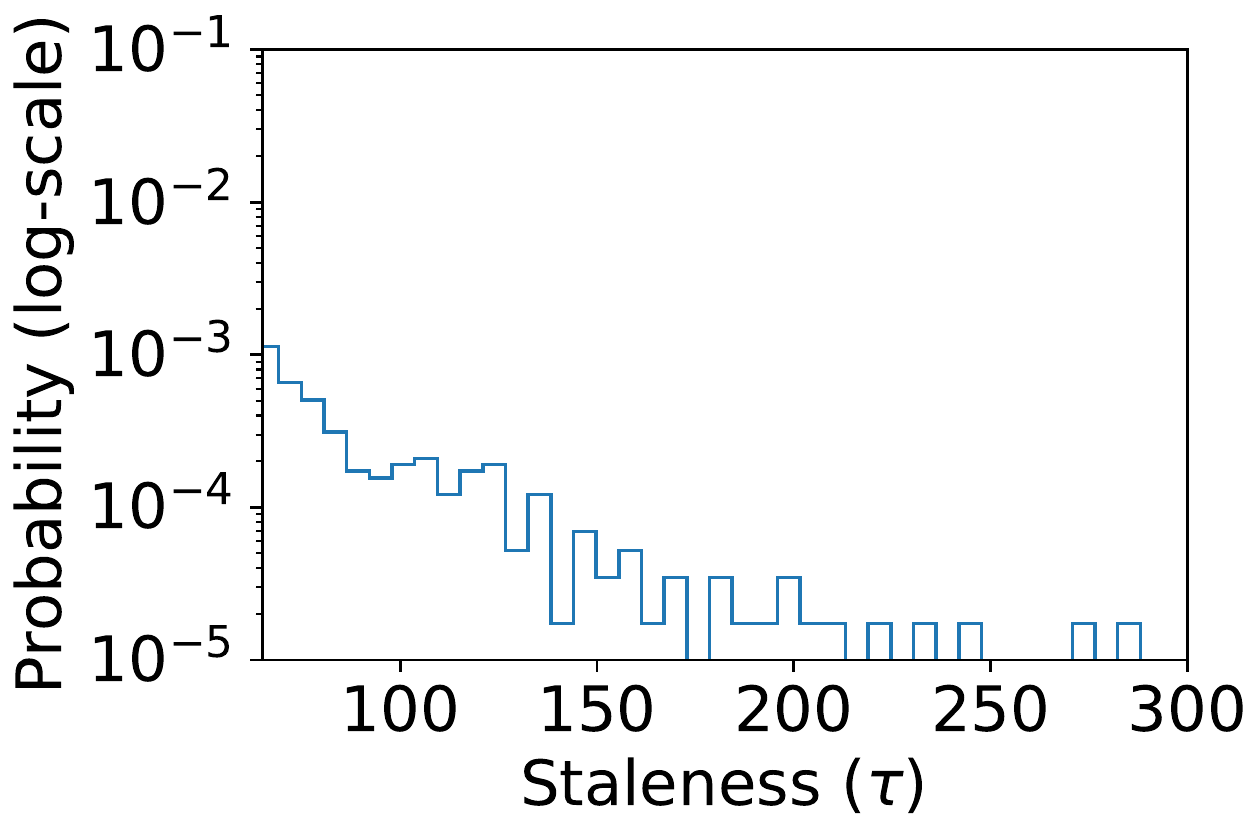}}
\caption{Staleness distribution of collected tweets follows a Gaussian distribution ($\tau < 65$) with a long tail ($\tau > 65$).
}
\label{fig:hashtagStaledist}
\end{figure}

\paragraph{Staleness distribution.}
We study the staleness distribution of the updates on our collected tweets, in order to set our experimental setup for evaluating \algo (\cref{sec:algoEval}). 
We assume that the round-trip latency per model update (gradient computation time plus network latency) follows an exponential distribution (as commonly done in the distributed learning literature~\cite{mitliagkas2016asynchrony,dutta2016short,lee2017speeding,al2020gradient}).
The network latency for downloading the model (123,330 parameters) 
and uploading the gradients is estimated to 1.1 second for 4G LTE and 3.8 seconds for 3G HSPA+~\cite{4gspeed}.
We then estimate the average computation latency to be 6 seconds, based on our latency measurements on the Raspberry Pi.
Therefore, we choose the exponential distribution with a minimum of $6+1.1=7.1$ seconds and a mean of $\frac{(6+1.1)+(6+3.8)}{2}=8.45$ seconds.
Given the exponential distribution for the round-trip latency and the timestamps of the tweets, 
we observe (in \cref{fig:hashtagStaledist}) that the staleness follows a Gaussian distribution with a long tail (as assumed in~\cite{zhang2015staleness}).
The long tail is due the presence of certain peak times with hundreds of tweets per second.

Noteworthy, for applications such as our Twitter hashtag recommender where the user activity (and therefore the per-user data creation) is concentrated in time, the difference between Online and Standard FL in terms of communication overhead (due to the gradient-model exchange) is negligible. 
For applications where the per-user data creation is spread across the day, the communication overhead of Online FL grows, as each user communicates more often with the server.

\subsection{\algo Performance} \label{sec:algoEval}
 
We now dissect the performance of \algo via an image classification application that involves Convolutional Neural Networks (CNNs).
We choose this benchmark due to its popularity for the evaluation of SGD-based approaches~\cite{zhang2015staleness,abadi2016deep,chilimbi2014project,zhao2018federated,mcmahan2017communication,kang2017neurosurgeon}.
We employ multiple scenarios involving various staleness distributions, data distributions, and a noise-based differentially private mechanism.

\paragraph{Image classification setup.}
We implement the models shown in \cref{table:cnn-params} in \system\footnote{We implement the CNN for E-MNIST on DL4J and the rest on our default CNN library.} to classify handwritten characters and colored images.
We use three publicly available datasets:
MNIST~\cite{mnist},
E-MNIST~\cite{cohen2017emnist} and CIFAR-100~\cite{cifar}.
MNIST consists of 70,000 examples of handwritten digits (10 classes) while
E-MNIST consists of 814,255 examples of handwritten characters and digits (62 classes). CIFAR-100 consists of 60,000 colour images in 100 classes, with 600 images per class.
We perform min-max scaling as a pre-processing step for the input features.

We split each dataset into \emph{training} / \emph{test sets}: 60,000 / 10,000 for MNIST, 697,932 / 116,323 for E-MNIST and 50,000 / 10,000 for CIFAR-100.
Unless stated otherwise, we set the aggregation parameter $K$ (\cref{sec:ada}) to 1 (for maximum update frequency), the mini-batch size to 100 examples~\cite{neyshabur2015path}, the $\epsilon$ (the Passive-Aggressive parameter) to 0.1 and the learning rate to $15*10^{-4}$ for CIFAR-100, $8*10^{-4}$ for E-MNIST, and $5*10^{-4}$ for MNIST. 

Since the training data present on mobile devices are typically collected by the users based on their local environment and usage, both the size and the distribution of the training data will typically heavily vary among users. 
Given the terminology of statistics, this means that the data are not Independent and Identically Distributed (\emph{non-IID}).
Following recent work on FL~\cite{wang2019beyond,yurochkin2019bayesian,wang2019adaptive,zhao2018federated}, we employ a non-IID version of MNIST. 
Based on the standard data decentralization scheme~\cite{mcmahan2017communication}, we sort the data by the label, divide them
into shards of size equal to $\frac{60000}{\text{2 * number of users}}$, and assign 2 shards to each user. Therefore, each user will contain examples for only a few labels.

\begin{table}%
\centering
\caption{CNN parameters.}
\scalebox{0.75}{%
\setlength\tabcolsep{1.5pt}
\begin{tabular}{|l|l|l|l|l|l|l|l|l|l|}
\hline
Dataset   & Parameters                                                        & Input   & Conv1                                                & Pool1                                             & Conv2                                                & Pool2                                             & FC1 & FC2 & FC3 \\ \hline
MNIST    & \begin{tabular}[c]{@{}l@{}}Kernel size\\ Strides\end{tabular} & 28$\times$28$\times$1 & \begin{tabular}[c]{@{}l@{}}5$\times$5$\times$8\\ 1$\times$1\end{tabular} & \begin{tabular}[c]{@{}l@{}}3$\times$3\\ 3$\times$3\end{tabular} & \begin{tabular}[c]{@{}l@{}}5$\times$5$\times$48\\ 1$\times$1\end{tabular} & \begin{tabular}[c]{@{}l@{}}2$\times$2\\ 2$\times$2\end{tabular} & 10  & --  & --  \\ \hline
E-MNIST    & \begin{tabular}[c]{@{}l@{}}Kernel size\\ Strides\end{tabular} & 28$\times$28$\times$1 & \begin{tabular}[c]{@{}l@{}}5$\times$5$\times$10\\ 1$\times$1\end{tabular} & \begin{tabular}[c]{@{}l@{}}2$\times$2\\ 2$\times$2\end{tabular} & \begin{tabular}[c]{@{}l@{}}5$\times$5$\times$10\\ 1$\times$1\end{tabular} & \begin{tabular}[c]{@{}l@{}}2$\times$2\\ 2$\times$2\end{tabular} & 15  & 62  & --  \\ \hline
CIFAR-100 & \begin{tabular}[c]{@{}l@{}}Kernel size\\ Strides\end{tabular} & 32$\times$32$\times$3 & \begin{tabular}[c]{@{}l@{}}3$\times$3$\times$16\\ 1$\times$1\end{tabular} & \begin{tabular}[c]{@{}l@{}}3$\times$3\\ 2$\times$2\end{tabular} & \begin{tabular}[c]{@{}l@{}}3$\times$3$\times$64\\ 1$\times$1\end{tabular} & \begin{tabular}[c]{@{}l@{}}4$\times$4\\ 4$\times$4\end{tabular} & 384 & 192 & 100 \\ \hline
\end{tabular}
}
\label{table:cnn-params}
\end{table}

\paragraph{Staleness awareness setup.} 
To be able to precisely compare \algo with its competitors, we control the staleness of the updates produced by the workers of \system.
Based on~\cite{zhang2015staleness} and the shape of the staleness distribution shown in \cref{fig:hashtagStaledist}, we employ Gaussian distributions for the staleness with two setups: $D1 := \cN (\mu=6, \sigma=2)$ and $D2 := \cN (\mu=12, \sigma=4)$, to measure the impact of increasing the staleness.
We set the expected percentage of non-stragglers ($s\%$) to 99.7\%, i.e., $\tau_\text{thres} = \mu + 3 \sigma$.
We evaluate the SGD algorithms on \system by using commercial Android devices from AWS.

We evaluate the performance of \algo against three learning algorithms: (i) \dynsgd~\cite{jiang2017heterogeneity}, a staleness-aware SGD algorithm employing an inverse dampening function (${\Lambda(\tau)=\frac{1}{\tau+1}}$), that \algo builds upon (\cref{sec:ada}), (ii) the standard SGD algorithm with synchronous updates (SSGD) that represents the ideal (\emph{staleness-free}) convergence behaviour, 
and (iii) \baselinesgd~\cite{mcmahan2017communication}, the standard \emph{staleness-unaware} SGD algorithm that is based on gradient averaging.

\begin{figure}[!ht]
\centering 
\includegraphics[width=\linewidth,keepaspectratio]{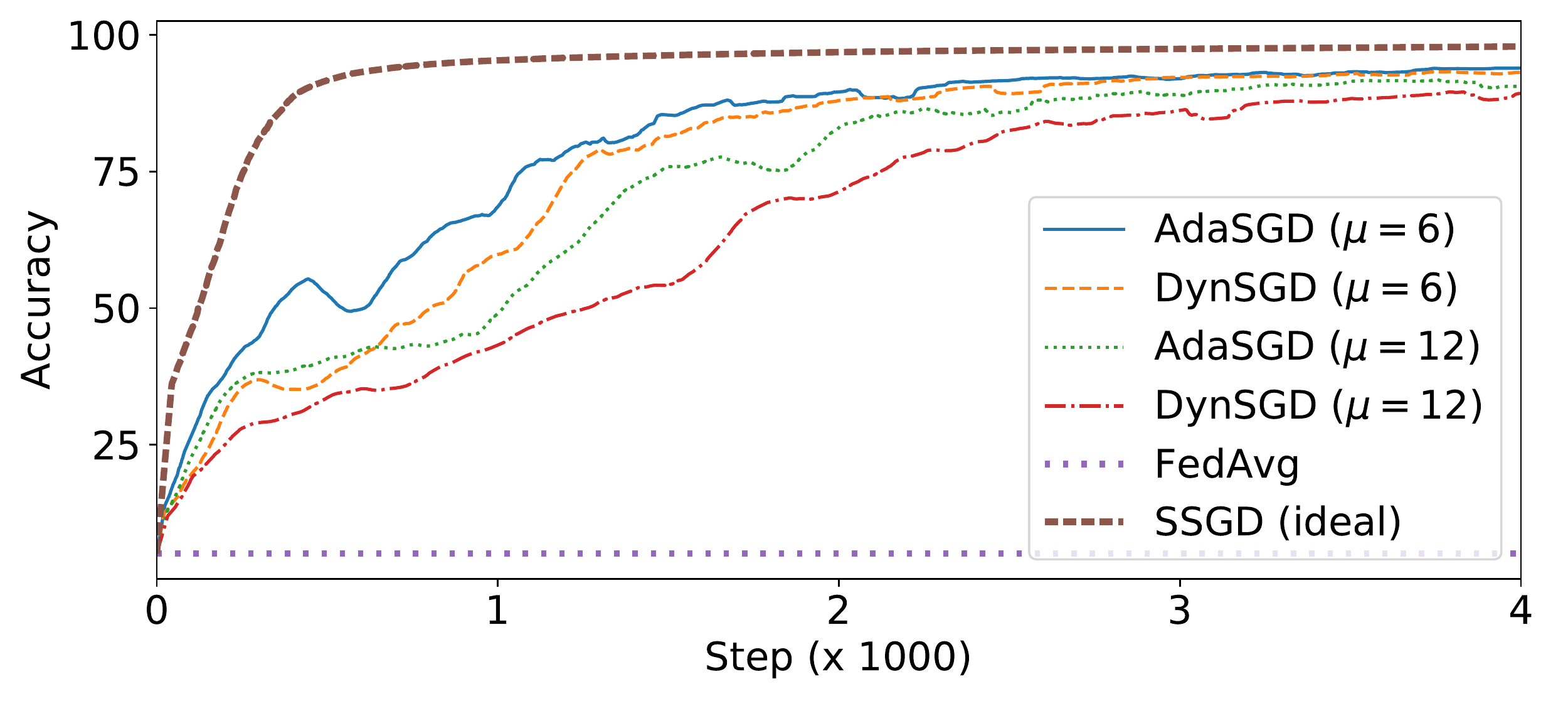}
\caption{Impact of staleness on learning.} 
\label{fig:impact_staleness}
\end{figure}

\paragraph{Staleness-based dampening.} 
\cref{fig:impact_staleness} depicts that \algo outperforms the alternative learning schemes for the non-IID version of MNIST.
As expected, the staleness-free scenario (SSGD) delivers the fastest (ideal) convergence, whereas the \emph{staleness-unaware} \baselinesgd diverges.
The comparison between the two staleness-aware algorithms (\dynsgd and \algo) shows that our solution (\algo) better adapts the dampening factor to the noise introduced by stale gradients (\cref{sec:ada}).
\algo reaches 80\% accuracy 14.4\% faster than \dynsgd for $D1$ and 18.4\% for $D2$.
\cref{fig:impact_staleness} also depicts the impact of staleness on \dynsgd and \algo.
We observe that the larger the staleness, the slower the convergence of both algorithms.
The advantage of \algo over \dynsgd grows with the amount of staleness as the larger amount of noise gives more leeway to \algo to benefit from its superior dampening scheme.

\begin{figure}[!ht]
\centering 
\subfloat[]{\label{subfig:outlier_acc}\includegraphics[width=0.5\linewidth,keepaspectratio]{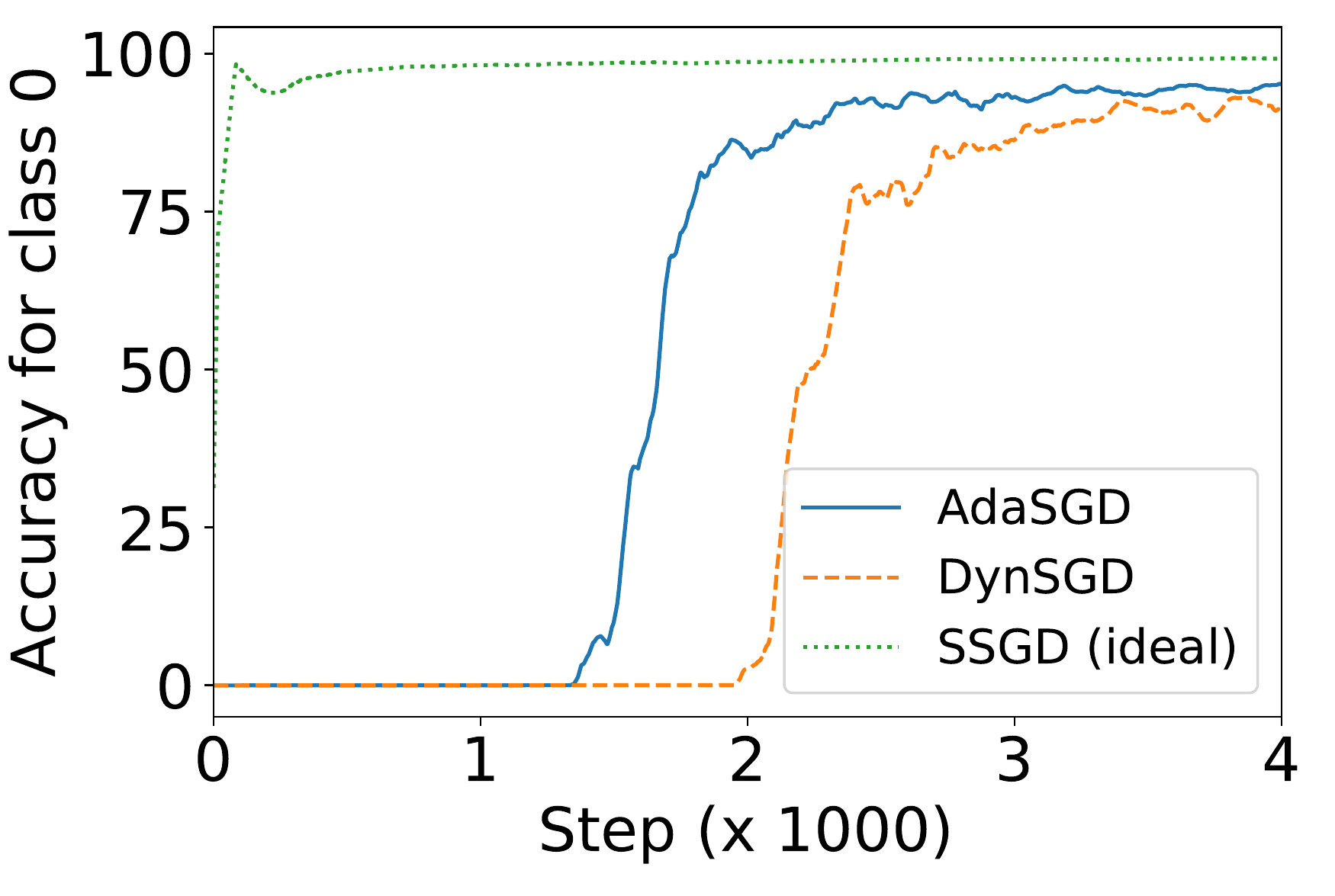}}
\subfloat[]{\label{subfig:outlier_dist}\includegraphics[width=0.5\linewidth,keepaspectratio]{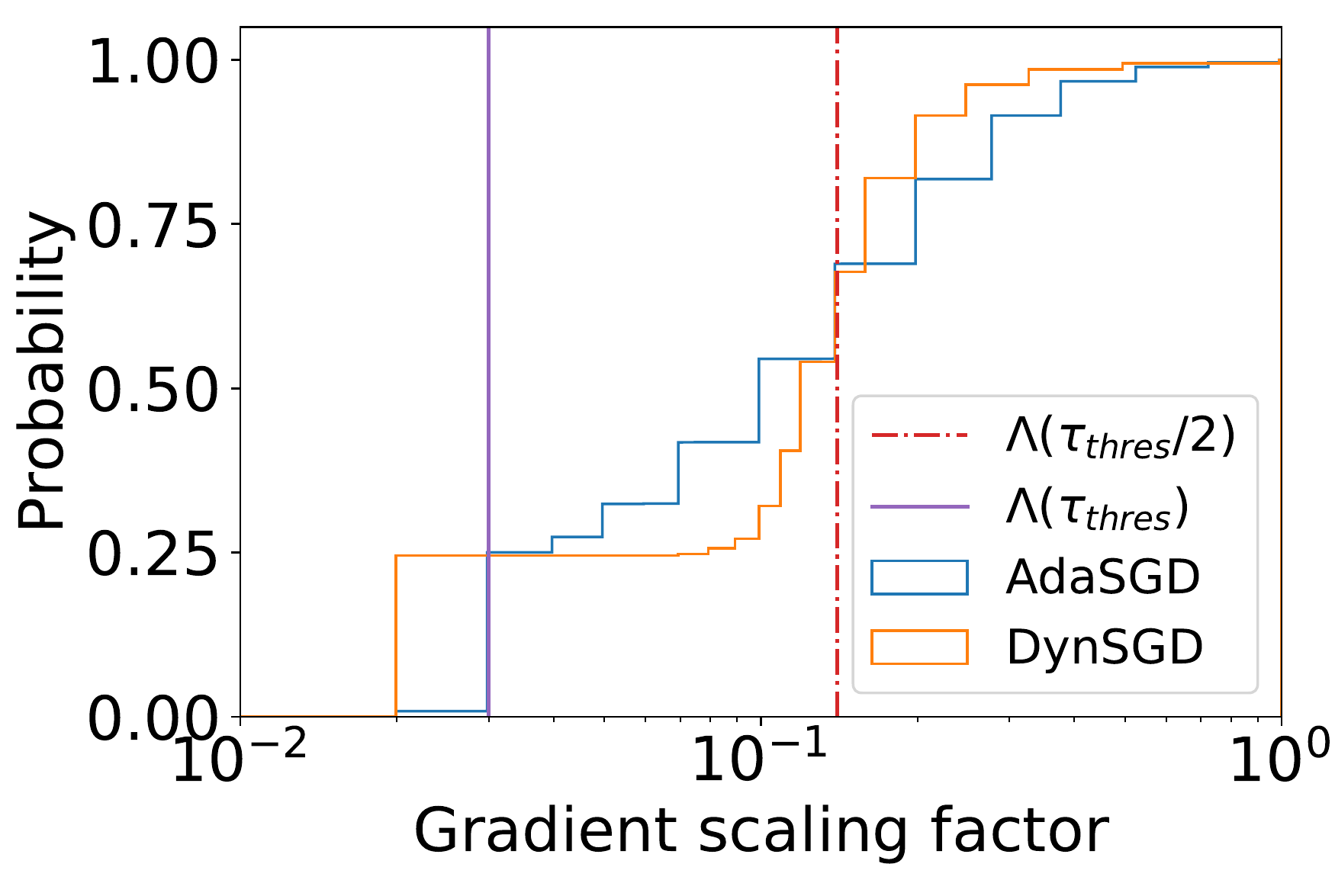}}%
\caption{Impact of long tail staleness on learning.} 
\label{fig:outlier}
\end{figure}

\paragraph{Similarity-based boosting.}
We evaluate the effectiveness of the similarity-based boosting property of \algo (\cref{sec:ada}) in the case of long tail staleness (\cref{fig:hashtagStaledist}).
We employ the non-IID MNIST dataset, $D1$ (thus $\tau_\text{thres}$ is 12) and set the staleness to $4 \cdot \tau_\text{thres} = 48$ for all the gradients computed on data with class 0.
This setup essentially captures the case where a particular label is only present in stragglers.
\cref{subfig:outlier_acc} shows that \algo incorporates the knowledge from class 0 much faster than \dynsgd.

\cref{subfig:outlier_dist} shows the CDF for the dampening values used to weight the gradients of \cref{subfig:outlier_acc}.
We mark the two points of interest regarding the $\tau_\text{thres}$ by vertical lines (as also shown in \cref{fig:scaling_factor}).
If \algo had no similarity-based boosting, all updates related to class 0 would almost not be taken into account, as they would be nullified by the exponential dampening function, therefore leading to a model with poor predictions for this class.
Given the low class similarity of the learning tasks involving class 0, \algo boosts their dampening value.
The second vertical line denotes the staleness value ($\frac{\tau_\text{thres}}{2} = 6$) for which \algo and \dynsgd give the same dampening value ($0.14$).
The slope of each curve at this point indicates that the dampening values for \dynsgd are more concentrated whereas the ones for \algo are more spread around this value.

\begin{figure}[!t]
\centering 
\subfloat[E-MNIST]{\label{subfig:accemnist}\includegraphics[width=0.5\linewidth,keepaspectratio]{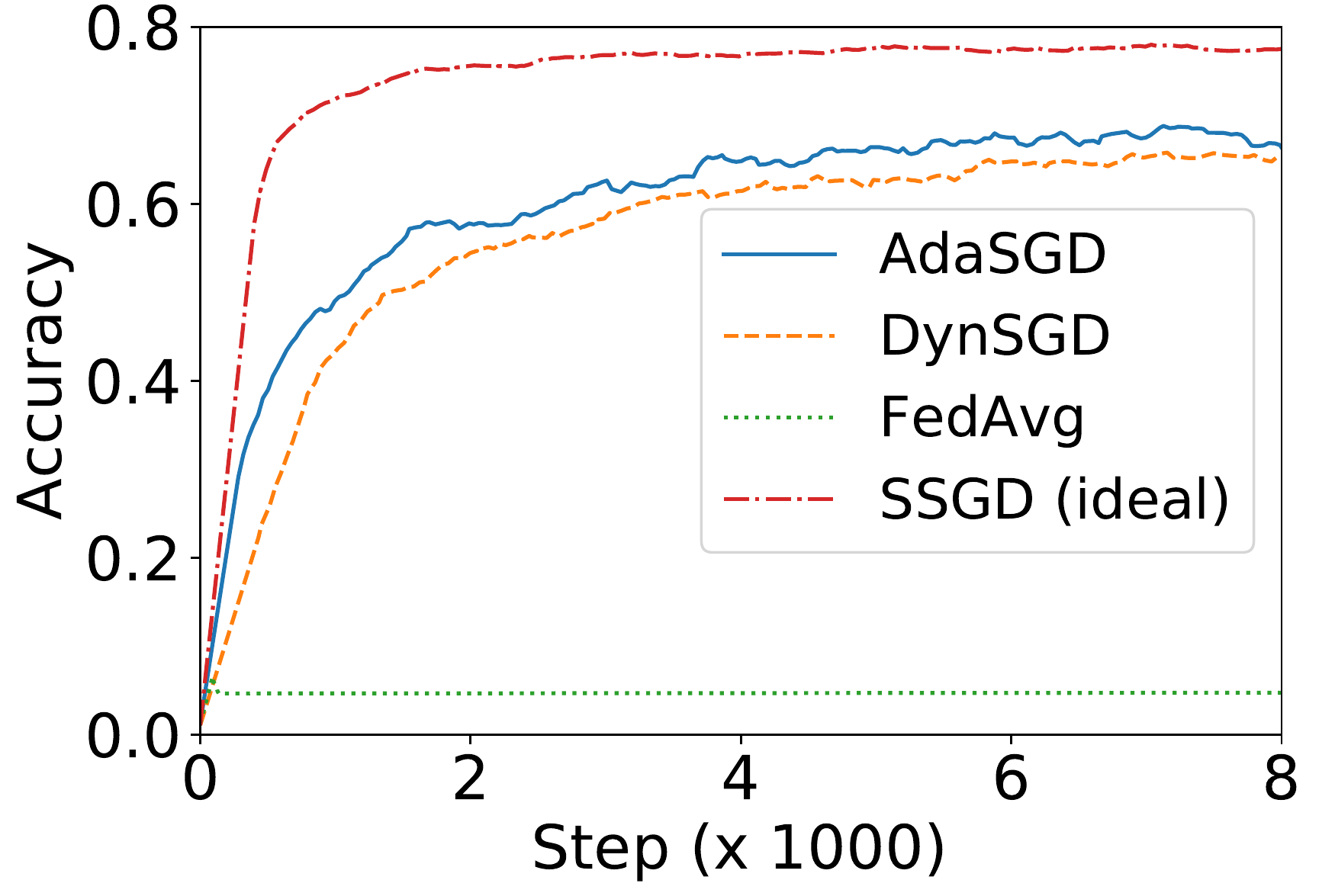}} 
\subfloat[CIFAR-100]{\label{subfig:acccifar}\includegraphics[width=0.5\linewidth,keepaspectratio]{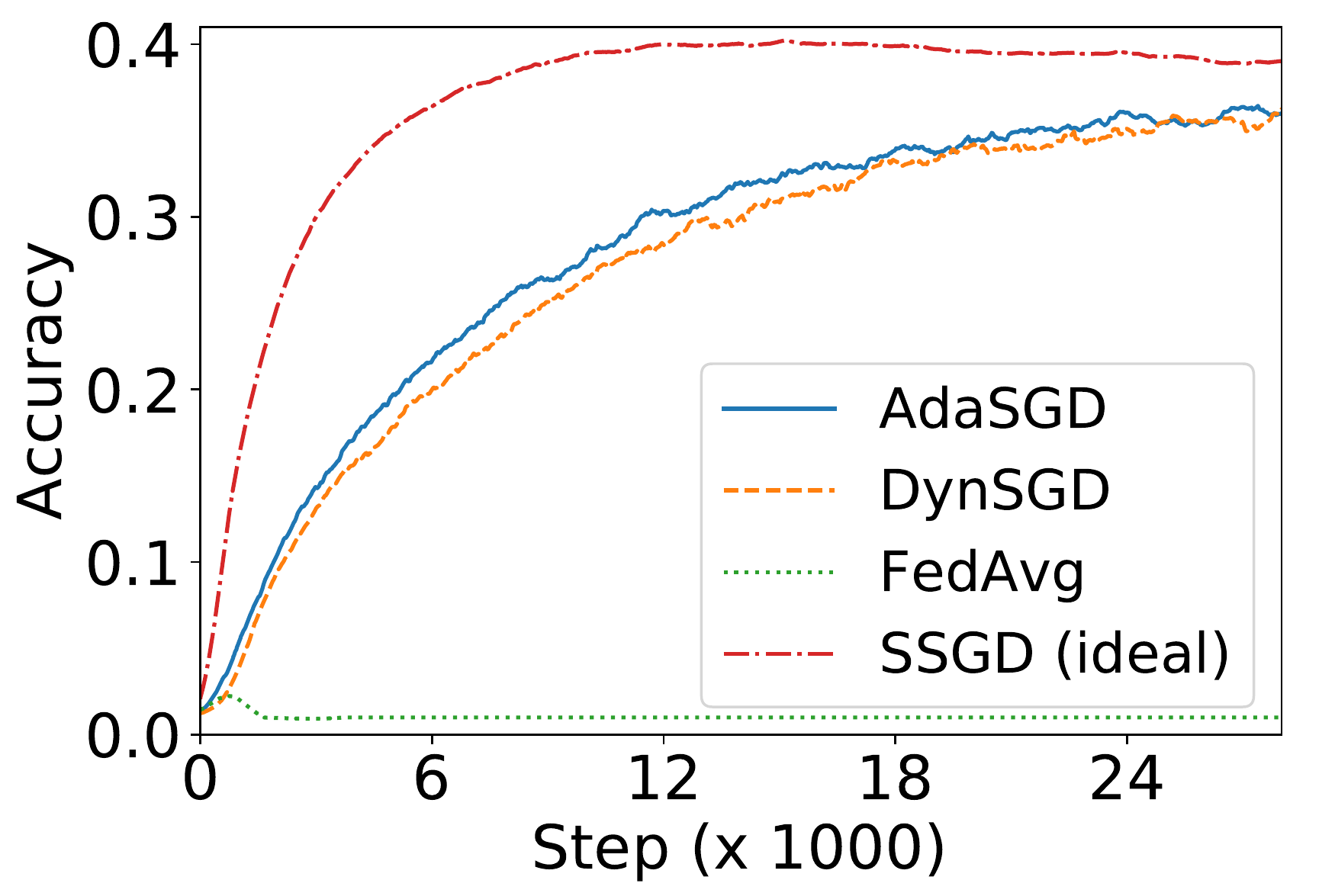}}
\caption{Staleness awareness with IID data.}
\label{fig:staleness}
\end{figure}

\paragraph{IID data.}
Although data are more likely to be non-IID in an FL environment, the data collected on mobile devices might in some cases be IID. %
We thus benchmark \algo under two additional datasets (E-MNIST and CIFAR-100) with the staleness following $D2$.
\cref{fig:staleness} shows that our observations from \cref{fig:impact_staleness} hold also with IID data.
As with non-IID data, \baselinesgd diverges also in the IID setting, and \algo performs better than \dynsgd on both datasets.

\paragraph{Differential privacy.}
Differential privacy~\cite{dwork2014algorithmic} is a popular technique for privacy-preserving FL with formal guarantees~\cite{bonawitz2017practical}.
We thus compare \algo against \dynsgd in a differentially private setup by perturbing the gradients as in~\cite{abadi2016deep}.
We keep the previous setup (IID data with $D2$) and employ the MNIST dataset.
Based on~\cite{wu2017bolt}, we fix the probability $\delta = 1/N^2 = \frac{1}{60000^2}$ and measure the privacy loss ($\epsilon$) with the \emph{moments accountant} approach~\cite{abadi2016deep} given the sampling ratio ($\frac{\text{mini-batch size}}{N} = \frac{100}{60000}$), the noise amplitude, and the total number of iterations.

\begin{figure}[!t]
\centering 
\includegraphics[width=\linewidth,keepaspectratio]{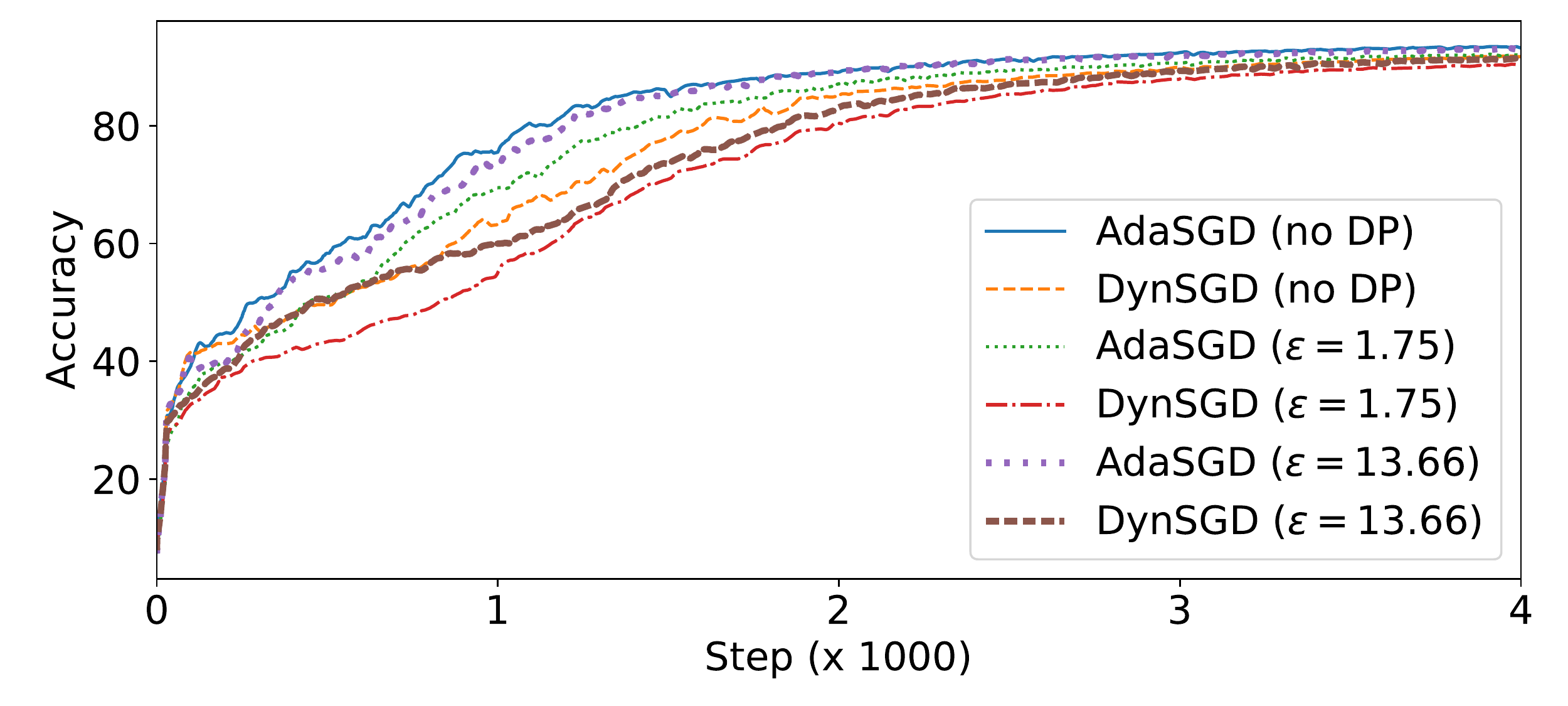}
\caption{Staleness awareness with differential privacy.} 
\label{fig:dp}
\end{figure}

\cref{fig:dp} demonstrates that the advantage of \algo over \dynsgd also holds in the differentially private setup.
A better privacy guarantee (i.e., smaller $\epsilon$) slows down the convergence for both staleness-aware learning schemes.

\begin{figure*}[!t]
\centering 
\subfloat[Request schedule]{\includegraphics[width=0.25\linewidth,keepaspectratio]{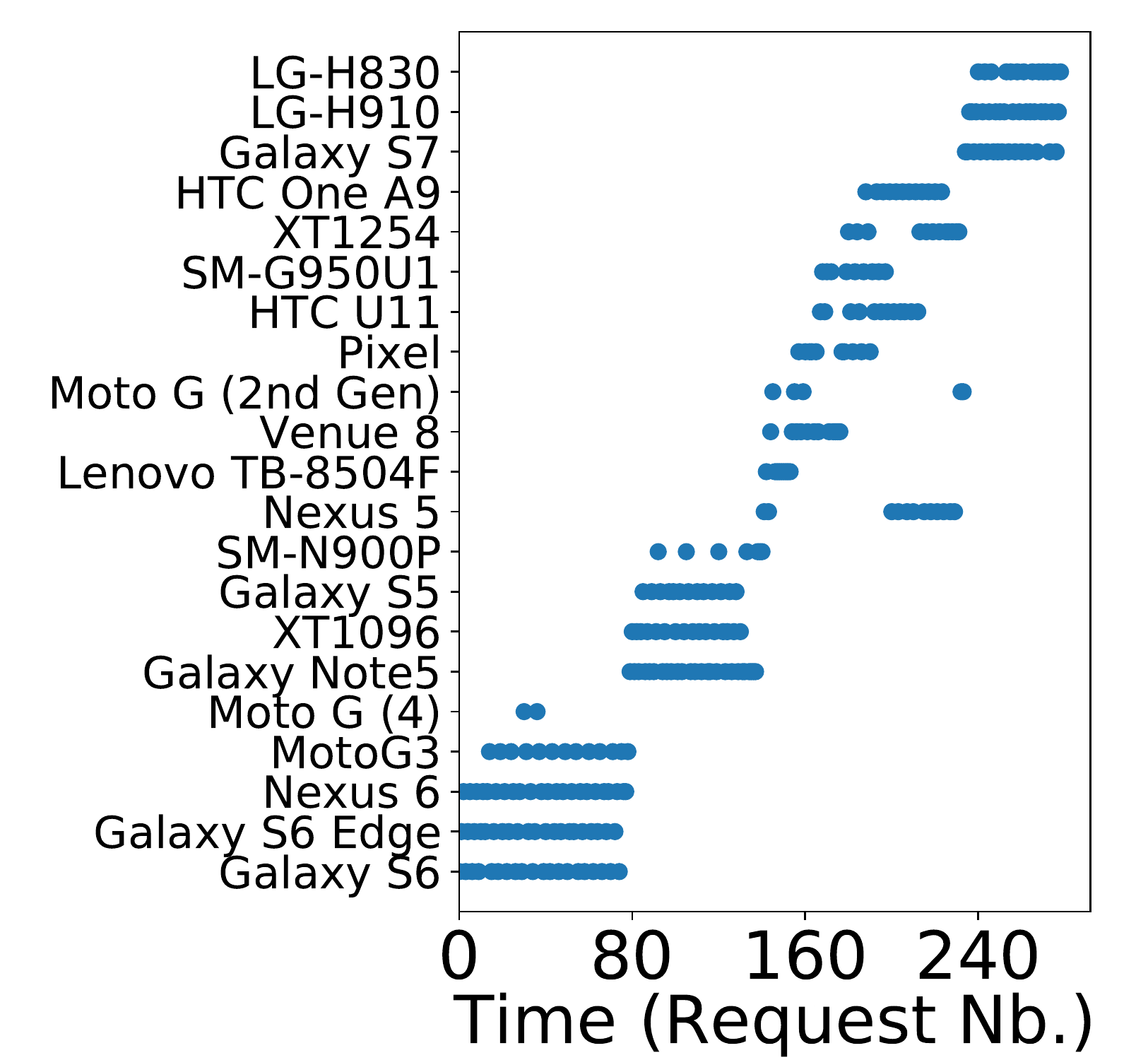}
\label{subfig:loginSchedule}}
\subfloat[error CDF]{\includegraphics[width=0.1875\linewidth,keepaspectratio]{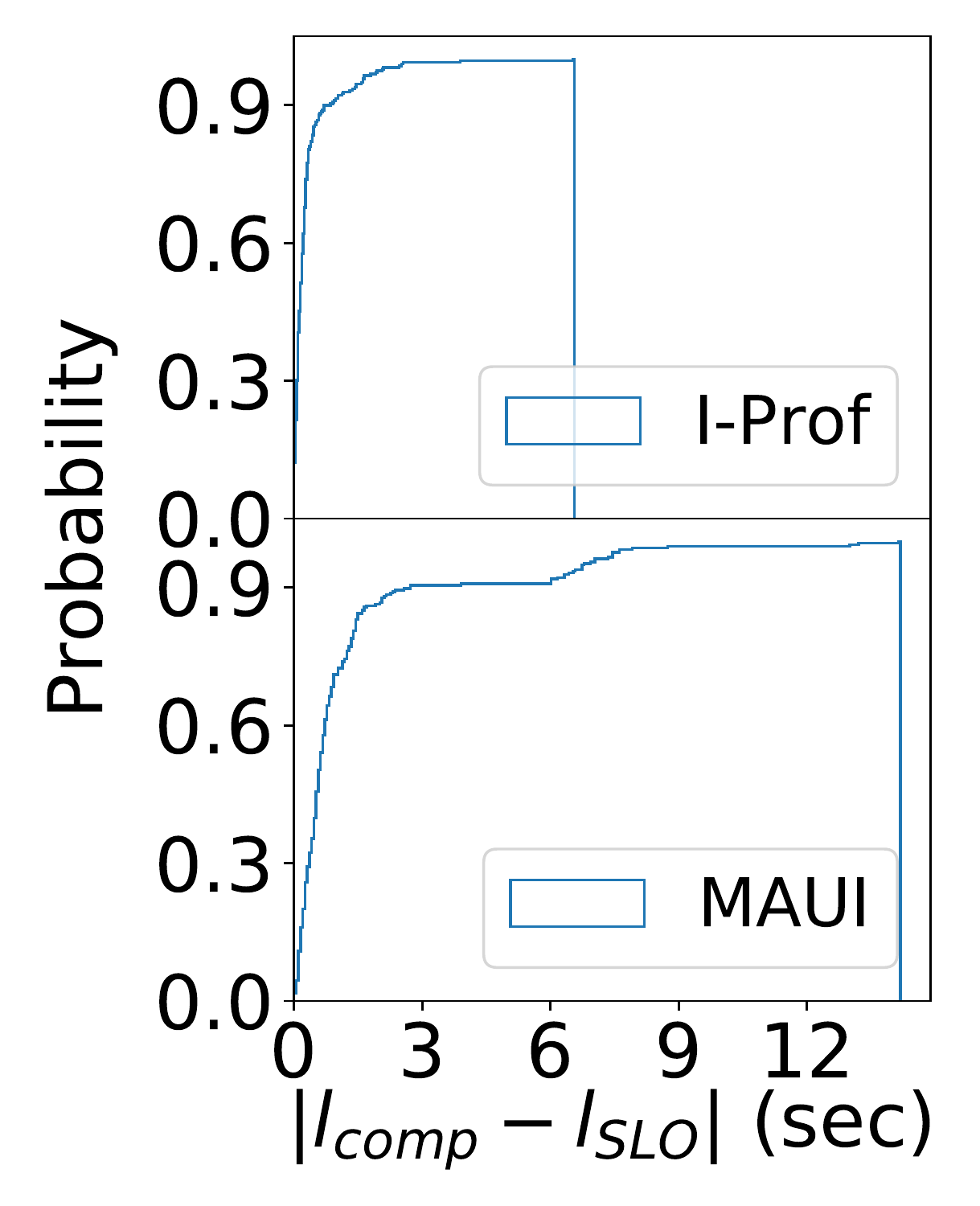}
\label{subfig:latcdf}}
\subfloat[Request computation time]{\includegraphics[width=0.375\linewidth,keepaspectratio]{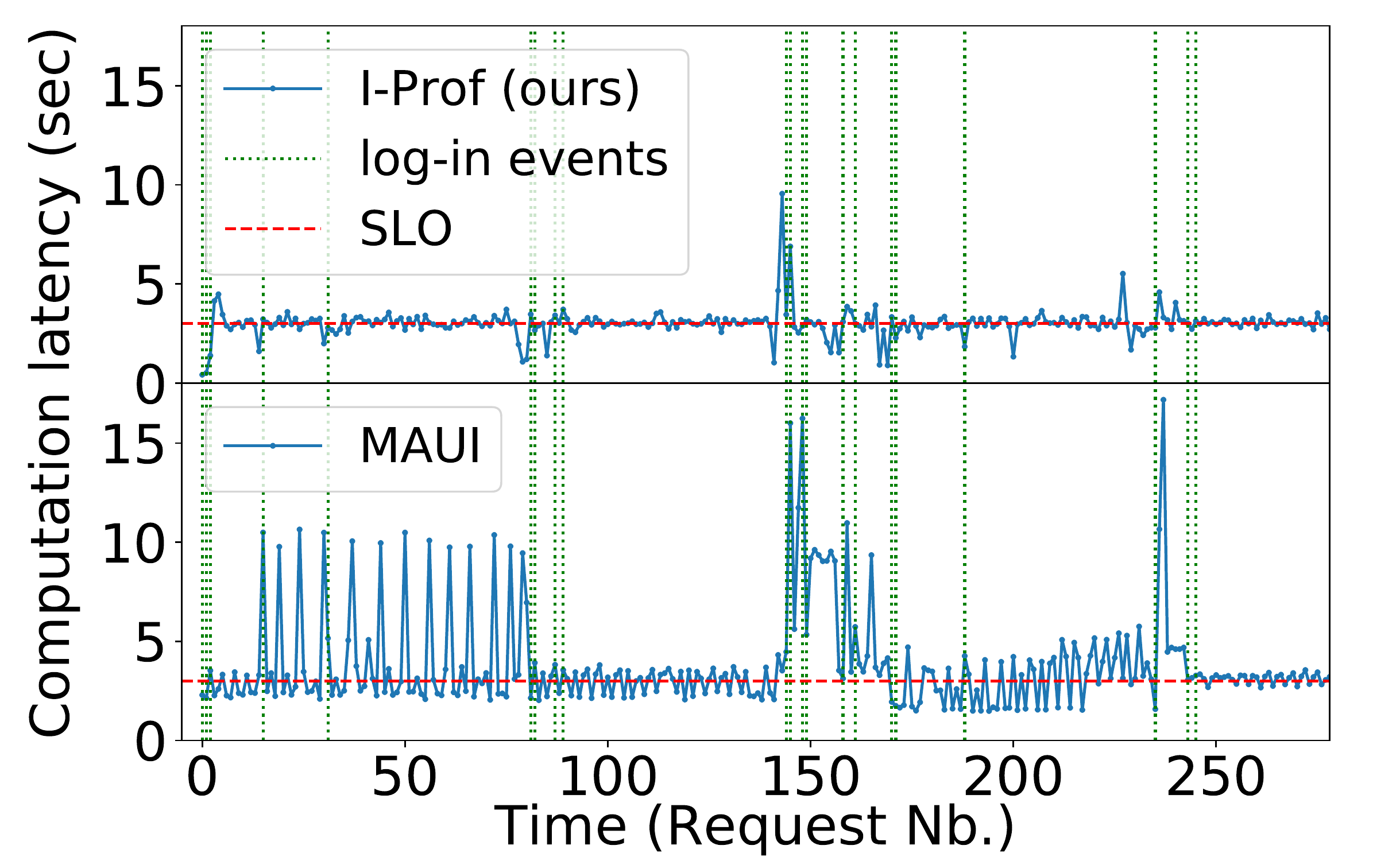}
\label{subfig:latdeviation}}
\subfloat[Profiler output]{\includegraphics[width=0.1875\linewidth,keepaspectratio]{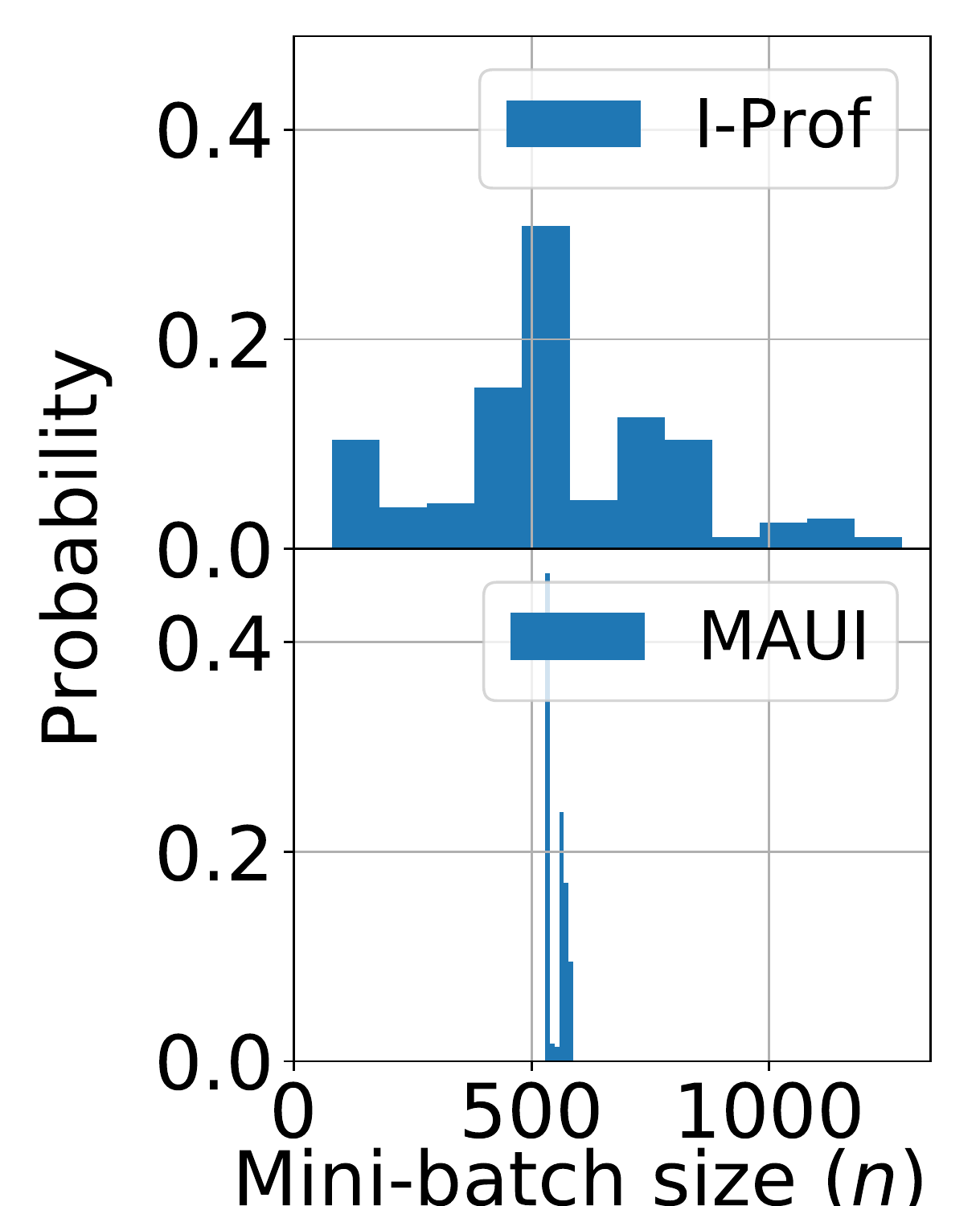}
\label{subfig:outVar}}
\caption{\profiler outperforms MAUI and drives the computation time closer to the SLO.
}
\label{fig:latProfEval}
\end{figure*}

\subsection{\profiler Performance} \label{sec:profilerEval}

We compare \profiler against the profiler of MAUI~\cite{cuervo2010maui}, a mobile device profiler aiming to identify the most energy-consuming parts of the code and offload them to the cloud.
MAUI predicts the energy by using a linear regression model (similar to the global model of \profiler) on the number of CPU cycles ($\hat{E} = \theta_0 \cdot n$), to essentially capture how the size of the workload affects the energy (as in~\cite{mittal2012empowering}).
We adapt the profiler of MAUI to our setup by replacing the CPU cycles with the mini-batch size for two main reasons. 
First, our workload has a static code path so the number of CPU cycles on a particular mobile device is directly proportional to the mini-batch size. 
Second, measuring the number of executed CPU cycles requires root access that is not available on AWS.

We bootstrap the global model of \profiler and the model of MAUI by pre-training on a training dataset.
To this end, we use 15 mobile devices in AWS (that are different from the ones used for the rest of the experiments), assign them learning tasks with increasing mini-batch size until the computation time becomes 2 times the SLO, and collect their device information for each task.
We rely on the same methodology to evaluate energy consumption but use only 3 mobile devices in our lab, as AWS prohibits energy measurements.

For testing, we use a different set of 20 commercial mobile devices in AWS, each performing requests for the image classification application (on MNIST), starting at different timestamps (log-in events) as shown in \cref{subfig:loginSchedule}.
In order to ensure a precise comparison with MAUI, we add a \emph{round-robin dispatcher} to the profiler component which alternates the requests from a given device between \profiler and MAUI.

\paragraph{Computation time SLO.}
 \cref{subfig:latcdf} shows that \profiler largely outperforms MAUI in terms of deviation from the computation time SLO.
90\% of approximately 280 learning tasks deviate from an SLO of 3 seconds by at most 0.75 seconds with \profiler and 2.7 seconds with MAUI.
This is the direct outcome of our  design decisions.
First, \profiler adds dynamic features 
(e.g., the temperature of the device) to train its global model (\cref{sec:profiler}).
As a result, the predictions are more accurate for the first request of each user. 
Second, \profiler uses a personalized model for each device that reduces the error (deviation from the SLO) with every subsequent request (\cref{subfig:latdeviation}).
\cref{subfig:outVar} shows that the personalized models of \profiler are able to output a wider range of mini-batch sizes that better match the capabilities of individual devices.
On the contrary, MAUI relies on a simple linear regression model which has acceptable accuracy for its use-case but is inefficient when profiling heterogeneous mobile devices.

\paragraph{Energy SLO.}
To assess the ability of \profiler to also target the energy SLO, we use the same setup as for the computation time, except on 5 mobile devices\footnote{AWS prohibits energy measurements so we only rely on devices available in our lab, listed in their log-in order: Honor 10, Galaxy S8, Galaxy S7, Galaxy S4 mini, Xperia E3.}.
We configure \profiler with a significantly smaller error margin, $\epsilon=6 * 10^{-5}$ (\cref{eq:paloss}), because the linear relation (capture by $\alpha$ as defined in \cref{sec:profiler}) is significantly smaller for the energy than for the computation time (as shown in \cref{fig:hypothesis}). 

\begin{figure}%
\centering 
\subfloat[\profiler (ours)]{\includegraphics[width=0.5\linewidth,keepaspectratio]{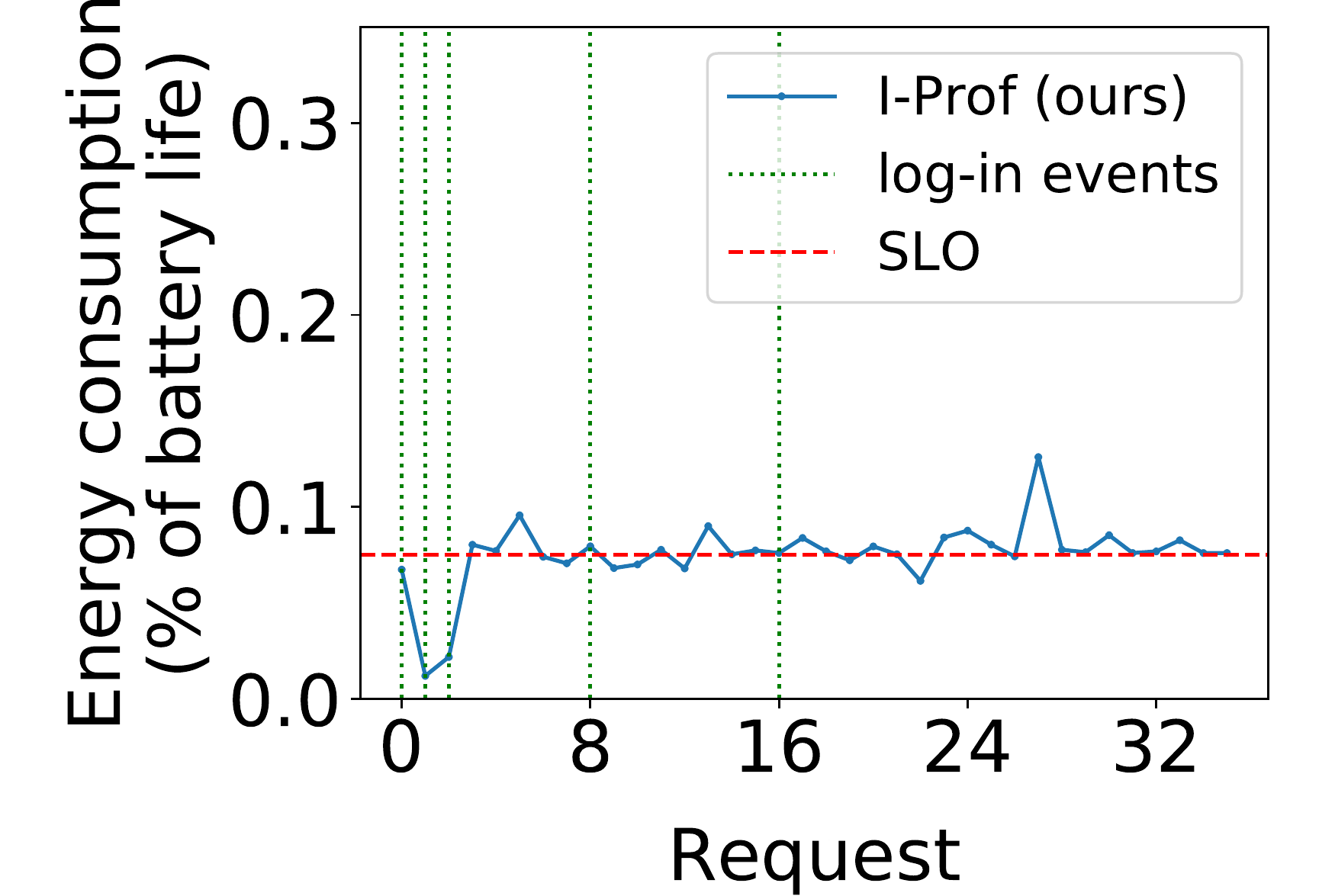}}
\subfloat[MAUI]{\includegraphics[width=0.5\linewidth,keepaspectratio]{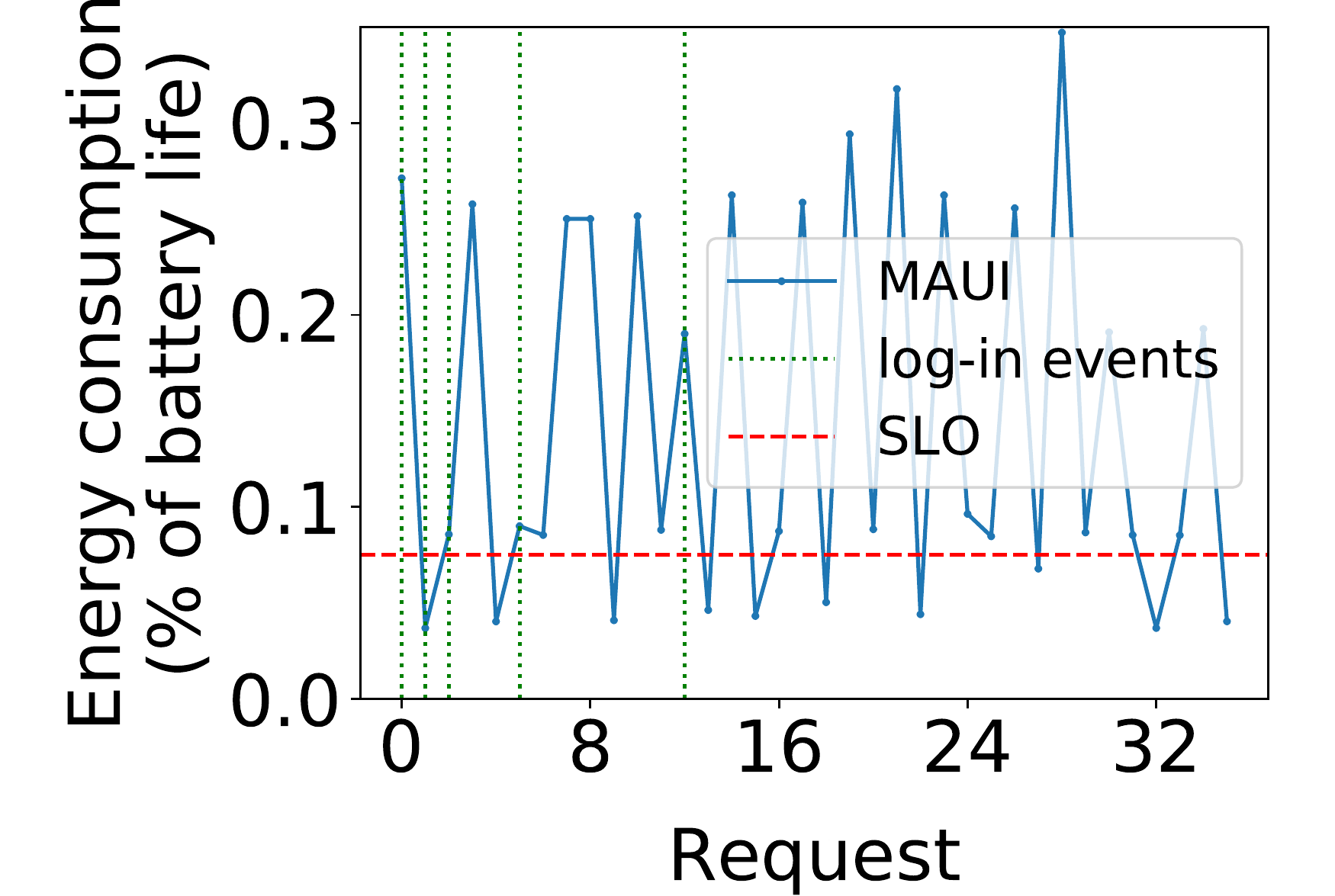}}
\vspace{2mm}
\caption{\profiler outperforms MAUI and drives the energy closer to the SLO.
}
\label{fig:energyProfEval}
\end{figure}

\cref{fig:energyProfEval} shows that \profiler significantly outperforms MAUI in terms of deviation from the energy SLO. 
90\% of 36 learning tasks deviate from an SLO of 0.075\% battery drop by at most 0.01\% for \profiler and 0.19\% for MAUI.
The observation that \profiler is able to closely match the latency SLO, while MAUI suffers from huge deviations, holds for the energy too.
The PA personalized models are able to quickly adapt to the state of the device as opposed to the linear model of MAUI that provides biased predictions.

\subsection{Resource Allocation} \label{sec:evalResourceAlloc}

We evaluate our resource allocation scheme (\cref{sec:impl}) and compare it against CALOREE~\cite{mishra2018caloree} which is a state of the art resource manager for mobile devices. The goal of CALOREE is to optimize resource allocation in order for the workload execution to meet its predefined deadline while minimizing the energy consumption.
To this end, CALOREE profiles the target device by running the workload with different resource configurations (i.e., number of cores, core frequency).
Since \system executes on non-rooted mobile devices, we can only adapt the number of big/little cores (but not their frequencies). 
By varying the number of cores allocated to our workload (i.e., gradient computation), we obtain the energy consumption of each possible configuration. From these configurations, CALOREE only selects those with the optimal energy consumption (the lower convex hull) which are packed in the so called \emph{performance hash table} (PHT). %

\paragraph{CALOREE on new devices.}
In their thorough evaluation, the authors of CALOREE used the same device for training and running the workloads.
Therefore, we first benchmark the performance of CALOREE when running on new devices.
We employ Galaxy S7 to collect the PHT and set the mini-batch size that \profiler gives for a latency SLO of 3 seconds (\cref{sec:profilerEval}).
We then run this workload with CALOREE on different mobile devices, as shown in \cref{table:caloree_coldstart}.

The performance of CALOREE degrades significantly when running on a different device than the one used for training.
The first line of \cref{table:caloree_coldstart} shows the baseline error when running on the same device.
The error increases more than 6$\times$ for a device with similar architecture and the same vendor (Galaxy S8) and more than 32$\times$ for a device of similar architecture but different vendor (Honor 9 and 10).
This significant increase for the error is due to the heterogeneity of the mobile devices which make PHTs not applicable across different device models.

\begin{table}[!t]
\centering
\caption{Performance of CALOREE~\cite{mishra2018caloree} on new devices.}
\begin{tabular}{|c|c|}
\hline
Running device             & Deadline error (\%) \\ \hline \hline
Galaxy S7 & 1.4        \\ \hline
Galaxy S8        & 9          \\ \hline
Honor 9          & 46         \\ \hline
Honor 10         & 255        \\ \hline
\end{tabular}
\label{table:caloree_coldstart}
\end{table}

\begin{figure}[!ht]
\centering
\includegraphics[width=\linewidth]{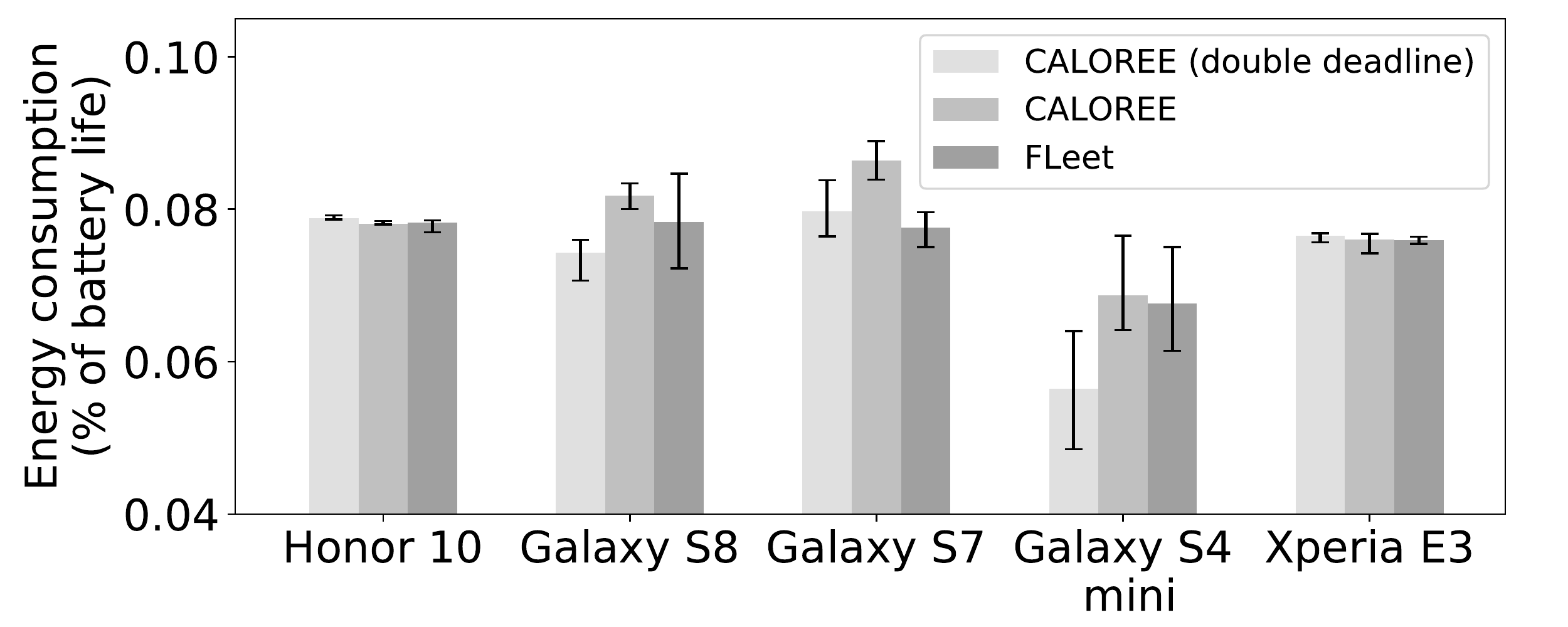}
\caption{Resource allocation of \system vs. CALOREE. %
  }
\label{fig:caloreeVSheuristic}
\end{figure}

\paragraph{CALOREE vs \system.}
We evaluate the resource allocation scheme of \system by comparing it to the ideal environment for CALOREE, i.e., training and running on the same device (a setup nevertheless difficult to achieve in FL with millions of devices).
Following the setup used for the energy SLO evaluation (\cref{sec:profilerEval}), we employ 5 devices and fix the size of the workload (mini-batch size) based on the output of \profiler.
In particular we set the mini-batch size to 280, 4320, 6720, 5280, 1200 for the devices shown in \cref{fig:caloreeVSheuristic} respectively.
We set the deadline of CALOREE either equal or double than the computation latency of \system.
We take 10 measurements and report on the median, 10th and 90th percentile.

\cref{fig:caloreeVSheuristic} shows the fact that in the ideal environment for CALOREE and even with double the time budget (giving more flexibility to CALOREE), \system has comparable energy consumption.
Since gradient computation is a compute intensive task with high temporal and spacial cache locality, the configuration changes performed by CALOREE negatively impact the execution time and cancel any energy saved by its advanced resource allocation scheme.
Additionally, the fewer configuration knobs available on non-rooted Android devices limit the full potential of CALOREE.

\subsection{Learning Task Assignment Control} \label{sec:evalController}

\begin{figure}[!ht]
\centering 
\subfloat[Based on mini-batch size]{\label{subfig:batchThres}\includegraphics[width=\linewidth,keepaspectratio]{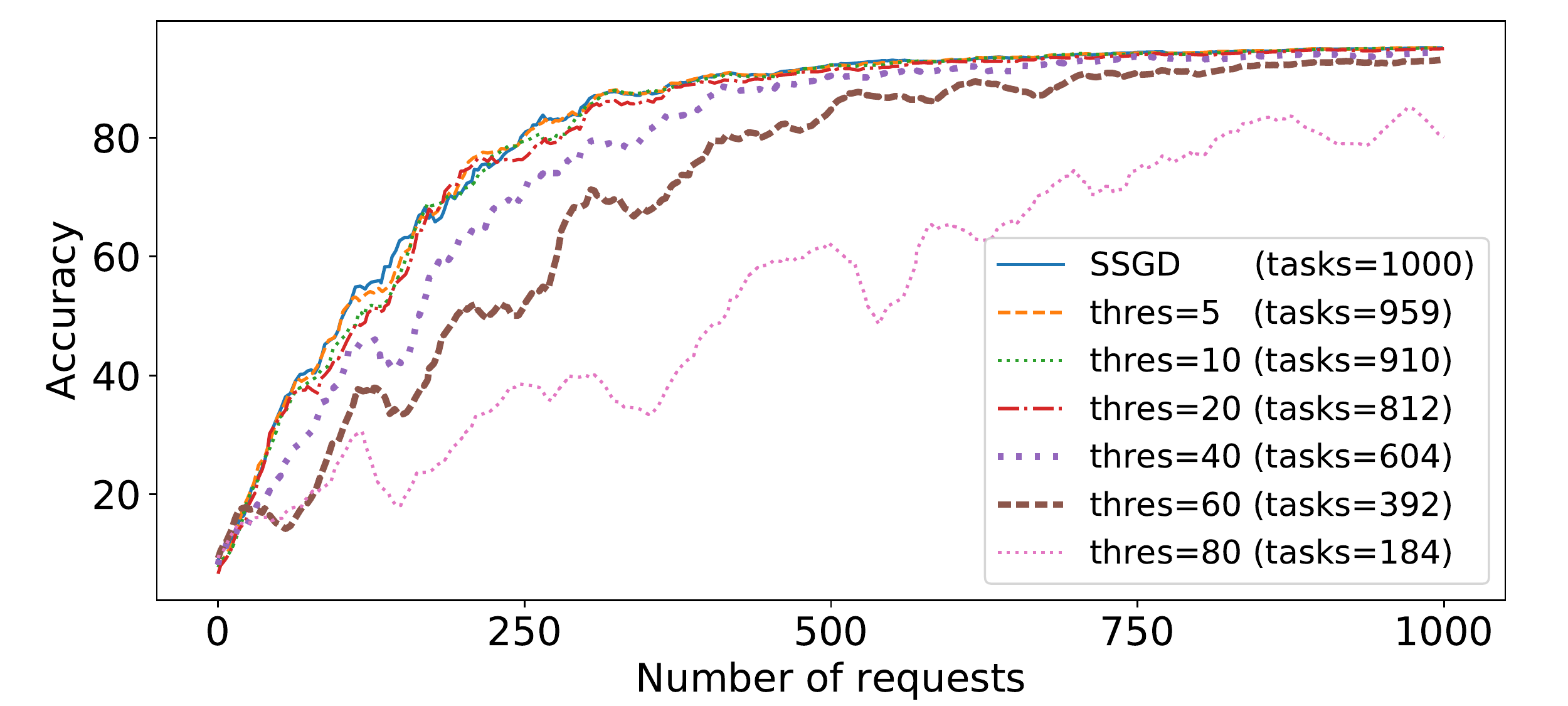}} 
\\
\subfloat[Based on similarity]{\label{subfig:simThres}\includegraphics[width=\linewidth,keepaspectratio]{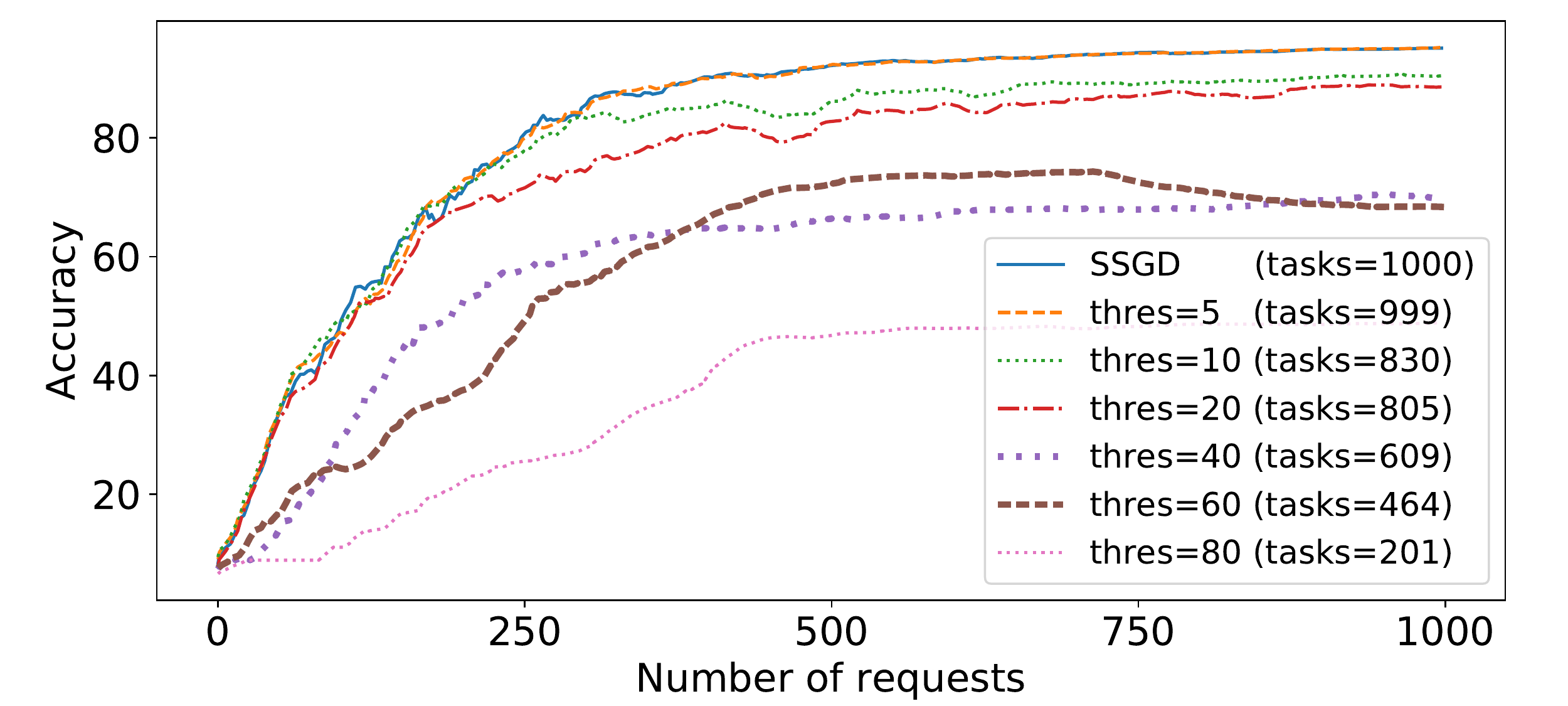}}
\caption{Threshold-based pruning.} 
\label{fig:thres}
\end{figure}

The controller of \system employs a threshold to prune learning tasks and thus control the trade-off between the cost of the gradient computations and the model prediction quality.
This threshold can be based either on the mini-batch size or on the similarity values (\cref{fig:arch}).
To evaluate this trade-off, we employ non-IID MNIST with the mini-batch size following a Gaussian distribution $\cN (\mu=100, \sigma=33)$ (based on the distribution of the output of \profiler shown in \cref{subfig:outVar}), and set the threshold to the $n-\text{th}$ percentile of the past values.
\cref{fig:thres} illustrates that a threshold on the mini-batch size is more effective in pruning the less useful gradient computations than a threshold on the similarity.
\cref{subfig:batchThres} shows that even dropping up to 39.2\% of the gradients (with the smallest mini-batch size) has a negligible impact on the accuracy (less than 2.2\%).
\cref{subfig:simThres} shows that one can drop 17\% of the most similar gradients with an accuracy impact of 4.8\%.

%% file: RelWork.tex
\section{Related Work}
\label{sec:relwork}

\paragraph{Distributed ML.}
Adam~\cite{chilimbi2014project} and TensorFlow~\cite{abadi2016tensorflow} adopt the parameter server architecture~\cite{li2014scaling} for scaling out on high-end machines, and typically require cross-worker communication.
\system also follows the parameter server architecture, by maintaining a global model on the server.
However, \system avoids cross-worker communication, which is impractical for mobile workers due to the device churn.

A common approach for large-scale ML is to control the amount of staleness for boosting convergence~\cite{cui2014exploiting,qiao2018litz}.
In Online FL, staleness cannot be controlled as this would impact the model update frequency.
The workers perform learning tasks asynchronously with end-to-end latencies that can differ significantly (due to device heterogeneity and network variability) or even become infinite (user disconnects).

Petuum~\cite{xing2015petuum} and TensorFlow handle faults (worker crashes) by checkpointing and repartitioning the model across the workers whenever failures are detected.
In a setting with mobile devices, such failures may appear very often, thus increasing the overhead for checkpointing and repartitioning.
\system does not require any fault-tolerance mechanism for its workers, as from a global learning perspective, they can be viewed as stateless.

\paragraph{Federated learning.}
In order to minimize the impact on mobile devices, Standard FL algorithms~\cite{jeong2018communication,mcmahan2017communication,smith2017federated,bonawitz2019towards} require the learning task to be executed only when the devices are idle, plugged in, and on a free wireless connection.
However, in \cref{sec:onlinefl}, we have shown that these requirements may drastically impact the performance of some applications.  
Noteworthy, techniques for reducing the communication overhead~\cite{jeong2018communication} or increasing the robustness against adversarial users~\cite{kardam,damaskinos2019aggregathor}, are orthogonal to the online characteristic so they can be adapted for \algo, and plugged into \system.

\paragraph{Performance prediction for mobile devices.}
Estimating the computation time or energy consumption of an application running on a mobile device is a very broad area of research.
Existing approaches~\cite{kwon2013mantis,chowdhury2015system,yoon2012appscope,hao2013estimating,carroll2010analysis,chu2011balancing} target multiple applications generally executing on a single device.
They typically benchmark the device or monitor hardware and OS-level counters that require root access.
In contrast, \system targets a single application executing in the same way across a large range of devices.
\profiler poses a negligible overhead, as it employs features only from the standard Android API to enable Online FL, and requires no benchmarking of new devices.
\profiler is designed to make predictions for \emph{unknown} devices.

Neurosurgeon~\cite{kang2017neurosurgeon} is a scheduler that minimizes the end-to-end computation time of inference tasks (whereas \system focuses on training tasks), by choosing the optimal partition for a neural network and offloading computations to the cloud.
The profiler of Neurosurgeon only uses workload-specific features (e.g., number of filters or neurons) to estimate computation time and energy, and ignores device-specific features.
By contrast, mobile phones, as targeted by \profiler\footnote{In their in-depth experimental evaluation the authors of~\cite{kang2017neurosurgeon} consider a single hardware platform and not Android mobile devices.}, exhibit a wide range of device-specific characteristics that significantly impact their latency and energy consumption (\cref{fig:hypothesis}).

Systems such as CALOREE~\cite{mishra2018caloree} and LEO~\cite{mishra2015probabilistic} profile mobile devices under different system configurations and train an ML model to determine the ones that minimize the energy consumption.
They rely on a control loop to switch between these configurations such that the application does not miss the preset deadline. 
Due to the restrictions of the standard Android API, the available knobs are limited in our setup. 
For our application (i.e., gradient computation), we show that a simple resource allocation scheme (\cref{sec:impl}) is preferable even in comparison with an ideal execution model.

%% file: Conclusion.tex
\section{Concluding Remarks} \label{sec:concl}
This paper presented \system, the first system that enables \emph{online} ML at the edge.
\system employs \profiler, a new \emph{ML-based profiler} which determines the ML workload that each device can perform within predefined energy and computation time SLOs.  
\system also makes use of \algo, a new \emph{staleness-aware learning}  algorithm that is optimized for Online FL.
We showed the performance of \profiler and \algo on commercial Android devices with popular benchmarks.
\profiler can be used for any iterative computation with embarrassingly-parallel CPU-bound tasks while \algo is specific to gradient-descent and thus ML applications.
In our performance evaluation we do not focus on network and scalability aspects that are orthogonal to our work and addressed in existing literature.
We also highlight that transferring the label and device information (\cref{fig:arch}) poses a negligible network overhead compared to transferring the relatively large FL learning models.
Finally, addressing biases is an important problem even in cloud-based online ML (not only FL) that we also do not address in this work. 
For Online FL we arguably need to keep some of these biases (e.g., recommend more politics to people that wake up earlier). 
Of course diversity is also crucial.

Although we believe \system to represent a significant advance for online learning at the edge, there is still room 
for improvement.
First, for the energy prediction, \profiler
requires access to the CPU usage that is considered as a security flaw on some Android builds and thus not exposed to all applications.
In this case, 
\profiler 
requires a set of additional permissions that belong to services from Android Runtime.
Second, the transfer of the label distribution from the worker to the server introduces a potential privacy leakage.
However, we highlight that the server has access only to the indices of the labels and not their values.
In this paper, we focus on the protection of the input features and mention the possibility to deactivate the similarity-based boosting feature of \algo in the case that this leakage is detrimental.
We plan to investigate noise addition techniques for bounding this leakage~\cite{dwork2014algorithmic} in our future work.
Finally, theoretically proving the convergence of \algo is non-trivial due to the unbounded staleness and the non Independent and Identically Distributed (non-IID) datasets among the workers.
In this respect, a dissimilarity assumption similar to~\cite{li2018federated} may facilitate the derivation of the proof.

\section{Acknowledgments} 
We thank Dr. Sonia Ben-Mokhtar, our shepherd Prof. Christine Julien, our colleagues from the DCL laboratory at EPFL and the anonymous reviewers, for their helpful feedback. This work was supported by the EPFL-Inria International Lab and by the French ANR project PAMELA n.~ANR-16-CE23-0016.

%% file: main.bbl

\begin{thebibliography}{86}

%%% the Web address.

\ifx \showCODEN    \undefined \def \showCODEN     #1{\unskip}     \fi
\ifx \showDOI      \undefined \def \showDOI       #1{#1}\fi
\ifx \showISBNx    \undefined \def \showISBNx     #1{\unskip}     \fi
\ifx \showISBNxiii \undefined \def \showISBNxiii  #1{\unskip}     \fi
\ifx \showISSN     \undefined \def \showISSN      #1{\unskip}     \fi
\ifx \showLCCN     \undefined \def \showLCCN      #1{\unskip}     \fi
\ifx \shownote     \undefined \def \shownote      #1{#1}          \fi
\ifx \showarticletitle \undefined \def \showarticletitle #1{#1}   \fi
\ifx \showURL      \undefined \def \showURL       {\relax}        \fi
% invisible to TeX
\providecommand\bibfield[2]{#2}
\providecommand\bibinfo[2]{#2}
\providecommand\natexlab[1]{#1}
\providecommand\showeprint[2][]{arXiv:#2}

\bibitem[\protect\citeauthoryear{Abadi, Barham, Chen, Chen, Davis, Dean, Devin,
  Ghemawat, Irving, Isard, et~al\mbox{.}}{Abadi et~al\mbox{.}}{2016a}]%
        {abadi2016tensorflow}
\bibfield{author}{\bibinfo{person}{Mart{\'\i}n Abadi}, \bibinfo{person}{Paul
  Barham}, \bibinfo{person}{Jianmin Chen}, \bibinfo{person}{Zhifeng Chen},
  \bibinfo{person}{Andy Davis}, \bibinfo{person}{Jeffrey Dean},
  \bibinfo{person}{Matthieu Devin}, \bibinfo{person}{Sanjay Ghemawat},
  \bibinfo{person}{Geoffrey Irving}, \bibinfo{person}{Michael Isard},
  {et~al\mbox{.}}} \bibinfo{year}{2016}\natexlab{a}.
\newblock \showarticletitle{TensorFlow: A system for large-scale machine
  learning}. In \bibinfo{booktitle}{\emph{OSDI}}. \bibinfo{pages}{265--283}.
\newblock


\bibitem[\protect\citeauthoryear{Abadi, Chu, Goodfellow, McMahan, Mironov,
  Talwar, and Zhang}{Abadi et~al\mbox{.}}{2016b}]%
        {abadi2016deep}
\bibfield{author}{\bibinfo{person}{Mart{\'\i}n Abadi}, \bibinfo{person}{Andy
  Chu}, \bibinfo{person}{Ian Goodfellow}, \bibinfo{person}{H~Brendan McMahan},
  \bibinfo{person}{Ilya Mironov}, \bibinfo{person}{Kunal Talwar}, {and}
  \bibinfo{person}{Li Zhang}.} \bibinfo{year}{2016}\natexlab{b}.
\newblock \showarticletitle{Deep learning with differential privacy}. In
  \bibinfo{booktitle}{\emph{CCS}}. \bibinfo{publisher}{{ACM}},
  \bibinfo{pages}{308--318}.
\newblock


\bibitem[\protect\citeauthoryear{Al-Lawati and Draper}{Al-Lawati and
  Draper}{2020}]%
        {al2020gradient}
\bibfield{author}{\bibinfo{person}{Haider Al-Lawati} {and}
  \bibinfo{person}{Stark~C Draper}.} \bibinfo{year}{2020}\natexlab{}.
\newblock \showarticletitle{Gradient Delay Analysis in Asynchronous Distributed
  Optimization}. In \bibinfo{booktitle}{\emph{ICASSP}}. IEEE,
  \bibinfo{pages}{4207--4211}.
\newblock


\bibitem[\protect\citeauthoryear{Altamimi, Abdrabou, Naik, and Nayak}{Altamimi
  et~al\mbox{.}}{2015}]%
        {altamimi2015energy}
\bibfield{author}{\bibinfo{person}{Majid Altamimi}, \bibinfo{person}{Atef
  Abdrabou}, \bibinfo{person}{Kshirasagar Naik}, {and} \bibinfo{person}{Amiya
  Nayak}.} \bibinfo{year}{2015}\natexlab{}.
\newblock \showarticletitle{Energy cost models of smartphones for task
  offloading to the cloud}.
\newblock \bibinfo{journal}{\emph{TETC}} \bibinfo{volume}{3},
  \bibinfo{number}{3} (\bibinfo{year}{2015}), \bibinfo{pages}{384--398}.
\newblock


\bibitem[\protect\citeauthoryear{Arm}{Arm}{2020}]%
        {neon}
\bibfield{author}{\bibinfo{person}{Arm}.} \bibinfo{year}{2020}\natexlab{}.
\newblock \bibinfo{title}{{SIMD ISAs | Neon}}.
\newblock
  \bibinfo{howpublished}{\url{https://developer.arm.com/architectures/instruction-sets/simd-isas/neon}}.
\newblock


\bibitem[\protect\citeauthoryear{Arm Mali Graphics Processing Units (GPUs)}{Arm
  Mali Graphics Processing Units (GPUs)}{2020}]%
        {mali2020}
Arm Mali Graphics Processing Units (GPUs) \bibinfo{year}{2020}\natexlab{}.
\newblock
  \bibinfo{howpublished}{\url{https://developer.arm.com/ip-products/graphics-and-multimedia/mali-gpus}}.
\newblock


\bibitem[\protect\citeauthoryear{AWS Device Farm}{AWS Device Farm}{2020}]%
        {deviceFarm}
AWS Device Farm \bibinfo{year}{2020}\natexlab{}.
\newblock \bibinfo{howpublished}{\url{https://aws.amazon.com/device-farm/}}.
\newblock


\bibitem[\protect\citeauthoryear{Bonawitz, Eichner, Grieskamp, Huba, Ingerman,
  Ivanov, Kiddon, Konecny, Mazzocchi, McMahan, et~al\mbox{.}}{Bonawitz
  et~al\mbox{.}}{2019}]%
        {bonawitz2019towards}
\bibfield{author}{\bibinfo{person}{Keith Bonawitz}, \bibinfo{person}{Hubert
  Eichner}, \bibinfo{person}{Wolfgang Grieskamp}, \bibinfo{person}{Dzmitry
  Huba}, \bibinfo{person}{Alex Ingerman}, \bibinfo{person}{Vladimir Ivanov},
  \bibinfo{person}{Chloe Kiddon}, \bibinfo{person}{Jakub Konecny},
  \bibinfo{person}{Stefano Mazzocchi}, \bibinfo{person}{H~Brendan McMahan},
  {et~al\mbox{.}}} \bibinfo{year}{2019}\natexlab{}.
\newblock \showarticletitle{Towards Federated Learning at Scale: System
  Design}.
\newblock \bibinfo{journal}{\emph{Proceedings of the 2nd SysML Conference}}
  (\bibinfo{year}{2019}).
\newblock


\bibitem[\protect\citeauthoryear{Bonawitz, Ivanov, Kreuter, Marcedone, McMahan,
  Patel, Ramage, Segal, and Seth}{Bonawitz et~al\mbox{.}}{2017}]%
        {bonawitz2017practical}
\bibfield{author}{\bibinfo{person}{Keith Bonawitz}, \bibinfo{person}{Vladimir
  Ivanov}, \bibinfo{person}{Ben Kreuter}, \bibinfo{person}{Antonio Marcedone},
  \bibinfo{person}{H~Brendan McMahan}, \bibinfo{person}{Sarvar Patel},
  \bibinfo{person}{Daniel Ramage}, \bibinfo{person}{Aaron Segal}, {and}
  \bibinfo{person}{Karn Seth}.} \bibinfo{year}{2017}\natexlab{}.
\newblock \showarticletitle{Practical secure aggregation for privacy-preserving
  machine learning}. In \bibinfo{booktitle}{\emph{CCS}}. ACM,
  \bibinfo{pages}{1175--1191}.
\newblock


\bibitem[\protect\citeauthoryear{Carroll, Heiser, et~al\mbox{.}}{Carroll
  et~al\mbox{.}}{2010}]%
        {carroll2010analysis}
\bibfield{author}{\bibinfo{person}{Aaron Carroll}, \bibinfo{person}{Gernot
  Heiser}, {et~al\mbox{.}}} \bibinfo{year}{2010}\natexlab{}.
\newblock \showarticletitle{An Analysis of Power Consumption in a Smartphone}.
  In \bibinfo{booktitle}{\emph{USENIX ATC}}, Vol.~\bibinfo{volume}{14}. Boston,
  MA, \bibinfo{pages}{21--21}.
\newblock


\bibitem[\protect\citeauthoryear{Chen, Dong, Li, and He}{Chen
  et~al\mbox{.}}{2018}]%
        {chen2018federated}
\bibfield{author}{\bibinfo{person}{Fei Chen}, \bibinfo{person}{Zhenhua Dong},
  \bibinfo{person}{Zhenguo Li}, {and} \bibinfo{person}{Xiuqiang He}.}
  \bibinfo{year}{2018}\natexlab{}.
\newblock \showarticletitle{Federated Meta-Learning for Recommendation}.
\newblock \bibinfo{journal}{\emph{arXiv preprint arXiv:1802.07876}}
  (\bibinfo{year}{2018}).
\newblock


\bibitem[\protect\citeauthoryear{Chilimbi, Suzue, Apacible, and
  Kalyanaraman}{Chilimbi et~al\mbox{.}}{2014}]%
        {chilimbi2014project}
\bibfield{author}{\bibinfo{person}{Trishul~M Chilimbi}, \bibinfo{person}{Yutaka
  Suzue}, \bibinfo{person}{Johnson Apacible}, {and} \bibinfo{person}{Karthik
  Kalyanaraman}.} \bibinfo{year}{2014}\natexlab{}.
\newblock \showarticletitle{Project Adam: Building an Efficient and Scalable
  Deep Learning Training System}. In \bibinfo{booktitle}{\emph{OSDI}},
  Vol.~\bibinfo{volume}{14}. \bibinfo{pages}{571--582}.
\newblock


\bibitem[\protect\citeauthoryear{Chowdhury, Kumar, Imam, Jabbar, Sapra,
  Aggarwal, Hindle, and Greiner}{Chowdhury et~al\mbox{.}}{2015}]%
        {chowdhury2015system}
\bibfield{author}{\bibinfo{person}{Shaiful~Alam Chowdhury},
  \bibinfo{person}{Luke~N Kumar}, \bibinfo{person}{Md~Toukir Imam},
  \bibinfo{person}{Mohomed Shazan~Mohomed Jabbar}, \bibinfo{person}{Varun
  Sapra}, \bibinfo{person}{Karan Aggarwal}, \bibinfo{person}{Abram Hindle},
  {and} \bibinfo{person}{Russell Greiner}.} \bibinfo{year}{2015}\natexlab{}.
\newblock \showarticletitle{A system-call based model of software energy
  consumption without hardware instrumentation}. In
  \bibinfo{booktitle}{\emph{IGSC}}. \bibinfo{pages}{1--6}.
\newblock


\bibitem[\protect\citeauthoryear{Chu, Lane, Lai, Pang, Meng, Guo, Li, and
  Zhao}{Chu et~al\mbox{.}}{2011}]%
        {chu2011balancing}
\bibfield{author}{\bibinfo{person}{David Chu}, \bibinfo{person}{Nicholas~D
  Lane}, \bibinfo{person}{Ted Tsung-Te Lai}, \bibinfo{person}{Cong Pang},
  \bibinfo{person}{Xiangying Meng}, \bibinfo{person}{Qing Guo},
  \bibinfo{person}{Fan Li}, {and} \bibinfo{person}{Feng Zhao}.}
  \bibinfo{year}{2011}\natexlab{}.
\newblock \showarticletitle{Balancing energy, latency and accuracy for mobile
  sensor data classification}. In \bibinfo{booktitle}{\emph{SenSys}}. ACM,
  \bibinfo{pages}{54--67}.
\newblock


\bibitem[\protect\citeauthoryear{Cohen, Afshar, Tapson, and van Schaik}{Cohen
  et~al\mbox{.}}{2017}]%
        {cohen2017emnist}
\bibfield{author}{\bibinfo{person}{Gregory Cohen}, \bibinfo{person}{Saeed
  Afshar}, \bibinfo{person}{Jonathan Tapson}, {and} \bibinfo{person}{Andr{\'e}
  van Schaik}.} \bibinfo{year}{2017}\natexlab{}.
\newblock \showarticletitle{EMNIST: an extension of MNIST to handwritten
  letters}.
\newblock \bibinfo{journal}{\emph{arXiv preprint arXiv:1702.05373}}
  (\bibinfo{year}{2017}).
\newblock


\bibitem[\protect\citeauthoryear{Crammer, Dekel, Keshet, Shalev-Shwartz, and
  Singer}{Crammer et~al\mbox{.}}{2006}]%
        {crammer2006online}
\bibfield{author}{\bibinfo{person}{Koby Crammer}, \bibinfo{person}{Ofer Dekel},
  \bibinfo{person}{Joseph Keshet}, \bibinfo{person}{Shai Shalev-Shwartz}, {and}
  \bibinfo{person}{Yoram Singer}.} \bibinfo{year}{2006}\natexlab{}.
\newblock \showarticletitle{Online passive-aggressive algorithms}.
\newblock \bibinfo{journal}{\emph{JMLR}} \bibinfo{volume}{7},
  \bibinfo{number}{Mar} (\bibinfo{year}{2006}), \bibinfo{pages}{551--585}.
\newblock


\bibitem[\protect\citeauthoryear{Crushh}{Crushh}{2017}]%
        {messageLength}
\bibfield{author}{\bibinfo{person}{Crushh}.} \bibinfo{year}{2017}\natexlab{}.
\newblock \bibinfo{title}{{Average text message length}}.
\newblock
  \bibinfo{howpublished}{\url{https://crushhapp.com/blog/k-wrap-it-up-mom}}.
\newblock


\bibitem[\protect\citeauthoryear{Cuervo, Balasubramanian, Cho, Wolman, Saroiu,
  Chandra, and Bahl}{Cuervo et~al\mbox{.}}{2010}]%
        {cuervo2010maui}
\bibfield{author}{\bibinfo{person}{Eduardo Cuervo}, \bibinfo{person}{Aruna
  Balasubramanian}, \bibinfo{person}{Dae-ki Cho}, \bibinfo{person}{Alec
  Wolman}, \bibinfo{person}{Stefan Saroiu}, \bibinfo{person}{Ranveer Chandra},
  {and} \bibinfo{person}{Paramvir Bahl}.} \bibinfo{year}{2010}\natexlab{}.
\newblock \showarticletitle{MAUI: making smartphones last longer with code
  offload}. In \bibinfo{booktitle}{\emph{MobiSys}}. ACM,
  \bibinfo{pages}{49--62}.
\newblock


\bibitem[\protect\citeauthoryear{Cui, Cipar, Ho, Kim, Lee, Kumar, Wei, Dai,
  Ganger, Gibbons, et~al\mbox{.}}{Cui et~al\mbox{.}}{2014}]%
        {cui2014exploiting}
\bibfield{author}{\bibinfo{person}{Henggang Cui}, \bibinfo{person}{James
  Cipar}, \bibinfo{person}{Qirong Ho}, \bibinfo{person}{Jin~Kyu Kim},
  \bibinfo{person}{Seunghak Lee}, \bibinfo{person}{Abhimanu Kumar},
  \bibinfo{person}{Jinliang Wei}, \bibinfo{person}{Wei Dai},
  \bibinfo{person}{Gregory~R Ganger}, \bibinfo{person}{Phillip~B Gibbons},
  {et~al\mbox{.}}} \bibinfo{year}{2014}\natexlab{}.
\newblock \showarticletitle{Exploiting Bounded Staleness to Speed Up Big Data
  Analytics}. In \bibinfo{booktitle}{\emph{USENIX ATC}}.
  \bibinfo{pages}{37--48}.
\newblock


\bibitem[\protect\citeauthoryear{Damaskinos, El~Mhamdi, Guerraoui, Guirguis,
  and Rouault}{Damaskinos et~al\mbox{.}}{2019}]%
        {damaskinos2019aggregathor}
\bibfield{author}{\bibinfo{person}{Georgios Damaskinos},
  \bibinfo{person}{El~Mahdi El~Mhamdi}, \bibinfo{person}{Rachid Guerraoui},
  \bibinfo{person}{Arsany Guirguis}, {and} \bibinfo{person}{S{\'e}bastien
  Louis~Alexandre Rouault}.} \bibinfo{year}{2019}\natexlab{}.
\newblock \showarticletitle{Aggregathor: Byzantine machine learning via robust
  gradient aggregation}. In \bibinfo{booktitle}{\emph{Conference on Machine
  Learning and Systems (SysML / MLSys)}}.
\newblock


\bibitem[\protect\citeauthoryear{Damaskinos, El~Mhamdi, Guerraoui, Patra,
  Taziki, et~al\mbox{.}}{Damaskinos et~al\mbox{.}}{2018}]%
        {kardam}
\bibfield{author}{\bibinfo{person}{Georgios Damaskinos},
  \bibinfo{person}{El~Mahdi El~Mhamdi}, \bibinfo{person}{Rachid Guerraoui},
  \bibinfo{person}{Rhicheek Patra}, \bibinfo{person}{Mahsa Taziki},
  {et~al\mbox{.}}} \bibinfo{year}{2018}\natexlab{}.
\newblock \showarticletitle{Asynchronous Byzantine Machine Learning (the case
  of SGD)}. In \bibinfo{booktitle}{\emph{ICML}}. \bibinfo{pages}{1153--1162}.
\newblock


\bibitem[\protect\citeauthoryear{Deeplearning4j}{Deeplearning4j}{2020}]%
        {dl4j}
\bibfield{author}{\bibinfo{person}{Deeplearning4j}.}
  \bibinfo{year}{2020}\natexlab{}.
\newblock \bibinfo{title}{DL4J}.
\newblock \bibinfo{howpublished}{\url{https://deeplearning4j.org/}}.
\newblock


\bibitem[\protect\citeauthoryear{Dhingra, Zhou, Fitzpatrick, Muehl, and
  Cohen}{Dhingra et~al\mbox{.}}{2016}]%
        {dhingra2016tweet2vec}
\bibfield{author}{\bibinfo{person}{Bhuwan Dhingra}, \bibinfo{person}{Zhong
  Zhou}, \bibinfo{person}{Dylan Fitzpatrick}, \bibinfo{person}{Michael Muehl},
  {and} \bibinfo{person}{William~W Cohen}.} \bibinfo{year}{2016}\natexlab{}.
\newblock \showarticletitle{Tweet2vec: Character-based distributed
  representations for social media}.
\newblock \bibinfo{journal}{\emph{arXiv preprint arXiv:1605.03481}}
  (\bibinfo{year}{2016}).
\newblock


\bibitem[\protect\citeauthoryear{Ding, Mishra, and Hoffmann}{Ding
  et~al\mbox{.}}{2019}]%
        {ding2019}
\bibfield{author}{\bibinfo{person}{Yi Ding}, \bibinfo{person}{Nikita Mishra},
  {and} \bibinfo{person}{Henry Hoffmann}.} \bibinfo{year}{2019}\natexlab{}.
\newblock \showarticletitle{Generative and Multi-Phase Learning for Computer
  Systems Optimization}. In \bibinfo{booktitle}{\emph{ISCA}} (Phoenix, Arizona)
  \emph{(\bibinfo{series}{ISCA ’19})}. \bibinfo{publisher}{Association for
  Computing Machinery}, \bibinfo{address}{New York, NY, USA},
  \bibinfo{pages}{39–52}.
\newblock
\showISBNx{9781450366694}
\urldef\tempurl%
\url{https://doi.org/10.1145/3307650.3326633}
\showDOI{\tempurl}


\bibitem[\protect\citeauthoryear{Dutta, Cadambe, and Grover}{Dutta
  et~al\mbox{.}}{2016}]%
        {dutta2016short}
\bibfield{author}{\bibinfo{person}{Sanghamitra Dutta}, \bibinfo{person}{Viveck
  Cadambe}, {and} \bibinfo{person}{Pulkit Grover}.}
  \bibinfo{year}{2016}\natexlab{}.
\newblock \showarticletitle{Short-dot: Computing large linear transforms
  distributedly using coded short dot products}. In
  \bibinfo{booktitle}{\emph{NIPS}}. \bibinfo{pages}{2100--2108}.
\newblock


\bibitem[\protect\citeauthoryear{Dwork, Roth, et~al\mbox{.}}{Dwork
  et~al\mbox{.}}{2014}]%
        {dwork2014algorithmic}
\bibfield{author}{\bibinfo{person}{Cynthia Dwork}, \bibinfo{person}{Aaron
  Roth}, {et~al\mbox{.}}} \bibinfo{year}{2014}\natexlab{}.
\newblock \showarticletitle{The algorithmic foundations of differential
  privacy}.
\newblock \bibinfo{journal}{\emph{Foundations and Trends{\textregistered} in
  Theoretical Computer Science}} \bibinfo{volume}{9}, \bibinfo{number}{3--4}
  (\bibinfo{year}{2014}), \bibinfo{pages}{211--407}.
\newblock


\bibitem[\protect\citeauthoryear{EsotericSoftware}{EsotericSoftware}{2020}]%
        {kryo}
\bibfield{author}{\bibinfo{person}{EsotericSoftware}.}
  \bibinfo{year}{2020}\natexlab{}.
\newblock \bibinfo{title}{Kryo}.
\newblock
  \bibinfo{howpublished}{\url{https://github.com/EsotericSoftware/kryo/}}.
\newblock


\bibitem[\protect\citeauthoryear{Gong and Zhang}{Gong and Zhang}{2016}]%
        {gong2016hashtag}
\bibfield{author}{\bibinfo{person}{Yuyun Gong} {and} \bibinfo{person}{Qi
  Zhang}.} \bibinfo{year}{2016}\natexlab{}.
\newblock \showarticletitle{Hashtag Recommendation Using Attention-Based
  Convolutional Neural Network}. In \bibinfo{booktitle}{\emph{IJCAI}}.
  \bibinfo{pages}{2782--2788}.
\newblock


\bibitem[\protect\citeauthoryear{Google}{Google}{2020a}]%
        {tffl}
\bibfield{author}{\bibinfo{person}{Google}.} \bibinfo{year}{2020}\natexlab{a}.
\newblock \bibinfo{title}{TensorFlow - Federated Learning}.
\newblock
  \bibinfo{howpublished}{\url{https://www.tensorflow.org/federated/federated_learning}}.
\newblock


\bibitem[\protect\citeauthoryear{Google}{Google}{2020b}]%
        {tfTextClassification}
\bibfield{author}{\bibinfo{person}{Google}.} \bibinfo{year}{2020}\natexlab{b}.
\newblock \bibinfo{title}{Tensorflow text classification}.
\newblock
\newblock
\urldef\tempurl%
\url{https://www.tensorflow.org/tutorials/text/text_classification_rnn#create_the_model}
\showURL{%
\tempurl}


\bibitem[\protect\citeauthoryear{government}{government}{2018}]%
        {ccpa}
\bibfield{author}{\bibinfo{person}{US government}.}
  \bibinfo{year}{2018}\natexlab{}.
\newblock \bibinfo{title}{{California Consumer Privacy Act of 2018 (CCPA)}}.
\newblock
  \bibinfo{howpublished}{\url{https://leginfo.legislature.ca.gov/faces/billTextClient.xhtml?bill_id=201720180AB375}}.
\newblock


\bibitem[\protect\citeauthoryear{Greenhalgh}{Greenhalgh}{2013}]%
        {greenhalgh2013big}
\bibfield{author}{\bibinfo{person}{P Greenhalgh}.}
  \bibinfo{year}{2013}\natexlab{}.
\newblock \showarticletitle{big. LITTLE Technology: The Future of Mobile}.
\newblock \bibinfo{journal}{\emph{ARM, White paper}} (\bibinfo{year}{2013}).
\newblock


\bibitem[\protect\citeauthoryear{Grid5000}{Grid5000}{2020}]%
        {g5k}
\bibfield{author}{\bibinfo{person}{Grid5000}.} \bibinfo{year}{2020}\natexlab{}.
\newblock \bibinfo{title}{Grid5000}.
\newblock \bibinfo{howpublished}{\url{https://www.grid5000.fr/}}.
\newblock


\bibitem[\protect\citeauthoryear{Halpern, Zhu, and Reddi}{Halpern
  et~al\mbox{.}}{2016}]%
        {halpern2016mobile}
\bibfield{author}{\bibinfo{person}{Matthew Halpern}, \bibinfo{person}{Yuhao
  Zhu}, {and} \bibinfo{person}{Vijay~Janapa Reddi}.}
  \bibinfo{year}{2016}\natexlab{}.
\newblock \showarticletitle{Mobile cpu's rise to power: Quantifying the impact
  of generational mobile cpu design trends on performance, energy, and user
  satisfaction}. In \bibinfo{booktitle}{\emph{HPCA}}. IEEE,
  \bibinfo{pages}{64--76}.
\newblock


\bibitem[\protect\citeauthoryear{Hao, Li, Halfond, and Govindan}{Hao
  et~al\mbox{.}}{2013}]%
        {hao2013estimating}
\bibfield{author}{\bibinfo{person}{Shuai Hao}, \bibinfo{person}{Ding Li},
  \bibinfo{person}{William~GJ Halfond}, {and} \bibinfo{person}{Ramesh
  Govindan}.} \bibinfo{year}{2013}\natexlab{}.
\newblock \showarticletitle{Estimating mobile application energy consumption
  using program analysis}. In \bibinfo{booktitle}{\emph{ICSE}}. IEEE Press,
  \bibinfo{pages}{92--101}.
\newblock


\bibitem[\protect\citeauthoryear{Hard, Rao, Mathews, Ramaswamy, Beaufays,
  Augenstein, Eichner, Kiddon, and Ramage}{Hard et~al\mbox{.}}{2018}]%
        {hard2018federated}
\bibfield{author}{\bibinfo{person}{Andrew Hard}, \bibinfo{person}{Kanishka
  Rao}, \bibinfo{person}{Rajiv Mathews}, \bibinfo{person}{Swaroop Ramaswamy},
  \bibinfo{person}{Fran{\c{c}}oise Beaufays}, \bibinfo{person}{Sean
  Augenstein}, \bibinfo{person}{Hubert Eichner}, \bibinfo{person}{Chlo{\'e}
  Kiddon}, {and} \bibinfo{person}{Daniel Ramage}.}
  \bibinfo{year}{2018}\natexlab{}.
\newblock \showarticletitle{Federated learning for mobile keyboard prediction}.
\newblock \bibinfo{journal}{\emph{arXiv preprint arXiv:1811.03604}}
  (\bibinfo{year}{2018}).
\newblock


\bibitem[\protect\citeauthoryear{How fast is 4G?}{How fast is 4G?}{2020}]%
        {4gspeed}
How fast is 4G? \bibinfo{year}{2020}\natexlab{}.
\newblock \bibinfo{howpublished}{\url{https://www.4g.co.uk/how-fast-is-4g/}}.
\newblock


\bibitem[\protect\citeauthoryear{Jeong, Oh, Kim, Park, Bennis, and Kim}{Jeong
  et~al\mbox{.}}{2018}]%
        {jeong2018communication}
\bibfield{author}{\bibinfo{person}{Eunjeong Jeong}, \bibinfo{person}{Seungeun
  Oh}, \bibinfo{person}{Hyesung Kim}, \bibinfo{person}{Jihong Park},
  \bibinfo{person}{Mehdi Bennis}, {and} \bibinfo{person}{Seong-Lyun Kim}.}
  \bibinfo{year}{2018}\natexlab{}.
\newblock \showarticletitle{Communication-Efficient On-Device Machine Learning:
  Federated Distillation and Augmentation under Non-IID Private Data}.
\newblock \bibinfo{journal}{\emph{arXiv preprint arXiv:1811.11479}}
  (\bibinfo{year}{2018}).
\newblock


\bibitem[\protect\citeauthoryear{Jiang, Cui, Zhang, and Yu}{Jiang
  et~al\mbox{.}}{2017}]%
        {jiang2017heterogeneity}
\bibfield{author}{\bibinfo{person}{Jiawei Jiang}, \bibinfo{person}{Bin Cui},
  \bibinfo{person}{Ce Zhang}, {and} \bibinfo{person}{Lele Yu}.}
  \bibinfo{year}{2017}\natexlab{}.
\newblock \showarticletitle{Heterogeneity-aware Distributed Parameter Servers}.
  In \bibinfo{booktitle}{\emph{SIGMOD}}. \bibinfo{pages}{463--478}.
\newblock


\bibitem[\protect\citeauthoryear{Kang, Hauswald, Gao, Rovinski, Mudge, Mars,
  and Tang}{Kang et~al\mbox{.}}{2017}]%
        {kang2017neurosurgeon}
\bibfield{author}{\bibinfo{person}{Yiping Kang}, \bibinfo{person}{Johann
  Hauswald}, \bibinfo{person}{Cao Gao}, \bibinfo{person}{Austin Rovinski},
  \bibinfo{person}{Trevor Mudge}, \bibinfo{person}{Jason Mars}, {and}
  \bibinfo{person}{Lingjia Tang}.} \bibinfo{year}{2017}\natexlab{}.
\newblock \showarticletitle{Neurosurgeon: Collaborative intelligence between
  the cloud and mobile edge}. In \bibinfo{booktitle}{\emph{ASPLOS}}.
  \bibinfo{pages}{615--629}.
\newblock


\bibitem[\protect\citeauthoryear{Kone{\v{c}}n{\`y}, McMahan, Ramage, and
  Richt{\'a}rik}{Kone{\v{c}}n{\`y} et~al\mbox{.}}{2016}]%
        {konevcny2016federated}
\bibfield{author}{\bibinfo{person}{Jakub Kone{\v{c}}n{\`y}},
  \bibinfo{person}{H~Brendan McMahan}, \bibinfo{person}{Daniel Ramage}, {and}
  \bibinfo{person}{Peter Richt{\'a}rik}.} \bibinfo{year}{2016}\natexlab{}.
\newblock \showarticletitle{Federated optimization: distributed machine
  learning for on-device intelligence}.
\newblock \bibinfo{journal}{\emph{arXiv preprint arXiv:1610.02527}}
  (\bibinfo{year}{2016}).
\newblock


\bibitem[\protect\citeauthoryear{Kowald, Pujari, and Lex}{Kowald
  et~al\mbox{.}}{2017}]%
        {kowald2017temporal}
\bibfield{author}{\bibinfo{person}{Dominik Kowald},
  \bibinfo{person}{Subhash~Chandra Pujari}, {and} \bibinfo{person}{Elisabeth
  Lex}.} \bibinfo{year}{2017}\natexlab{}.
\newblock \showarticletitle{Temporal effects on hashtag reuse in twitter: A
  cognitive-inspired hashtag recommendation approach}. In
  \bibinfo{booktitle}{\emph{WWW}}. \bibinfo{pages}{1401--1410}.
\newblock


\bibitem[\protect\citeauthoryear{Krizhevsky}{Krizhevsky}{2009}]%
        {cifar}
\bibfield{author}{\bibinfo{person}{Alex Krizhevsky}.}
  \bibinfo{year}{2009}\natexlab{}.
\newblock \bibinfo{title}{Cifar dataset}.
\newblock
  \bibinfo{howpublished}{{\url{https://www.cs.toronto.edu/~kriz/cifar.html}}}.
\newblock


\bibitem[\protect\citeauthoryear{Kwon, Lee, Yi, Kwon, Yang, Chun, Huang,
  Maniatis, Naik, and Paek}{Kwon et~al\mbox{.}}{2013}]%
        {kwon2013mantis}
\bibfield{author}{\bibinfo{person}{Yongin Kwon}, \bibinfo{person}{Sangmin Lee},
  \bibinfo{person}{Hayoon Yi}, \bibinfo{person}{Donghyun Kwon},
  \bibinfo{person}{Seungjun Yang}, \bibinfo{person}{Byung-Gon Chun},
  \bibinfo{person}{Ling Huang}, \bibinfo{person}{Petros Maniatis},
  \bibinfo{person}{Mayur Naik}, {and} \bibinfo{person}{Yunheung Paek}.}
  \bibinfo{year}{2013}\natexlab{}.
\newblock \showarticletitle{Mantis: Automatic performance prediction for
  smartphone applications}. In \bibinfo{booktitle}{\emph{USENIX ATC}}.
  \bibinfo{pages}{297--308}.
\newblock


\bibitem[\protect\citeauthoryear{Kywe, Hoang, Lim, and Zhu}{Kywe
  et~al\mbox{.}}{2012}]%
        {kywe2012recommending}
\bibfield{author}{\bibinfo{person}{Su~Mon Kywe}, \bibinfo{person}{Tuan-Anh
  Hoang}, \bibinfo{person}{Ee-Peng Lim}, {and} \bibinfo{person}{Feida Zhu}.}
  \bibinfo{year}{2012}\natexlab{}.
\newblock \showarticletitle{On recommending hashtags in twitter networks}. In
  \bibinfo{booktitle}{\emph{International Conference on Social Informatics}}.
  Springer, \bibinfo{pages}{337--350}.
\newblock


\bibitem[\protect\citeauthoryear{Labs}{Labs}{2020}]%
        {matrixBench}
\bibfield{author}{\bibinfo{person}{Primate Labs}.}
  \bibinfo{year}{2020}\natexlab{}.
\newblock \bibinfo{title}{Matrix multiplication benchmark}.
\newblock \bibinfo{howpublished}{\url{https://browser.geekbench.com}}.
\newblock


\bibitem[\protect\citeauthoryear{Lecun}{Lecun}{1998}]%
        {mnist}
\bibfield{author}{\bibinfo{person}{Yann Lecun}.}
  \bibinfo{year}{1998}\natexlab{}.
\newblock \bibinfo{title}{MNIST dataset}.
\newblock \bibinfo{howpublished}{{\url{http://yann.lecun.com/exdb/mnist/}}}.
\newblock


\bibitem[\protect\citeauthoryear{Lee, Lam, Pedarsani, Papailiopoulos, and
  Ramchandran}{Lee et~al\mbox{.}}{2017}]%
        {lee2017speeding}
\bibfield{author}{\bibinfo{person}{Kangwook Lee}, \bibinfo{person}{Maximilian
  Lam}, \bibinfo{person}{Ramtin Pedarsani}, \bibinfo{person}{Dimitris
  Papailiopoulos}, {and} \bibinfo{person}{Kannan Ramchandran}.}
  \bibinfo{year}{2017}\natexlab{}.
\newblock \showarticletitle{Speeding up distributed machine learning using
  codes}.
\newblock \bibinfo{journal}{\emph{IEEE Transactions on Information Theory}}
  \bibinfo{volume}{64}, \bibinfo{number}{3} (\bibinfo{year}{2017}),
  \bibinfo{pages}{1514--1529}.
\newblock


\bibitem[\protect\citeauthoryear{Li, Andersen, Park, Smola, Ahmed, Josifovski,
  Long, Shekita, and Su}{Li et~al\mbox{.}}{2014}]%
        {li2014scaling}
\bibfield{author}{\bibinfo{person}{Mu Li}, \bibinfo{person}{David~G Andersen},
  \bibinfo{person}{Jun~Woo Park}, \bibinfo{person}{Alexander~J Smola},
  \bibinfo{person}{Amr Ahmed}, \bibinfo{person}{Vanja Josifovski},
  \bibinfo{person}{James Long}, \bibinfo{person}{Eugene~J Shekita}, {and}
  \bibinfo{person}{Bor-Yiing Su}.} \bibinfo{year}{2014}\natexlab{}.
\newblock \showarticletitle{Scaling distributed machine learning with the
  parameter server}. In \bibinfo{booktitle}{\emph{OSDI}}.
  \bibinfo{pages}{583--598}.
\newblock


\bibitem[\protect\citeauthoryear{Li, Sahu, Zaheer, Sanjabi, Talwalkar, and
  Smith}{Li et~al\mbox{.}}{2018}]%
        {li2018federated}
\bibfield{author}{\bibinfo{person}{Tian Li}, \bibinfo{person}{Anit~Kumar Sahu},
  \bibinfo{person}{Manzil Zaheer}, \bibinfo{person}{Maziar Sanjabi},
  \bibinfo{person}{Ameet Talwalkar}, {and} \bibinfo{person}{Virginia Smith}.}
  \bibinfo{year}{2018}\natexlab{}.
\newblock \showarticletitle{Federated optimization for heterogeneous networks}.
\newblock \bibinfo{journal}{\emph{arXiv preprint arXiv:1812.06127}}
  (\bibinfo{year}{2018}).
\newblock


\bibitem[\protect\citeauthoryear{Liu and Lee}{Liu and Lee}{2015}]%
        {liu2015empirical}
\bibfield{author}{\bibinfo{person}{Yan Liu} {and} \bibinfo{person}{Jack~YB
  Lee}.} \bibinfo{year}{2015}\natexlab{}.
\newblock \showarticletitle{An empirical study of throughput prediction in
  mobile data networks}. In \bibinfo{booktitle}{\emph{GLOBECOM}}. IEEE,
  \bibinfo{pages}{1--6}.
\newblock


\bibitem[\protect\citeauthoryear{McMahan, Moore, Ramage, Hampson, and
  y~Arcas}{McMahan et~al\mbox{.}}{2017}]%
        {mcmahan2017communication}
\bibfield{author}{\bibinfo{person}{Brendan McMahan}, \bibinfo{person}{Eider
  Moore}, \bibinfo{person}{Daniel Ramage}, \bibinfo{person}{Seth Hampson},
  {and} \bibinfo{person}{Blaise~Aguera y Arcas}.}
  \bibinfo{year}{2017}\natexlab{}.
\newblock \showarticletitle{Communication-Efficient Learning of Deep Networks
  from Decentralized Data}. In \bibinfo{booktitle}{\emph{AISTATS}}.
  \bibinfo{pages}{1273--1282}.
\newblock


\bibitem[\protect\citeauthoryear{media}{media}{2013}]%
        {prism}
\bibfield{author}{\bibinfo{person}{Vox media}.}
  \bibinfo{year}{2013}\natexlab{}.
\newblock \bibinfo{title}{{NSA's PRISM}}.
\newblock
  \bibinfo{howpublished}{\url{https://www.theverge.com/2013/7/17/4517480/nsa-spying-prism-surveillance-cheat-sheet}}.
\newblock


\bibitem[\protect\citeauthoryear{media}{media}{2018}]%
        {fbca}
\bibfield{author}{\bibinfo{person}{Vox media}.}
  \bibinfo{year}{2018}\natexlab{}.
\newblock \bibinfo{title}{{The Facebook and Cambridge Analytica scandal}}.
\newblock
  \bibinfo{howpublished}{\url{https://www.vox.com/policy-and-politics/2018/3/23/17151916/facebook-cambridge-analytica-trump-diagram}}.
\newblock


\bibitem[\protect\citeauthoryear{Mishne, Dalton, Li, Sharma, and Lin}{Mishne
  et~al\mbox{.}}{2013}]%
        {mishne2013fast}
\bibfield{author}{\bibinfo{person}{Gilad Mishne}, \bibinfo{person}{Jeff
  Dalton}, \bibinfo{person}{Zhenghua Li}, \bibinfo{person}{Aneesh Sharma},
  {and} \bibinfo{person}{Jimmy Lin}.} \bibinfo{year}{2013}\natexlab{}.
\newblock \showarticletitle{Fast data in the era of big data: Twitter's
  real-time related query suggestion architecture}. In
  \bibinfo{booktitle}{\emph{SIGMOD}}. ACM, \bibinfo{pages}{1147--1158}.
\newblock


\bibitem[\protect\citeauthoryear{Mishra, Imes, Lafferty, and Hoffmann}{Mishra
  et~al\mbox{.}}{2018}]%
        {mishra2018caloree}
\bibfield{author}{\bibinfo{person}{Nikita Mishra}, \bibinfo{person}{Connor
  Imes}, \bibinfo{person}{John~D Lafferty}, {and} \bibinfo{person}{Henry
  Hoffmann}.} \bibinfo{year}{2018}\natexlab{}.
\newblock \showarticletitle{CALOREE: Learning control for predictable latency
  and low energy}.
\newblock \bibinfo{journal}{\emph{ASPLOS}} \bibinfo{volume}{53},
  \bibinfo{number}{2} (\bibinfo{year}{2018}), \bibinfo{pages}{184--198}.
\newblock


\bibitem[\protect\citeauthoryear{Mishra, Zhang, Lafferty, and Hoffmann}{Mishra
  et~al\mbox{.}}{2015}]%
        {mishra2015probabilistic}
\bibfield{author}{\bibinfo{person}{Nikita Mishra}, \bibinfo{person}{Huazhe
  Zhang}, \bibinfo{person}{John~D Lafferty}, {and} \bibinfo{person}{Henry
  Hoffmann}.} \bibinfo{year}{2015}\natexlab{}.
\newblock \showarticletitle{A probabilistic graphical model-based approach for
  minimizing energy under performance constraints}.
\newblock \bibinfo{journal}{\emph{ASPLOS}} \bibinfo{volume}{50},
  \bibinfo{number}{4} (\bibinfo{year}{2015}), \bibinfo{pages}{267--281}.
\newblock


\bibitem[\protect\citeauthoryear{Mitliagkas, Zhang, Hadjis, and
  R{\'e}}{Mitliagkas et~al\mbox{.}}{2016}]%
        {mitliagkas2016asynchrony}
\bibfield{author}{\bibinfo{person}{Ioannis Mitliagkas}, \bibinfo{person}{Ce
  Zhang}, \bibinfo{person}{Stefan Hadjis}, {and} \bibinfo{person}{Christopher
  R{\'e}}.} \bibinfo{year}{2016}\natexlab{}.
\newblock \showarticletitle{Asynchrony begets momentum, with an application to
  deep learning}. In \bibinfo{booktitle}{\emph{Annual Allerton Conference on
  Communication, Control, and Computing}}. IEEE, \bibinfo{pages}{997--1004}.
\newblock


\bibitem[\protect\citeauthoryear{Mittal, Kansal, and Chandra}{Mittal
  et~al\mbox{.}}{2012}]%
        {mittal2012empowering}
\bibfield{author}{\bibinfo{person}{Radhika Mittal}, \bibinfo{person}{Aman
  Kansal}, {and} \bibinfo{person}{Ranveer Chandra}.}
  \bibinfo{year}{2012}\natexlab{}.
\newblock \showarticletitle{Empowering developers to estimate app energy
  consumption}. In \bibinfo{booktitle}{\emph{MobiCom}}. ACM,
  \bibinfo{pages}{317--328}.
\newblock


\bibitem[\protect\citeauthoryear{Neyshabur, Salakhutdinov, and
  Srebro}{Neyshabur et~al\mbox{.}}{2015}]%
        {neyshabur2015path}
\bibfield{author}{\bibinfo{person}{Behnam Neyshabur}, \bibinfo{person}{Ruslan~R
  Salakhutdinov}, {and} \bibinfo{person}{Nati Srebro}.}
  \bibinfo{year}{2015}\natexlab{}.
\newblock \showarticletitle{Path-sgd: Path-normalized optimization in deep
  neural networks}. In \bibinfo{booktitle}{\emph{NIPS}}.
  \bibinfo{pages}{2422--2430}.
\newblock


\bibitem[\protect\citeauthoryear{Nishio and Yonetani}{Nishio and
  Yonetani}{2019}]%
        {nishio2019client}
\bibfield{author}{\bibinfo{person}{Takayuki Nishio} {and} \bibinfo{person}{Ryo
  Yonetani}.} \bibinfo{year}{2019}\natexlab{}.
\newblock \showarticletitle{Client selection for federated learning with
  heterogeneous resources in mobile edge}. In \bibinfo{booktitle}{\emph{ICC}}.
  IEEE, \bibinfo{pages}{1--7}.
\newblock


\bibitem[\protect\citeauthoryear{{NVIDIA}}{{NVIDIA}}{2020}]%
        {nvidia2020}
\bibfield{author}{\bibinfo{person}{{NVIDIA}}.} \bibinfo{year}{2020}\natexlab{}.
\newblock \bibinfo{title}{{CUDA GPUs | NVIDIA Developer}}.
\newblock \bibinfo{howpublished}{\url{https://developer.nvidia.com/cuda-gpus}}.
\newblock


\bibitem[\protect\citeauthoryear{Otsuka, Wallace, and Chiu}{Otsuka
  et~al\mbox{.}}{2014}]%
        {otsuka2014design}
\bibfield{author}{\bibinfo{person}{Eriko Otsuka}, \bibinfo{person}{Scott~A
  Wallace}, {and} \bibinfo{person}{David Chiu}.}
  \bibinfo{year}{2014}\natexlab{}.
\newblock \showarticletitle{Design and evaluation of a twitter hashtag
  recommendation system}. In \bibinfo{booktitle}{\emph{IDEAS}}.
  \bibinfo{pages}{330--333}.
\newblock


\bibitem[\protect\citeauthoryear{Ouyang, Garraghan, McKee, Townend, and
  Xu}{Ouyang et~al\mbox{.}}{2016}]%
        {ouyang2016straggler}
\bibfield{author}{\bibinfo{person}{Xue Ouyang}, \bibinfo{person}{Peter
  Garraghan}, \bibinfo{person}{David McKee}, \bibinfo{person}{Paul Townend},
  {and} \bibinfo{person}{Jie Xu}.} \bibinfo{year}{2016}\natexlab{}.
\newblock \showarticletitle{Straggler detection in parallel computing systems
  through dynamic threshold calculation}. In \bibinfo{booktitle}{\emph{AINA}}.
  IEEE, \bibinfo{pages}{414--421}.
\newblock


\bibitem[\protect\citeauthoryear{Phan, Pallez, Ibrahim, and Raghavan}{Phan
  et~al\mbox{.}}{2019}]%
        {phan2019new}
\bibfield{author}{\bibinfo{person}{Tien-Dat Phan}, \bibinfo{person}{Guillaume
  Pallez}, \bibinfo{person}{Shadi Ibrahim}, {and} \bibinfo{person}{Padma
  Raghavan}.} \bibinfo{year}{2019}\natexlab{}.
\newblock \showarticletitle{A new framework for evaluating straggler detection
  mechanisms in mapreduce}.
\newblock \bibinfo{journal}{\emph{TOMPECS}} \bibinfo{volume}{4},
  \bibinfo{number}{3} (\bibinfo{year}{2019}), \bibinfo{pages}{1--23}.
\newblock


\bibitem[\protect\citeauthoryear{Qian, Wang, Gerber, Mao, Sen, and
  Spatscheck}{Qian et~al\mbox{.}}{2011}]%
        {qian2011profiling}
\bibfield{author}{\bibinfo{person}{Feng Qian}, \bibinfo{person}{Zhaoguang
  Wang}, \bibinfo{person}{Alexandre Gerber}, \bibinfo{person}{Zhuoqing Mao},
  \bibinfo{person}{Subhabrata Sen}, {and} \bibinfo{person}{Oliver Spatscheck}.}
  \bibinfo{year}{2011}\natexlab{}.
\newblock \showarticletitle{Profiling resource usage for mobile applications: a
  cross-layer approach}. In \bibinfo{booktitle}{\emph{MobiSys}}. ACM,
  \bibinfo{pages}{321--334}.
\newblock


\bibitem[\protect\citeauthoryear{Qiao, Aghayev, Yu, Chen, Ho, Gibson, and
  Xing}{Qiao et~al\mbox{.}}{2018}]%
        {qiao2018litz}
\bibfield{author}{\bibinfo{person}{Aurick Qiao}, \bibinfo{person}{Abutalib
  Aghayev}, \bibinfo{person}{Weiren Yu}, \bibinfo{person}{Haoyang Chen},
  \bibinfo{person}{Qirong Ho}, \bibinfo{person}{Garth~A Gibson}, {and}
  \bibinfo{person}{Eric~P Xing}.} \bibinfo{year}{2018}\natexlab{}.
\newblock \showarticletitle{Litz: Elastic framework for high-performance
  distributed machine learning}. In \bibinfo{booktitle}{\emph{USENIX ATC}}.
  \bibinfo{pages}{631--644}.
\newblock


\bibitem[\protect\citeauthoryear{Qualcomm}{Qualcomm}{2020}]%
        {adreno2020}
\bibfield{author}{\bibinfo{person}{Qualcomm}.} \bibinfo{year}{2020}\natexlab{}.
\newblock \bibinfo{title}{{Adreno™ Graphics Processing Units}}.
\newblock
  \bibinfo{howpublished}{\url{https://developer.qualcomm.com/software/adreno-gpu-sdk/gpu}}.
\newblock


\bibitem[\protect\citeauthoryear{request}{request}{2016}]%
        {messagesPerDay}
\bibfield{author}{\bibinfo{person}{Text request}.}
  \bibinfo{year}{2016}\natexlab{}.
\newblock \bibinfo{title}{{Average text messages per day}}.
\newblock
  \bibinfo{howpublished}{\url{https://www.textrequest.com/blog/how-many-texts-people-send-per-day/}}.
\newblock


\bibitem[\protect\citeauthoryear{Ryffel, Trask, Dahl, Wagner, Mancuso,
  Rueckert, and Passerat-Palmbach}{Ryffel et~al\mbox{.}}{2018}]%
        {ryffel2018generic}
\bibfield{author}{\bibinfo{person}{Theo Ryffel}, \bibinfo{person}{Andrew
  Trask}, \bibinfo{person}{Morten Dahl}, \bibinfo{person}{Bobby Wagner},
  \bibinfo{person}{Jason Mancuso}, \bibinfo{person}{Daniel Rueckert}, {and}
  \bibinfo{person}{Jonathan Passerat-Palmbach}.}
  \bibinfo{year}{2018}\natexlab{}.
\newblock \showarticletitle{A generic framework for privacy preserving deep
  learning}.
\newblock \bibinfo{journal}{\emph{arXiv preprint arXiv:1811.04017}}
  (\bibinfo{year}{2018}).
\newblock


\bibitem[\protect\citeauthoryear{S20.ai}{S20.ai}{2020}]%
        {s20ai}
\bibfield{author}{\bibinfo{person}{S20.ai}.} \bibinfo{year}{2020}\natexlab{}.
\newblock \bibinfo{title}{S20.ai}.
\newblock \bibinfo{howpublished}{\url{https://www.s20.ai/}}.
\newblock


\bibitem[\protect\citeauthoryear{Smith, Chiang, Sanjabi, and Talwalkar}{Smith
  et~al\mbox{.}}{2017}]%
        {smith2017federated}
\bibfield{author}{\bibinfo{person}{Virginia Smith}, \bibinfo{person}{Chao-Kai
  Chiang}, \bibinfo{person}{Maziar Sanjabi}, {and} \bibinfo{person}{Ameet~S
  Talwalkar}.} \bibinfo{year}{2017}\natexlab{}.
\newblock \showarticletitle{Federated multi-task learning}. In
  \bibinfo{booktitle}{\emph{NIPS}}. \bibinfo{pages}{4424--4434}.
\newblock


\bibitem[\protect\citeauthoryear{Snips}{Snips}{2020}]%
        {snips}
\bibfield{author}{\bibinfo{person}{Snips}.} \bibinfo{year}{2020}\natexlab{}.
\newblock \bibinfo{title}{Snips - Using Voice to Make Technology Disappear}.
\newblock \bibinfo{howpublished}{\url{https://snips.ai/}}.
\newblock


\bibitem[\protect\citeauthoryear{Statista}{Statista}{2018}]%
        {traffic}
\bibfield{author}{\bibinfo{person}{Statista}.} \bibinfo{year}{2018}\natexlab{}.
\newblock \bibinfo{title}{Percentage of all global web pages served to mobile
  phones from 2009 to 2018}.
\newblock
  \bibinfo{howpublished}{\url{https://www.statista.com/statistics/241462/global-mobile-phone-website-traffic-share/}}.
\newblock


\bibitem[\protect\citeauthoryear{Tweepy}{Tweepy}{2020}]%
        {tweepy}
Tweepy \bibinfo{year}{2020}\natexlab{}.
\newblock
  \bibinfo{howpublished}{\url{https://tweepy.readthedocs.io/en/latest/}}.
\newblock


\bibitem[\protect\citeauthoryear{Union}{Union}{2016}]%
        {gdpr}
\bibfield{author}{\bibinfo{person}{European Union}.}
  \bibinfo{year}{2016}\natexlab{}.
\newblock \bibinfo{title}{{Regulation 2016/679 of the European Parliament and
  of the Council of 27 April 2016 on the protection of natural persons with
  regard to the processing of personal data and on the free movement of such
  data, and repealing Directive 95/46/EC (GDPR)}}.
\newblock
  \bibinfo{howpublished}{\url{https://eur-lex.europa.eu/legal-content/EN/TXT/PDF/?uri=CELEX:32016R0679}}.
\newblock


\bibitem[\protect\citeauthoryear{Wang, Tuor, Salonidis, Leung, Makaya, He, and
  Chan}{Wang et~al\mbox{.}}{2019b}]%
        {wang2019adaptive}
\bibfield{author}{\bibinfo{person}{Shiqiang Wang}, \bibinfo{person}{Tiffany
  Tuor}, \bibinfo{person}{Theodoros Salonidis}, \bibinfo{person}{Kin~K Leung},
  \bibinfo{person}{Christian Makaya}, \bibinfo{person}{Ting He}, {and}
  \bibinfo{person}{Kevin Chan}.} \bibinfo{year}{2019}\natexlab{b}.
\newblock \showarticletitle{Adaptive federated learning in resource constrained
  edge computing systems}.
\newblock \bibinfo{journal}{\emph{IEEE Journal on Selected Areas in
  Communications}} \bibinfo{volume}{37}, \bibinfo{number}{6}
  (\bibinfo{year}{2019}), \bibinfo{pages}{1205--1221}.
\newblock


\bibitem[\protect\citeauthoryear{Wang, Song, Zhang, Song, Wang, and Qi}{Wang
  et~al\mbox{.}}{2019a}]%
        {wang2019beyond}
\bibfield{author}{\bibinfo{person}{Zhibo Wang}, \bibinfo{person}{Mengkai Song},
  \bibinfo{person}{Zhifei Zhang}, \bibinfo{person}{Yang Song},
  \bibinfo{person}{Qian Wang}, {and} \bibinfo{person}{Hairong Qi}.}
  \bibinfo{year}{2019}\natexlab{a}.
\newblock \showarticletitle{Beyond inferring class representatives: User-level
  privacy leakage from federated learning}. In
  \bibinfo{booktitle}{\emph{INFOCOM}}. IEEE, \bibinfo{pages}{2512--2520}.
\newblock


\bibitem[\protect\citeauthoryear{Wikipedia}{Wikipedia}{2019}]%
        {bhattacharyya}
\bibfield{author}{\bibinfo{person}{Wikipedia}.}
  \bibinfo{year}{2019}\natexlab{}.
\newblock \bibinfo{title}{Bhattacharyya coefficient}.
\newblock
  \bibinfo{howpublished}{\url{https://en.wikipedia.org/wiki/Bhattacharyya_distance}}.
\newblock


\bibitem[\protect\citeauthoryear{Wu, Li, Kumar, Chaudhuri, Jha, and
  Naughton}{Wu et~al\mbox{.}}{2017}]%
        {wu2017bolt}
\bibfield{author}{\bibinfo{person}{Xi Wu}, \bibinfo{person}{Fengan Li},
  \bibinfo{person}{Arun Kumar}, \bibinfo{person}{Kamalika Chaudhuri},
  \bibinfo{person}{Somesh Jha}, {and} \bibinfo{person}{Jeffrey Naughton}.}
  \bibinfo{year}{2017}\natexlab{}.
\newblock \showarticletitle{Bolt-on differential privacy for scalable
  stochastic gradient descent-based analytics}. In
  \bibinfo{booktitle}{\emph{SIGMOD}}. \bibinfo{pages}{1307--1322}.
\newblock


\bibitem[\protect\citeauthoryear{Xing, Ho, Dai, Kim, Wei, Lee, Zheng, Xie,
  Kumar, and Yu}{Xing et~al\mbox{.}}{2015}]%
        {xing2015petuum}
\bibfield{author}{\bibinfo{person}{Eric~P Xing}, \bibinfo{person}{Qirong Ho},
  \bibinfo{person}{Wei Dai}, \bibinfo{person}{Jin-Kyu Kim},
  \bibinfo{person}{Jinliang Wei}, \bibinfo{person}{Seunghak Lee},
  \bibinfo{person}{Xun Zheng}, \bibinfo{person}{Pengtao Xie},
  \bibinfo{person}{Abhimanu Kumar}, {and} \bibinfo{person}{Yaoliang Yu}.}
  \bibinfo{year}{2015}\natexlab{}.
\newblock \showarticletitle{Petuum: A New Platform for Distributed Machine
  Learning on Big Data}. In \bibinfo{booktitle}{\emph{KDD}}.
  \bibinfo{pages}{1335--1344}.
\newblock


\bibitem[\protect\citeauthoryear{Yang, Andrew, Eichner, Sun, Li, Kong, Ramage,
  and Beaufays}{Yang et~al\mbox{.}}{2018}]%
        {yang2018applied}
\bibfield{author}{\bibinfo{person}{Timothy Yang}, \bibinfo{person}{Galen
  Andrew}, \bibinfo{person}{Hubert Eichner}, \bibinfo{person}{Haicheng Sun},
  \bibinfo{person}{Wei Li}, \bibinfo{person}{Nicholas Kong},
  \bibinfo{person}{Daniel Ramage}, {and} \bibinfo{person}{Fran{\c{c}}oise
  Beaufays}.} \bibinfo{year}{2018}\natexlab{}.
\newblock \showarticletitle{Applied federated learning: Improving google
  keyboard query suggestions}.
\newblock \bibinfo{journal}{\emph{arXiv preprint arXiv:1812.02903}}
  (\bibinfo{year}{2018}).
\newblock


\bibitem[\protect\citeauthoryear{Yoon, Kim, Jung, Kang, and Cha}{Yoon
  et~al\mbox{.}}{2012}]%
        {yoon2012appscope}
\bibfield{author}{\bibinfo{person}{Chanmin Yoon}, \bibinfo{person}{Dongwon
  Kim}, \bibinfo{person}{Wonwoo Jung}, \bibinfo{person}{Chulkoo Kang}, {and}
  \bibinfo{person}{Hojung Cha}.} \bibinfo{year}{2012}\natexlab{}.
\newblock \showarticletitle{AppScope: Application Energy Metering Framework for
  Android Smartphone Using Kernel Activity Monitoring}. In
  \bibinfo{booktitle}{\emph{USENIX ATC}}, Vol.~\bibinfo{volume}{12}.
  \bibinfo{pages}{1--14}.
\newblock


\bibitem[\protect\citeauthoryear{Yurochkin, Agarwal, Ghosh, Greenewald, Hoang,
  and Khazaeni}{Yurochkin et~al\mbox{.}}{2019}]%
        {yurochkin2019bayesian}
\bibfield{author}{\bibinfo{person}{Mikhail Yurochkin}, \bibinfo{person}{Mayank
  Agarwal}, \bibinfo{person}{Soumya Ghosh}, \bibinfo{person}{Kristjan
  Greenewald}, \bibinfo{person}{Nghia Hoang}, {and} \bibinfo{person}{Yasaman
  Khazaeni}.} \bibinfo{year}{2019}\natexlab{}.
\newblock \showarticletitle{Bayesian Nonparametric Federated Learning of Neural
  Networks}. In \bibinfo{booktitle}{\emph{ICML}}. \bibinfo{pages}{7252--7261}.
\newblock


\bibitem[\protect\citeauthoryear{Zhang, Gupta, Lian, and Liu}{Zhang
  et~al\mbox{.}}{2016}]%
        {zhang2015staleness}
\bibfield{author}{\bibinfo{person}{Wei Zhang}, \bibinfo{person}{Suyog Gupta},
  \bibinfo{person}{Xiangru Lian}, {and} \bibinfo{person}{Ji Liu}.}
  \bibinfo{year}{2016}\natexlab{}.
\newblock \showarticletitle{Staleness-aware async-sgd for distributed deep
  learning}. In \bibinfo{booktitle}{\emph{IJCAI}}. \bibinfo{pages}{2350--2356}.
\newblock


\bibitem[\protect\citeauthoryear{Zhao, Li, Lai, Suda, Civin, and Chandra}{Zhao
  et~al\mbox{.}}{2018}]%
        {zhao2018federated}
\bibfield{author}{\bibinfo{person}{Yue Zhao}, \bibinfo{person}{Meng Li},
  \bibinfo{person}{Liangzhen Lai}, \bibinfo{person}{Naveen Suda},
  \bibinfo{person}{Damon Civin}, {and} \bibinfo{person}{Vikas Chandra}.}
  \bibinfo{year}{2018}\natexlab{}.
\newblock \showarticletitle{Federated learning with non-iid data}.
\newblock \bibinfo{journal}{\emph{arXiv preprint arXiv:1806.00582}}
  (\bibinfo{year}{2018}).
\newblock


\end{thebibliography}
